\PassOptionsToPackage{unicode}{hyperref}
\PassOptionsToPackage{hyphens}{url}
\PassOptionsToPackage{dvipsnames,svgnames,x11names}{xcolor}
\documentclass[
  a4paper,
]{article}
\usepackage{xcolor}
\usepackage{amsmath,amssymb}
\usepackage{iftex}
\ifPDFTeX
  \usepackage[T1]{fontenc}
  \usepackage[utf8]{inputenc}
  \usepackage{textcomp}%
\else%
  \usepackage{unicode-math}%
  \defaultfontfeatures{Scale=MatchLowercase}
  \defaultfontfeatures[\rmfamily]{Ligatures=TeX,Scale=1}
\fi
\usepackage{lmodern}
\ifPDFTeX\else
\fi
\IfFileExists{upquote.sty}{\usepackage{upquote}}{}
\IfFileExists{microtype.sty}{%
  \usepackage[]{microtype}
  \UseMicrotypeSet[protrusion]{basicmath}%
}{}
\makeatletter
\@ifundefined{KOMAClassName}{%
  \IfFileExists{parskip.sty}{%
    \usepackage{parskip}
  }{%
    \setlength{\parindent}{0pt}
    \setlength{\parskip}{6pt plus 2pt minus 1pt}}
}{%
  \KOMAoptions{parskip=half}}
\makeatother
\usepackage{longtable,booktabs,array}
\usepackage{caption}
\usepackage{calc}%
\usepackage{etoolbox}
\makeatletter
\patchcmd\longtable{\par}{\if@noskipsec\mbox{}\fi\par}{}{}
\makeatother
\IfFileExists{footnotehyper.sty}{\usepackage{footnotehyper}}{\usepackage{footnote}}
\makesavenoteenv{longtable}
\usepackage{graphicx}
\makeatletter
\newsavebox\pandoc@box
\newcommand*\pandocbounded[1]{%
  \sbox\pandoc@box{#1}%
  \Gscale@div\@tempa{\textheight}{\dimexpr\ht\pandoc@box+\dp\pandoc@box\relax}%
  \Gscale@div\@tempb{\linewidth}{\wd\pandoc@box}%
  \ifdim\@tempb\p@<\@tempa\p@\let\@tempa\@tempb\fi%
  \ifdim\@tempa\p@<\p@\scalebox{\@tempa}{\usebox\pandoc@box}%
  \else\usebox{\pandoc@box}%
  \fi%
}
\def\fps@figure{htbp}
\makeatother
\newlength{\cslhangindent}
\newlength{\csllabelwidth}
 {\begin{list}{}{%
  \setlength{\itemindent}{0pt}
  \setlength{\leftmargin}{0pt}
  \setlength{\parsep}{0pt}
  \ifodd #1
   \setlength{\leftmargin}{\cslhangindent}
   \setlength{\itemindent}{-1\cslhangindent}
  \fi
  \setlength{\itemsep}{#2\baselineskip}}}
 {\end{list}}
\usepackage{calc}

\usepackage[a4paper,margin=22mm]{geometry}
\usepackage{microtype}
\usepackage{booktabs,longtable,array,calc}
\usepackage{multirow}
\usepackage{caption}
\usepackage{float}
\usepackage{fontspec}
\usepackage{titlesec}
\titleformat{\section}{\large\bfseries}{}{0em}{}
\titleformat{\subsection}{\normalsize\bfseries}{}{0em}{}
\titleformat{\subsubsection}{\normalsize\itshape}{}{0em}{}
\titleformat{\paragraph}[runin]{\bfseries}{}{0em}{}[.]
\usepackage{fancyhdr}
\AtBeginEnvironment{longtable}{\footnotesize}
\usepackage{etoolbox}
\usepackage[section,above]{placeins}
\usepackage{bookmark}
\IfFileExists{xurl.sty}{\usepackage{xurl}}{}%
\makeatletter
\@ifundefined{xmpquote}{}{}
\makeatother
\hypersetup{
  pdftitle={Reasoning in the loop: antibody lead optimization by probing mechanism and measurement},
  pdfauthor={Samuel Stanton, Aaron Solomon, Emily Delaney, Melanie Montgomery, Crystal Humphries, Timothy Acker, Ryan Henrici, Charbel Eid, Eddie Abrams, Peyton Greenside},
  colorlinks=true,
  linkcolor={NavyBlue},
  filecolor={Maroon},
  citecolor={NavyBlue},
  urlcolor={NavyBlue},
  pdfcreator={LaTeX via pandoc}}

\usepackage[authoryear,round]{natbib}

\title{Reasoning in the loop: antibody lead optimization by probing mechanism and measurement}
\author{Samuel Stanton\textsuperscript{1}\footnote{Corresponding author: samuelstanton@anthropic.com}\hspace{2em}Aaron Solomon\textsuperscript{2}\hspace{2em}Emily Delaney\textsuperscript{2}\hspace{2em}Melanie Montgomery\textsuperscript{2} \\[1ex] Crystal Humphries\textsuperscript{2}\hspace{2em}Timothy Acker\textsuperscript{2}\hspace{2em}Ryan Henrici\textsuperscript{2} \\[1ex] Charbel Eid\textsuperscript{2}\hspace{2em}Eddie Abrams\textsuperscript{2}\hspace{2em}Peyton Greenside\textsuperscript{2} \\[1.5ex] {\small \textsuperscript{1}Anthropic \hspace{2em} \textsuperscript{2}BigHat Biosciences}}
\date{}

\begin{document}
\maketitle

\begin{abstract}

Lab-in-the-loop systems, in which a model proposes molecules, automated assays measure their properties and the results retrain the model, are a popular approach to antibody lead optimization.
These systems optimize for a specification defined with metrics computed from instrument signals, so their success depends on how well those metrics serve as proxies for the properties a drug candidate needs.
When a metric is a poor proxy, neither a better model nor more data will fix it, and identifying the problem usually requires expert review.
We report two campaigns run on BigHat Biosciences' Milliner platform, one on an anti-albumin nanobody (VHH) and one on an anti-mesothelin single-chain variable fragment (scFv), in which a research team and a general-purpose AI agent, Claude, brought scientific reasoning into the loop.
Over seven rounds and about 1,010 design variants we built libraries from antibody lineage, structure and biophysics, and found from the raw curves that on each scaffold one thermal metric did not measure what its name implied.
The libraries were built as controlled comparisons, so the results also show what each mutation contributes to each property.
We fit a physical model of the curves, used it to choose metrics suited to each scaffold, and changed the experimental protocol, varying protein concentration and heating rate, to test the model against pre-registered predictions.
Both leads improve on their parents in affinity, thermal stability and aggregation while improving or maintaining yield and purity.

\end{abstract}

\section{1. Introduction}\label{introduction}

Therapeutic antibodies have transformed the treatment of cancer and autoimmune disease \citep{carter2018}.
An antibody drug candidate must be potent, manufacturable at scale, stable in storage, and safe in patients.
Lead optimization is the step that takes a therapeutic program from an identified starting molecule to IND-enabling studies (i.e., the formal safety work required before a first trial in humans).
The longer and more expensive the lead optimization step is, the fewer clinical candidates a program has to choose from, and ultimately the fewer programs are ready to proceed to the clinic.
Because physical experiments are slow and expensive, a lead-optimization campaign can make only a few hundred high-fidelity measurements.
Each experiment should therefore teach as much as possible, which rules out brute-force or random search over the vast space of possible mutations.
A popular alternative is a closed lab-in-the-loop system, in which a model proposes molecules, the lab measures them, and the results retrain the model.
This formulation is widely shared, from machine-learning-guided directed evolution \citep{wu2019} to Bayesian optimization and generative design of biomolecules \citep{romero2013,stanton2022lambo,khan2023,gruver2023,li2023,rapp2024}.

\begin{figure*}[t]
\centering
\includegraphics[width=0.95\textwidth,alt={Cartoon of the two assays. Panel a shows a sensor tip carrying antigen, with antibody binding and then letting go, the reflected spectrum shifting with the amount bound, and the shift plotted over time. Panel b shows protein in a heated capillary going from mostly folded to mostly aggregated, the emitted and scattered light at each stage, and the two signals plotted against temperature.}]{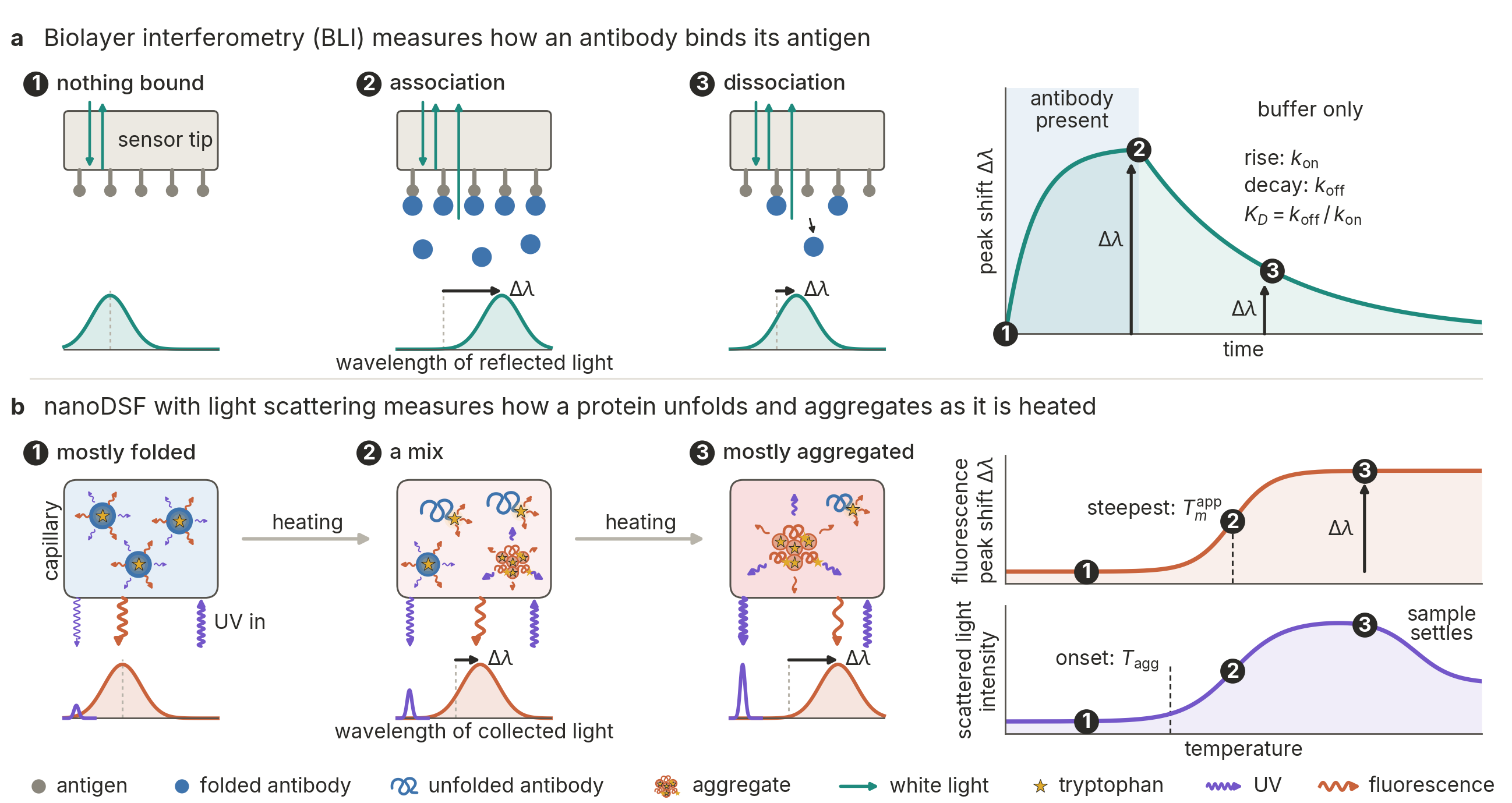}
\caption{\textbf{How the binding and thermal signals are produced.} Numbers link each state to a point on its curve. (a) Biolayer interferometry. White light sent down a sensor tip reflects from the tip surface and from the protein layer, and the two reflections interfere, so more bound antibody moves the reflected peak to longer wavelength. One peak is drawn for clarity. Fitting a kinetic model to the curve gives the on-rate $k_{\mathrm{on}}$, the off-rate $k_{\mathrm{off}}$ and $K_D$. (b) nanoDSF with light scattering. UV light is absorbed by tryptophan, which fluoresces, and scattered by particles. Tryptophan emits at a longer wavelength when unfolding exposes it to water, and larger particles scatter far more, until aggregates settle out of the light path and the signal falls. The order of events is illustrative.}\label{fig:assays}
\end{figure*}

Every lab-in-the-loop system needs a formal specification of the molecule to produce, made up of objective functions (i.e., reward functions) and constraint functions, metrics that say which molecules are admissible and preferable for a program.
Most of what a program cares about cannot be measured directly during lead optimization, so the specification is written in terms of \emph{in vitro} readouts that serve as proxies, for example a thermal stability readout taken over a few hours as a proxy for years of shelf life.
The specification is usually chosen before the first round and then held fixed, so the whole campaign rests on the alignment between these proxy metrics and the program's true goals.
How good that alignment is depends on how each metric is made.
An instrument does not report an intrinsic property of a molecule, it records a signal, and an analysis turns that signal into a metric based on a set of assumptions.
For example, to measure how tightly an antibody binds its target, we use a sensor to follow the two coming together and letting go, and if a simple physics model fits the resulting curve well, its parameters serve as binding metrics (Figure~\ref{fig:assays}a).
Similarly, to measure how readily an antibody aggregates (i.e., sticks to itself and forms clumps), we heat a sample, measure the light it scatters, and report the temperature where the signal starts to rise (Figure~\ref{fig:assays}b).

To a lab-in-the-loop system, both of these metrics are simply labels to predict and improve.
If one of them turns out to be a poor proxy, neither a better predictive model nor more data will fix it, and the problem can only be found by going back to the raw signal and the analysis that produced the metric.
That means inspecting individual instrument traces, understanding what the prior analysis assumed, and testing a better model of the assay, which is usually the job of a biophysicist or experimental biologist who knows both the instrument and the molecule.
This is not the role typically envisioned for AI in lead optimization.
Even recent agentic campaigns have used general-purpose models chiefly to orchestrate specialized design tools under a predefined protocol \citep{shaneh2026}, leaving what to measure and how to read it to people.
In this report we ask what a research team and a general-purpose AI agent can accomplish when they work together on live optimization campaigns.

We report two lead-optimization campaigns run between June and September 2026 on BigHat Biosciences' Milliner platform, which carried out every experiment, one on an anti-albumin VHH (the single variable domain of a camelid heavy-chain antibody) and the second on an anti-mesothelin scFv (the two variable domains of an antibody joined by a linker), over seven rounds and about 1,010 design variants (1,232 wells).
A lab-in-the-loop campaign typically builds each library with specialized machine learning models, trained on earlier rounds to propose the designs predicted to score best \citep{frey2025}.
In this work a human researcher and Claude, working in Claude Code, took a different approach to both library design and measurement.
To design libraries they reasoned as antibody engineers would, tracing each parent's lineage through the patent and structural literature, running small experiments to find which positions mattered, and proposing mutations that tested specific hypotheses about binding and stability.
To interpret the results they asked what developability means apart from any one assay, and which intrinsic properties of the molecule the thermal metrics were meant to estimate.
Reading the raw curves with that question in mind, they found that on the VHH the aggregation onset sat near 40 °C regardless of the melt, while on the scFv the reported melt marked the sample precipitating.
They then built a physical model of the curves, replaced those metrics with an aggregation rate on the VHH and a solution collapse temperature on the scFv, and predicted how each metric should respond to changes in protein concentration and heating rate before running those experiments.
The VHH lead binds albumin about nine times more tightly than its parent, melts 8 °C higher and shows no aggregation onset across the thermal ramp, and the scFv lead binds mesothelin about six times more tightly and precipitates 18 °C higher.

\section{2. Background}\label{background}

\begin{figure*}[t]
\centering
\includegraphics[width=0.95\textwidth,alt={Three panels of two signals against temperature, scattered light and fluorescence shift. In the reference panel, a dot on the steepest point of the fluorescence curve marks the apparent melting temperature, and a dot where the scattering curve departs from its baseline marks the aggregation onset. With half the protein, or heated three times faster, the scattering curve and its onset move to higher temperature while the fluorescence curve stays where it was.}]{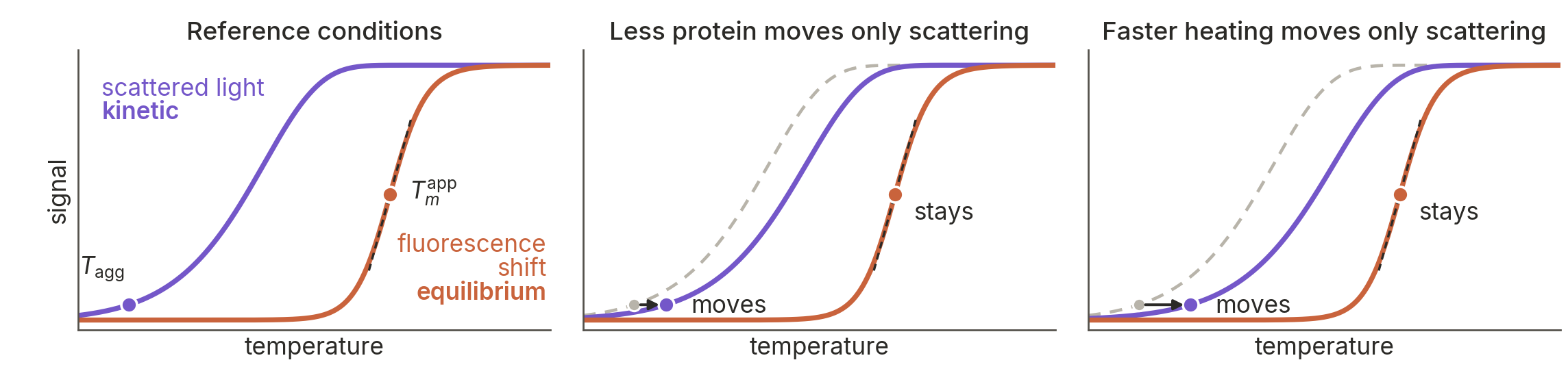}
\caption{\textbf{How $T_m^{\mathrm{app}}$ and $T_{\mathrm{agg}}$ are computed, and how they respond to the protocol.} Schematic of the textbook expectation, with shifts exaggerated. $T_m^{\mathrm{app}}$ (orange dot) is the steepest point of the fluorescence shift. $T_{\mathrm{agg}}$ (purple dot) is where the scattered light departs from its baseline, drawn here at an illustrative level. Halving the protein or heating faster moves $T_{\mathrm{agg}}$ to higher temperature and leaves $T_m^{\mathrm{app}}$ in place. Gray dashed curves repeat the reference conditions. The fluorescence is an equilibrium signal, so it has no memory of how the sample got there, while the scattering is kinetic and depends on the path.}\label{fig:meltevent}
\end{figure*}

We pose lead optimization as the problem of finding sequences $x$, near a parent sequence $x_0$, that satisfy a specification as closely as possible (i.e., that minimize their divergence from it) in a small number of rounds.
The specification is written on metrics such as $K_D$, $T_m^{\mathrm{app}}$ and $T_{\mathrm{agg}}$, so to understand what is being optimized we need to know how each metric is made.

\textbf{What the assays measure.} To measure binding, a sensor records a signal as an antibody associates with its antigen and then lets go, and fitting a kinetic model to the whole curve gives an off-rate $k_{\mathrm{off}}$ and a dissociation constant $K_D$ (Figure~\ref{fig:assays}a).
We report the off-rate as $\mathrm{p}k_{\mathrm{off}} = -\log_{10} k_{\mathrm{off}}$, so that higher is slower dissociation and tighter binding.
To measure thermal stability, a small capillary of protein solution is heated at a steady rate under UV light, and two signals are recorded as the temperature rises (Figure~\ref{fig:assays}b).
The first is intrinsic fluorescence: tryptophan residues absorb the UV light and emit at a longer wavelength, longer still when unfolding exposes them to water, so the emission spectrum shifts as the protein unfolds.
The second is scattered light: particles scatter the UV light, and larger particles scatter far more, so the signal rises as molecules clump together.
The instrument's software reduces these curves to two metrics.
The first is $T_m^{\mathrm{app}} = \arg\max_T |d\bar\lambda/dT|$, the steepest point of the fluorescence shift, where $\bar\lambda(T)$ is the mean emission wavelength.
The second, $T_{\mathrm{agg}}$, is the temperature at which the scattered intensity departs meaningfully from its baseline, as called by the software.

\textbf{A physical model of heated protein.} To interpret these metrics we need a physical model.
In the simplest one, each molecule is folded, unfolded, or aggregated (i.e., clumped).
Folding and unfolding are reversible, so at each temperature the two populations settle at an equilibrium, and the folded fraction is
\begin{equation}\label{eq:phi} \phi_N(T) = \Big[1 + \exp\!\Big(\frac{\Delta H}{R}\Big(\frac{1}{T_m} - \frac{1}{T}\Big)\Big)\Big]^{-1}, \end{equation}
where $\Delta H$ is the enthalpy of unfolding, $R$ is the gas constant, and the thermodynamic melting temperature $T_m$ is the temperature at which half the molecules are folded.

Aggregation, by contrast, is irreversible \citep{lumry1954, sanchezruiz1992}, and it can start from either folded or unfolded protein \citep{chi2003, andrews2007}.
Because aggregated molecules do not disaggregate, the amount aggregated depends on how long the sample has spent at each temperature, not only on the temperature itself.
Suppose each molecule aggregates at a rate $c\,k(T)$, proportional to the protein concentration $c$, with a rate constant $k(T)$ that grows with temperature.
If the sample heats at a rate $\beta$, the fraction $m$ of protein not yet aggregated falls as
\begin{equation}\label{eq:agg} \frac{dm}{dT} = -\frac{c}{\beta}\,k(T)\,m. \end{equation}
Unfolding is therefore an \emph{equilibrium} process, in which only the current temperature matters, however the sample got there.
Aggregation is a \emph{kinetic} process, in which the amount aggregated depends on the rate of the reaction and the time the sample has spent at each temperature.
Neither $c$ nor $\beta$ appears in Equation~\ref{eq:phi}, so $T_m$ depends on neither.
Both appear in Equation~\ref{eq:agg}, so the temperature at which aggregation appears moves higher with faster heating \citep{sanchezruiz1988, goyal2014} or less protein \citep{andrews2007} (Figure~\ref{fig:meltevent}).

\textbf{From metrics to labels.} The reported $T_m^{\mathrm{app}}$ is a \textit{summary statistic} of the fluorescence signal, the steepest point of a curve, while $T_m$ is a \textit{parameter} of the physical model.
The statistic estimates the parameter only if the protein unfolds in two states, reversibly, and at equilibrium \citep{greenfield2006, kotov2021}.
Otherwise it is only an \textit{apparent} melting temperature, as the superscript indicates, and it can change with protein concentration or heating rate.
An optimization loop sees neither the signal nor the physics, only a label, which we write $y_i = f(x_i; \omega) + \epsilon_i$ for sequence $x_i$ under protocol $\omega$ (e.g., instrument, sample concentration, heating rate).
Replicates average away $\epsilon_i$, but they cannot correct an $f$ that reports a different quantity, a problem known as \textit{misspecification}.
In lead optimization every design stays near $x_0$, so the distribution of sequences changes relatively little from one round to the next.
If a label reflects a stable property of the sequence, a model trained on one round should therefore predict the next, and a failure to do so suggests a problem with the label rather than with the model.

\section{3. Related work}\label{related-work}

\textbf{Model-based optimization.} Machine learning entered protein engineering through directed evolution, where models trained on measured variants choose what to test next \citep{romero2013, wu2019, yang2019}.
Generative methods followed, both for \textit{de novo} design \citep{dauparas2022, watson2023} and for optimization loops that pair a generative model with a discriminative model that ranks variants \citep{stanton2022lambo, gruver2023}.
For antibodies, models have predicted antigen specificity \citep{mason2021}, proposed mutations from language models alone \citep{hie2023}, co-optimized affinity and non-specific binding \citep{makowski2022}, and run a loop across four targets and eleven lead antibodies \citep{frey2025}.
The model may not even be the main constraint, since general-purpose language models can be trained to solve constrained biophysical sequence optimization tasks that specialized solvers also solve, leaving the solver an advantage mainly in efficiency \citep{chen2024}.
Objectives come from the program's target candidate profile (i.e., the properties a candidate must have, which can be revised as the program learns), and computational guidelines drawn from clinical-stage antibodies offer one screen for developability \citep{raybould2019}.
Most of this work treats reported assay values as correct labels. We also tested whether those values measured the intended property.

\textbf{Agents in scientific campaigns.} Recent work hands whole campaigns to general-purpose agents.
Such systems have designed nanobodies against SARS-CoV-2 \citep{swanson2025}, generated and triaged hypotheses across biomedical problems \citep{gottweis2026}, executed diverse biomedical research tasks \citep{biomni2026}, and automated the production of research reports \citep{lu2026}.
Closest to this work, an agent given a written protocol and open-source design tools ran \textit{de novo} binder design campaigns against sixteen targets, with no human input during optimization \citep{shaneh2026}, and a related system supplies campaign-level control for high-throughput binder design \citep{jeon2026}.
The former leaves open questions, since the agent primarily orchestrates pre-existing tools, mini-proteins are not yet a proven therapeutic protein modality, biologists wrote the protocol and configured the experiments, and binding is only one of several properties that decide a discovery program.
A recent position paper argues that agents are not built for autonomous discovery, in part because problem selection depends on human judgment and training corpora omit the tacit knowledge of laboratory practice, and endorses agents as co-scientists, with a person retaining at least one of problem specification, experimental work or interpretation \citep{bisht2026}.
Our campaigns fit that model. Researchers provided the molecule specifications and ran the experiments, and the agent did much of the analysis and interpretation.

\textbf{Developability metrics as surrogate objectives.} Whether a molecule can be manufactured, stored and delivered as a drug (i.e., its developability) cannot be measured directly during lead optimization, so programs rely on panels of rapid biophysical assays.
\citet{jain2017} ran a dozen such assays on 137 antibodies in advanced clinical stages and found that the number of ``red flags'' a molecule carries declines with advancement toward approval.
The resulting metrics are surrogates for the properties a program cares about, and clinical trials face the same problem with surrogate endpoints (i.e., measurements that stand in for a clinical outcome).
A surrogate is useful when effects on it predict effects on the endpoint, but correlation alone does not establish this, because an intervention can move a surrogate through a mechanism independent of the process of interest \citep{fleming1996}.
Optimization sharpens the problem, since selecting hard on a proxy tends to break the relationship that justified it \citep{manheim2018}, as seen with learned reward models \citep{gao2023}.
For thermal metrics, both the failure and its diagnosis are long established (Section 2).
Irreversible aggregation shifts the apparent transition temperature lower rather than adding noise \citep{lepock1992}, aggregation propensity need not follow thermodynamic stability \citep{ebo2020}, and the standard test is whether a metric moves with heating rate or concentration \citep{sanchezruiz1988, goyal2014}.

\begin{table}[t]
\centering
\footnotesize
\begin{tabular}{llllll}
\toprule
 & & $K_D$ (nM) & $\hat{T}_m$ or $T_m^{\mathrm{app}}$ (°C) & $T_{\mathrm{agg}}$ (°C) & Yield (µg/mL) \\
\midrule
\multirow{4}{*}{VHH} & Specification & $\leq 1.6$ & $\geq 70$ & $\geq 70$ & $> 10$ \\
 & Supplied baseline & 16 & 61.8 & -- & 21.5 \\
 & Parent, in campaign & 10.3 & 62.2 & 40--45 & 43 \\
 & Lead & 1.15 & 70.6 & no onset to 85 & 51 \\
\midrule
\multirow{5}{*}{scFv} & Specification & $\leq 1.5$ & $\geq 65$ & $\geq 65$ & $> 10$ \\
 & Supplied baseline & 15.2 & 54.9 & -- & 23.8 \\
 & Parent, in campaign (DR1) & 3.7 & 54.6 & 50.5 & 16 \\
 & Parent, in campaign (DR0) & 4.3 & 55.0 & 48.3 & 39 \\
 & Lead & 0.63 & 75.1 & 57.7 & 41 \\
\bottomrule
\end{tabular}
\caption{Original specifications, supplied baselines, in-campaign parent values, and the two leads. Yield is purified protein per volume of cell-free reaction. VHH melt values are identified melts ($\hat{T}_m$) and scFv values are as reported ($T_m^{\mathrm{app}}$), for reasons given in Section 7. The VHH lead shows no scattering onset across the thermal ramp, so its $T_{\mathrm{agg}}$ is a bound. scFv parent values come from design rounds 0 and 1 (DR0, DR1), and comparisons in the text use DR1. Full tables with $n$ are in Appendix H.}
\label{tab:gates}
\end{table}

\section{4. Setup}\label{setup}

\textbf{Platform.} Both campaigns ran on BigHat Biosciences' automated Milliner platform, which expresses variants cell-free (i.e., without growing cells) and measures binding, thermal stability, aggregation, purity and yield at high throughput.
All experimental and computational work sits in one system of record, Reccy, which an agent can reach through a Model Context Protocol (MCP) connector, a standard interface between a model and external tools.
Raw instrument traces are stored beside the summary metrics and the quality-control flags.
The analysis in Section 6 relies on this access to the raw traces. Appendix A gives the methods for each assay.

\textbf{Campaigns.} The VHH campaign started from Alb8, a 115-residue anti-albumin domain from Ablynx \citep{ablynx2006} (Figure~\ref{fig:structure}), and the scFv campaign from a 259-residue anti-mesothelin fragment derived from Medarex clone 7B1 \citep{medarex}, both supplied by BigHat with baseline measurements.
Between 29 June and 4 September 2026 we ran four design rounds on the VHH and three on the scFv. Each round tested 176 wells of design variants, and results typically returned in 9 to 10 days.
The AI agent was Claude in Claude Code, primarily Claude Fable 5, directed by S.S.

\textbf{Specifications.} The campaign specifications cover affinity, $T_m^{\mathrm{app}}$, $T_{\mathrm{agg}}$ and yield (Table~\ref{tab:gates}, which also shows where each lead finished).
The supplied baselines and the in-campaign parent measurements differ, most of all for the scFv's affinity, and Table~\ref{tab:gates} gives both.

\textbf{Campaign journal.} We kept an append-only journal of every design, analysis, finding and decision, with an author on each record (387 records through the lead decisions).
Predictions for the later design rounds were written to file before the plates ran.

\section{5. Designing by biophysical reasoning}\label{design}

\begin{figure*}[t]
\centering
\includegraphics[width=\linewidth,keepaspectratio,alt={Five panels of evidence for the design strategies: literature relatives of Alb8 with their off-rates, a schematic of the tyrosine scan, the six VHH factorial mutations on the crystal structure with their effects on off-rate, melt and aggregation, the scFv final-round cube by collapse temperature and off-rate, and Walsh effects of the eight somatic reversions on binding.}]{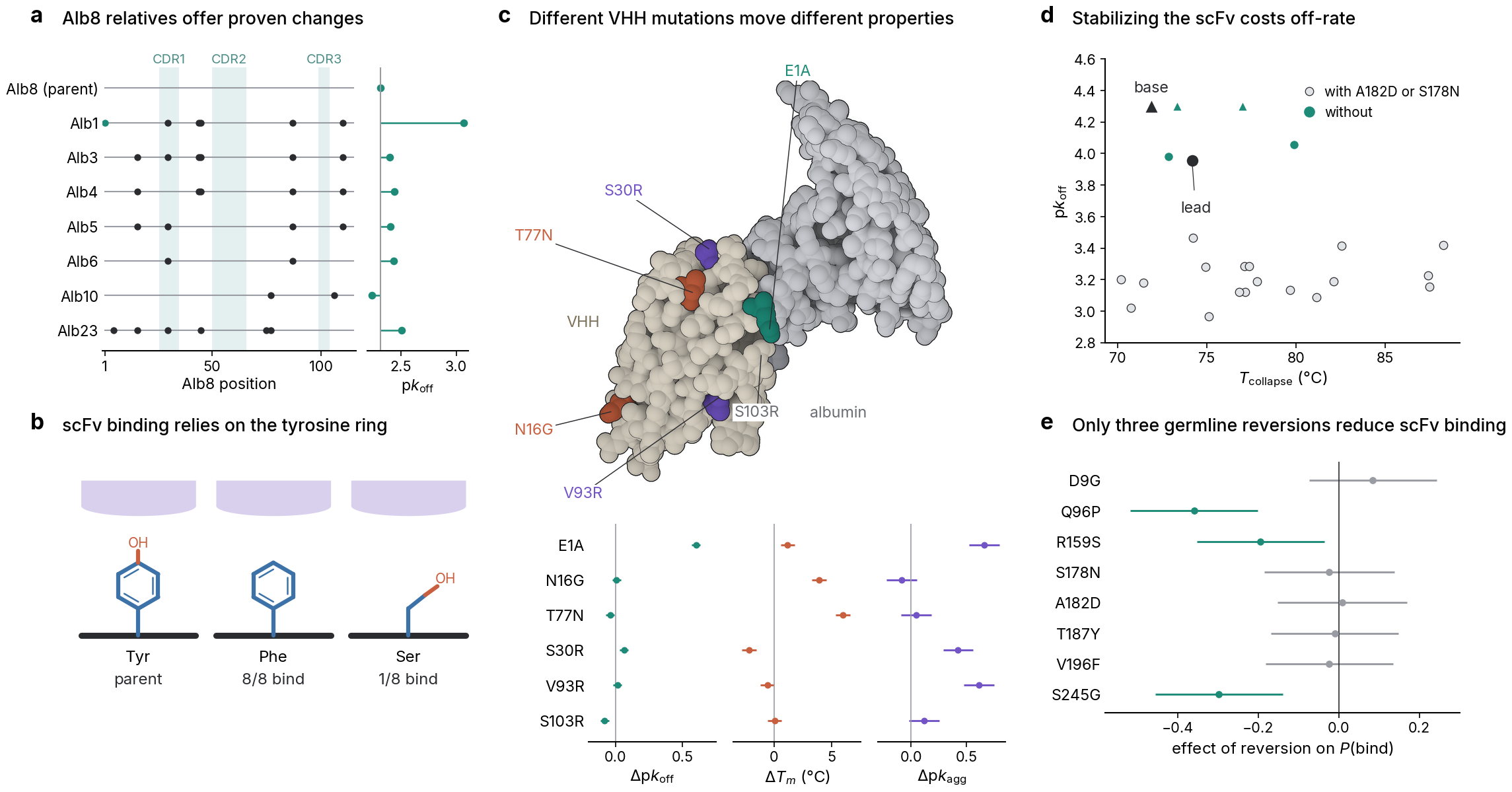}
\caption{\textbf{Evidence for each design strategy.} (a) Published relatives of Alb8. Dots mark substitutions (E1A in teal), shading marks the CDRs, and the right panel gives each molecule's round-0 off-rate. (b) Schematic of the tyrosine scan on the scFv, with the fraction of variants that bound. (c) The six mutations of the VHH final-round factorial on the Alb8:albumin structure (PDB 8Z8V; \citealp{mima2024}), colored by the property each moves most (teal off-rate, orange melt, purple aggregation), and the effect of adding each one, $\pm 2\sigma$ ($n = 63$, 52 and 57 designs). (d) The scFv final-round factorial on the base design (the round 1 control), by collapse temperature (where the sample falls out of solution) and off-rate. Triangles have $\mathrm{p}k_{\mathrm{off}} \geq 4.3$. (e) Effect of reverting each somatic mutation to germline on the probability of binding, $\pm 2\sigma$ ($n = 58$). Positions that carry binding are in teal.}\label{fig:strategies}
\end{figure*}

In both campaigns we designed libraries as antibody engineers and biophysicists might, with each part of a library testing a stated hypothesis about what a part of the molecule was doing.
The agent proposed hypotheses and mutations to test them, and the research team set the design strategy, reviewed the designs and chose which went on each plate.
Mechanistic reasoning is powerful but can slip into post-hoc rationalization, since more than one mechanism can often explain a result.
We therefore built libraries as designed experiments and read them with a simple statistical model.

\textbf{Priors from lineage and structure.} An antibody descends from its germline (i.e., the inherited gene sequences) through somatic mutations acquired as the immune system refines it, and a therapeutic parent often carries further engineering, so its relatives offer a ready-made library of changes that have already worked.
On the first day the agent identified the VHH as Ablynx's Alb8, matching at all 115 positions, and traced the scFv to Medarex clone 7B1, from which it differs at eight positions.
On the VHH, substitutions from the published successor Alb23 \citep{ablynx2012} and other members of the Alb family \citep{ablynx2006} went onto the round-0 plate, and every one of those designs bound (Figure~\ref{fig:strategies}a).
They supplied E1A, the largest affinity gain on this scaffold, and N16G, T77N and L45P, all of which reached the lead.
On the scFv, all five designs reverting the three differences in CDR-H3 (the heavy-chain loop at the center of most binding sites), singly or together, bound, three of them beyond the assay's dissociation floor (i.e., too slow to resolve in the measurement window), and one reversion, G230Y, carried the largest affinity gain on this scaffold.
On the VHH, structure bounded the search, since a related VHH's complex with albumin \citep{defelice2023} showed what could change without touching the binding mode (Figure~\ref{fig:structure}).

\textbf{Substitutions as chemical questions.} Amino acids differ in known ways, so a well-chosen substitution asks a precise question about a position, and a scan repeats that question across a region.
On the scFv we asked whether eight tyrosines in CDR-H3 contributed through their aromatic ring or their hydroxyl, replacing each with phenylalanine (ring, no hydroxyl) and with serine (hydroxyl, no ring).
All 8 phenylalanine variants bound, against 1 of 8 serine variants, which suggests that binding relies on the ring (Figure~\ref{fig:strategies}b).
On the VHH, aspartate at position 1 was far worse than alanine or serine, consistent with an acidic side chain facing an acidic patch on albumin.
In both campaigns, alanine scans of the CDRs and the Vernier-zone framework positions, which sit beneath the CDR loops and affect binding without contacting the antigen \citep{foote1992, makabe2008}, mapped what each position contributed.
On the scFv they also found two packing repairs, Q91A and S245A, each worth 6--7 °C on the collapse temperature, where the sample falls out of solution (Section 6), and we conjecture that these help the variable domains stay together, since an scFv's domains stabilize each other \citep{jager1999}.

\textbf{Separating what a mutation does.} Binding, folding and self-association are distinct physical processes, so we asked which of them each mutation moved.
On the VHH, the first charge hypothesis failed.
Cationic substitutions intended to speed association with albumin did not do so in any variant that still bound.
The raw curves then pointed elsewhere, since E1A and S30R together left the melt near the parent's and cut the rise in scattering about sixfold.
Across VHH designs, surface cationic mutations (S30R, S103R, V93R) and removal of anionic residues (E1A, D73A) reduced aggregation with little or no gain in melt.
The VHH parent began to aggregate about 20 °C below its melt, while still folded, so we treated charge as a way to keep folded molecules apart, separate from their stability, an established strategy for single domains in which the useful sign of charge depends on the scaffold \citep{dudgeon2012, perchiacca2014}.

\begin{figure*}[t]
\centering
\includegraphics[width=\linewidth,keepaspectratio,alt={Three panels of measured points with fitted curves. Panel a shows a binding curve with its kinetic fit. Panels b and c show fluorescence and scattered light against temperature for the VHH and scFv parents, with capillary cartoons above and the reported metrics marked.}]{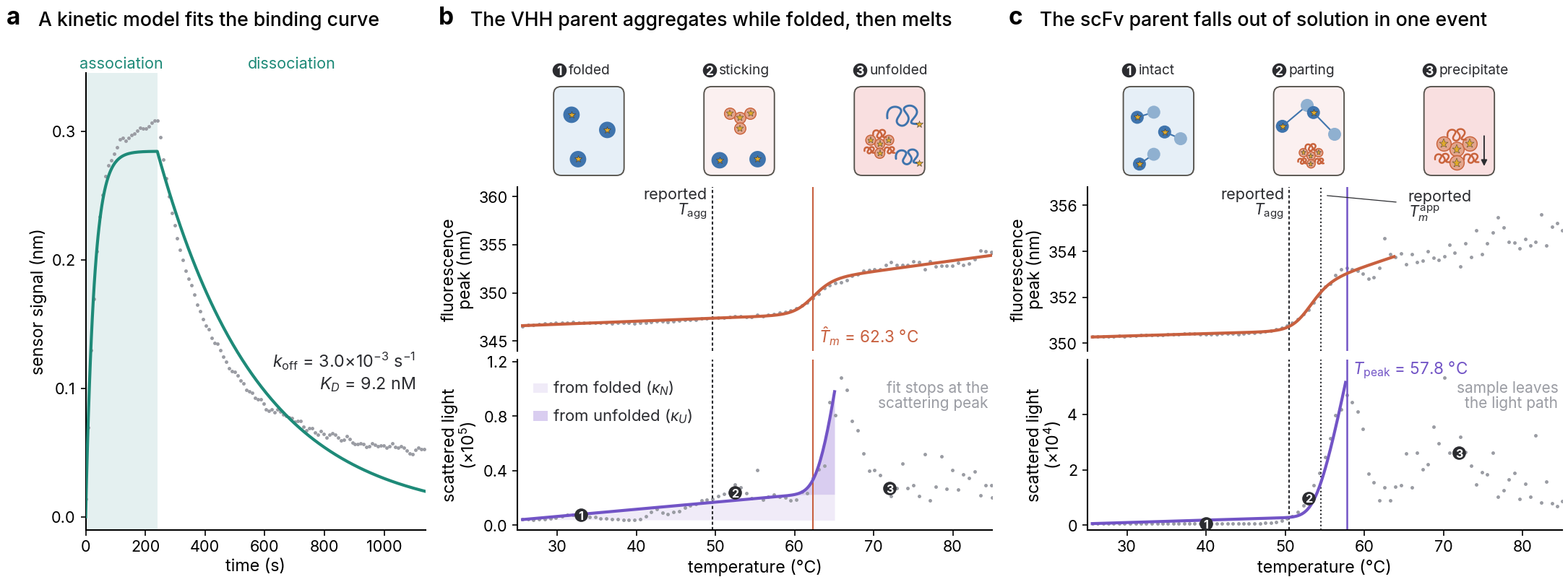}
\caption{\textbf{Reading whole curves.} Gray points are one well and colored lines are fits. (a) The VHH parent in the binding assay, with the fitted 1:1 kinetic model. (b) The VHH parent in the thermal assay, with Equations~\ref{eq:F} and \ref{eq:S} fit to fluorescence (top) and 266 nm scattered light (bottom). Shading splits the scattering fit above the instrument baseline into the two terms of Equation~\ref{eq:S}. The dashed line is the reported $T_{\mathrm{agg}}$, which in this well (49.6 °C) falls later than the parent's usual 40--45 °C, after the scattering has already tripled. (c) The scFv parent in round 1, fit the same way up to the scattering peak. $T_{\mathrm{peak}}$ is shown before correction for concentration, and the dashed and dotted lines are the reported $T_{\mathrm{agg}}$ and $T_m^{\mathrm{app}}$. The fluorescence fit in (c) is descriptive, since the sample leaves solution before an unfolded baseline is reached, so its midpoint is not a melt (Appendix D.4). Numbers link the cartoons to the scattering curves.}\label{fig:fit}
\end{figure*}

\textbf{Combining modules and balancing trade-offs.} Mutations that act through separate mechanisms can be treated as modules and combined, so that a gain from one module offsets a cost from another.
On the scFv, reversions to germline that improved stability also cost binding, and in round 1 designs carrying them as a module bound in 0 of 7 cases without G230Y and Y238A, against 14 of 15 with them (G230Y tightened binding and Y238A offset its cost in aggregation).
In the same round, a simple control combining G230Y and Y238A with the two packing repairs outperformed the variants designed from new hypotheses, including added disulfide bonds that did not improve stability, and it became the starting point of the final round.
Further stabilizing mutations cost off-rate, so the scFv lead took two of the milder ones, T187Y and V196A (Figure~\ref{fig:strategies}d, Appendix C.6).

\textbf{Checking the reasoning.} We tested these hypotheses with combinatorial libraries, in which each mutation is either present or absent, so that each design is a point on a Boolean lattice.
If only a few mutations carry a function, a small sample of the lattice is enough to find them \citep{poelwijk2019}.
We fit these samples with Bayesian linear regression in the Walsh basis, the Fourier basis of Boolean functions \citep{poelwijk2016}, which gives each mutation or combination of mutations its own coefficient and uncertainty (Appendix C.3).
On the scFv, a sample of 63 of the 256 combinations of the eight somatic mutations showed that only three carried binding, and the effects at the other five were consistent with zero (Figure~\ref{fig:strategies}e).
In the final round, a full factorial (i.e., all combinations) of stabilizing mutations at four of those positions tested them directly, and A182D, V196F and T187Y added 8.1, 4.2 and 2.0 °C of collapse temperature (Appendix C.6).
On the VHH, a full factorial over six mutations in the final round showed that E1A carried the off-rate gain, N16G and T77N the melt, and E1A, S30R and V93R slowed aggregation, with effects that added (Figure~\ref{fig:strategies}c, Appendix C.5).
We tested computational tools the same way, and on the scFv co-folding scores did not separate binders from decoys designed for another target (Appendix F.2).
On the VHH, off-rate could be predicted from one round to the next, while the reported thermal metrics could not (Section 6).

\section{6. Reassessing the thermal metrics}\label{thermal-metrics}

A heated protein can do two things.
It can unfold and refold, quickly and reversibly, so the fraction unfolded depends only on the temperature.
It can also stick to other molecules, and since aggregates rarely come apart, the amount aggregated grows with time and with protein concentration.
The first process has a natural temperature, $T_m$.
The second does not, since how much has aggregated at a given temperature depends on how the sample got there.

\textbf{The first clues.} Our thermal assay (nanoDSF) watches both processes at once (Figure~\ref{fig:assays}b).
As the sample heats, the fluorescence shifts when the protein unfolds and the scattered light rises when it aggregates.
At the start, the specifications were written on two summaries of these curves, $T_m^{\mathrm{app}}$ and $T_{\mathrm{agg}}$.

\begin{figure*}[t]
\centering
\includegraphics[width=0.95\linewidth,keepaspectratio,alt={Fluorescence and scattered light against temperature with parent and lead overlaid for the VHH and the scFv, and paired shifts per design for the VHH melt, the VHH scattering onset and the scFv collapse at half load and at the faster ramp.}]{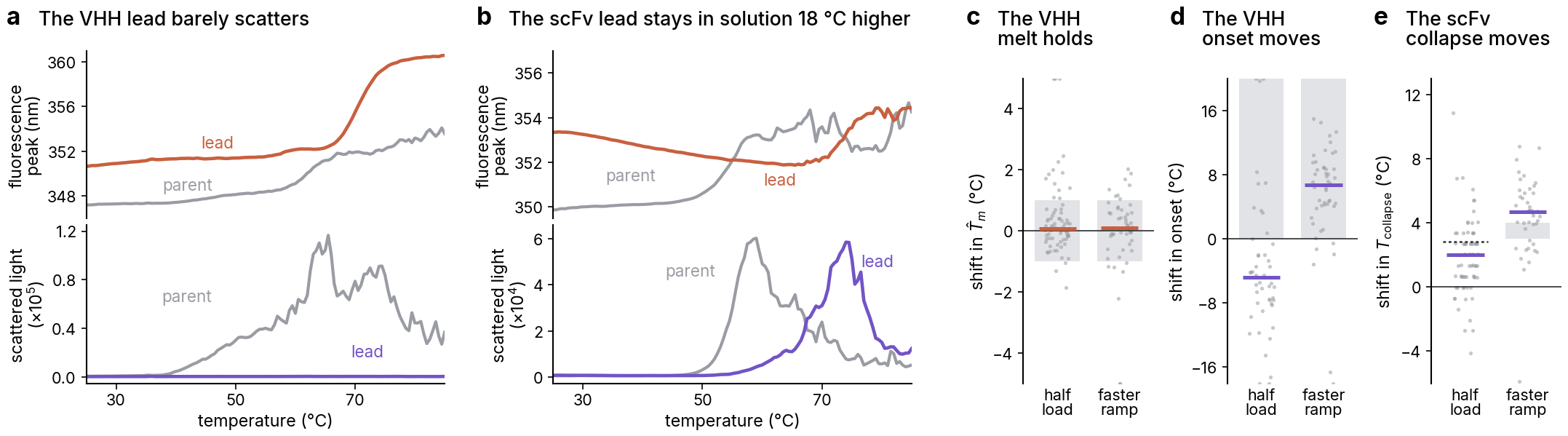}
\caption{\textbf{Parents and leads, and the test of the model.} (a) VHH and (b) scFv fluorescence (top) and 266 nm scattered light (bottom) against temperature, averaged over each molecule's wells in one round. The scFv parent is from round 1 and its lead from round 2. (c--e) Paired shifts per design at half the protein concentration (half load) and at the faster ramp, for the VHH identified melt, the VHH scattering onset and the scFv collapse temperature. An equilibrium quantity should hold and a kinetic one should move. Bars are medians, gray bands are predicted ranges (in d, the predicted direction), the dotted line is a predicted value, and points beyond the axes are drawn at the edge ($n = 74$ and 55, 49 and 49, 74 and 42).}\label{fig:traces}
\end{figure*}

After the first VHH design round, models trained on round 0 predicted round 1 $\mathrm{p}k_{\mathrm{off}}$ with a Spearman correlation of $\rho = 0.59$, while on reported $T_m^{\mathrm{app}}$ and $T_{\mathrm{agg}}$ none of seven models, built on sequence, structural and protein-language-model features, exceeded $\rho = 0.07$, a sign of misspecification (Section 2).
BigHat had independently raised the same concern.
Over the four rounds, their quality controls flagged 491 of 916 VHH wells as aggregating before a $T_m^{\mathrm{app}}$ call, and their scientists cautioned that a $T_m^{\mathrm{app}}$ above $T_{\mathrm{agg}}$ can reflect the melting of an aggregate and that $T_{\mathrm{agg}}$ depends on protein concentration.
After round 1 they provided the raw curves, which confirmed these flags.
Simple summaries of the same curves could be predicted across rounds ($\rho = 0.38$), so the curves held a signal that the reported metrics missed, which raised the question of what those metrics were measuring.

\textbf{A model of the curves.} Binding offered a template.
We read it by fitting a kinetic model to the whole BLI curve (Figure~\ref{fig:fit}a), and the fitted rates describe an intrinsic property of the molecule.
Why not read developability the same way?
We modeled each whole thermal curve as the two processes of Section 2 (Figure~\ref{fig:fit}b, Appendix D).
Fluorescence reports where the emission peak sits, and folded and unfolded molecules emit at different wavelengths, so the observed peak is an average over the two populations:
\begin{equation}\label{eq:F} F(T) = (a_N + s_N T)\,\phi_N(T) + (a_U + s_U T)\,\big[1 - \phi_N(T)\big]. \end{equation}
Each state has its own slowly drifting baseline, with intercept $a$ and slope $s$, and the gap between the two baselines is the shift on unfolding.
Only the current temperature appears, so $F$ has no memory.
We substitute Equation~\ref{eq:phi} for $\phi_N$ and fit the baselines, $T_m$ and $\Delta H$ to the measured curve.
$T_m$ sets where the curve crosses from one baseline to the other and $\Delta H$ sets how sharply it does, so the fit gives $\phi_N(T)$ at every temperature.
We write $\hat{T}_m$ for the model's estimate of $T_m$, the \textit{identified melt}.

Scattering reports particles, which accumulate as in Equation~\ref{eq:agg}, so we model its slope above the instrument baseline:
\begin{equation}\label{eq:S} \frac{dS}{dT} = \kappa_N\, \phi_N(T)^2 + \kappa_U\, \big[1 - \phi_N(T)\big]^2. \end{equation}
Below the melt $\phi_N \approx 1$, so the slope is $\kappa_N$, the rate of growth from folded protein.
Above the melt $\phi_N \approx 0$, so the slope is $\kappa_U$, the rate of growth from unfolded protein.
The squares come from a mean-field argument for early aggregation, in which two molecules meeting at random are both folded with probability $\phi_N^2$.
This is a phenomenological ansatz and not a derived rate law, since it neglects mixed encounters and later growth of aggregates, and a form linear in $\phi_N$ fits about as well (Appendix D).
The value of $S$ at any temperature is everything accumulated on the way there, which is the sample's history.
If scattering rises while $\phi_N$ is still near one, only folded protein can be responsible, and if it waits until the melt, unfolding is involved.

\textbf{What to measure.} Each metric should report one physical process, so what to measure follows from what the sample is doing.
Where a transition is visible, the identified melt $\hat{T}_m$ reports the equilibrium.
Where folded protein aggregates well below its melt, the fitted rate $\kappa_N$ reports the kinetics, and dividing by the protein concentration gives $k_{\mathrm{agg}} = \kappa_N/c$, which we report as $\mathrm{p}k_{\mathrm{agg}} = -\log_{10} k_{\mathrm{agg}}$, so that higher is slower, as with $\mathrm{p}k_{\mathrm{off}}$.
Where the sample precipitates before unfolding completes, there is no melt to report.
The scattering curve instead rises as aggregates grow and falls as they leave the light path, peaking at a temperature $T_{\mathrm{peak}}$ that shifts with the protein concentration $c$.
Correcting the peak to a reference concentration $c_{\mathrm{ref}}$ gives the collapse temperature,
\begin{equation}\label{eq:collapse} T_{\mathrm{collapse}} = T_{\mathrm{peak}} - \gamma_c \ln\!\big(c / c_{\mathrm{ref}}\big), \end{equation}
where the slope $\gamma_c$ is fit on paired concentrations (Appendix D.5).
Fitted melts and concentration-scaled aggregation rates are established quantities \citep{kotov2021, andrews2007}.

\textbf{Validating the model.} The model makes a sharp prediction, since equilibrium quantities should ignore protein concentration and heating rate, and kinetic ones should not.
A metric chosen only because it could be predicted would have no reason to pass this test.
For the final round of each campaign the team agreed on two protein concentrations and two heating rates, and the agent predicted which metrics should move before the plates ran (Table~\ref{tab:e1}).
BigHat reconfigured the platform to run it, doubling the synthesis scale.

\section{7. Results}\label{results}

\begin{figure}[t]
\centering
\includegraphics[width=0.92\linewidth,keepaspectratio,alt={Final-round designs against the affinity and revised thermal constraints. (a) VHH: off-rate against the aggregation metric. (b) scFv: off-rate against collapse temperature.}]{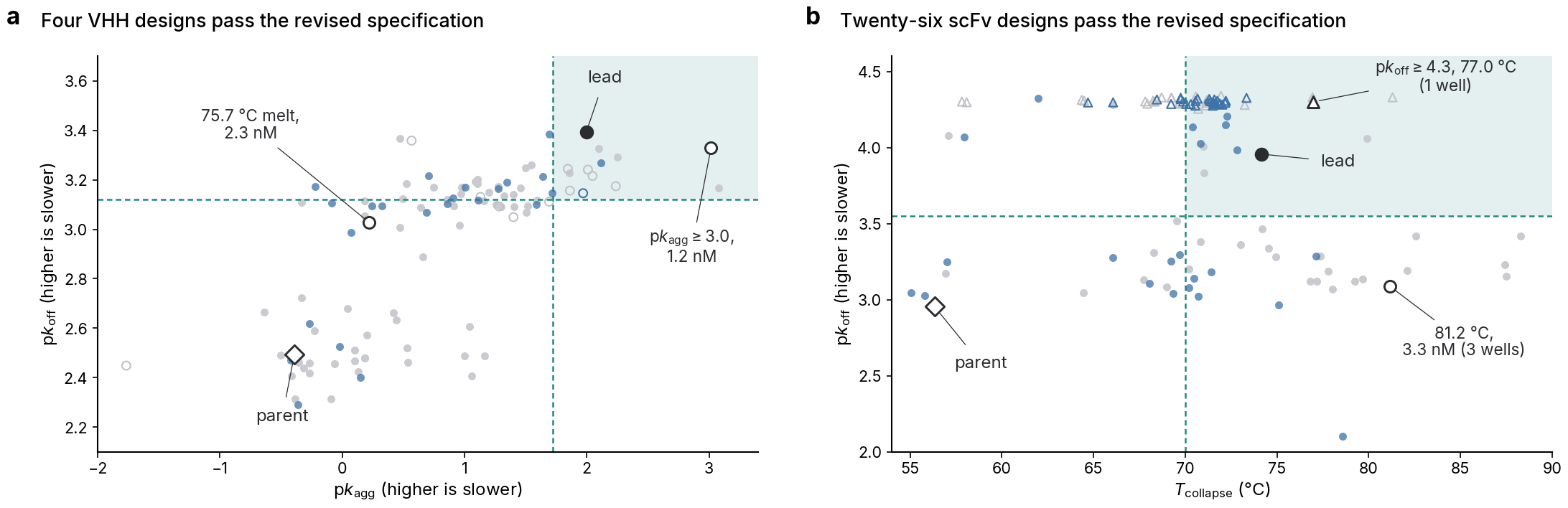}
\caption{\textbf{Final-round designs against the revised specifications.} (a) VHH: off-rate against the aggregation metric \(\mathrm{p}k_{\mathrm{agg}}\). (b) scFv: off-rate against \(T_{\mathrm{collapse}}\), where the sample falls out of solution. Dashed lines are two of the constraints and shading marks designs that clear them. Open circles are aggregation rates at a fitting bound, and open triangles have $\mathrm{p}k_{\mathrm{off}} \geq 4.3$. The scFv parent is from round 1. Gray points fail a constraint not plotted (melt, high-molecular-weight species, purity or yield). Of the 26 scFv designs that pass, 7 have a measured $K_D$ and 19 have $\mathrm{p}k_{\mathrm{off}} \geq 4.3$. Labeled designs mark the ends of each frontier (Section 7).}\label{fig:gates}
\end{figure}

Both leads improve on their parents in affinity, thermal stability and aggregation, hold expression and purity, and pass the revised specifications (Table~\ref{tab:gates}, Figure~\ref{fig:gates}, Appendix H).
We take each molecule in turn, from its parent's curves through the test to the lead.

\textbf{Two processes on the VHH.} As the VHH parent heats, two separate things happen (Figure~\ref{fig:fit}b).
From about 40 °C the scattering climbs while the fluorescence barely changes, so particles are forming while nearly every molecule is still folded.
Only near 62 °C does the fluorescence shift, as the protein unfolds in a real, reproducible melt.
The melt is an equilibrium, so it depends only on the temperature the sample has reached.
The aggregation is kinetic, so the amount of protein aggregated depends on how long the sample has been warm and how much protein is present.

In the final round the onset sat near 40 °C regardless of the melt (median 40.2 °C, $\rho = -0.11$ with the melt, $n = 129$ wells), while molecules with the same onset differed more than tenfold in $\kappa_N$.
On a ramp, $T_{\mathrm{agg}}$ is a time disguised as a temperature, marking when the signal first leaves its baseline and not how fast it grows afterward.
It was a poor proxy for aggregation on this scaffold and could not rank designs.
We therefore used $\mathrm{p}k_{\mathrm{agg}}$ for aggregation and the identified melt for stability.

In the test, the identified melt moved about 0.1 °C at half the concentration (half load) and at the faster ramp, and $\mathrm{p}k_{\mathrm{agg}}$ moved 0.03 log units at half load, as predicted (Figure~\ref{fig:traces}c).
The onset rose at the faster ramp but moved the wrong way at half load (Figure~\ref{fig:traces}d), so a replicated load series is needed.

\textbf{The VHH lead} carries eight mutations from Alb8 (16 V-gene differences from the human germline IGHV3-23, against 13 for Alb8) and passes the original $K_D$, melt and yield constraints, with its $T_{\mathrm{agg}}$ right-censored above the ramp maximum of 85 °C (Appendix D.3).
Off-rate is 0.90 log slower than the parent's on two wells, and $K_D$ is 1.15 nM against 10.3 nM (Figure~\ref{fig:sensorgrams}a).
The identified melt is 70.6 °C against 62.2 °C, clearing the 70 °C constraint by about one replicate standard deviation.
$\mathrm{p}k_{\mathrm{agg}}$ is $+2.00$ on one identified well against $-0.39$ (Figure~\ref{fig:traces}a), and we detect no high-molecular-weight (HMW) species (i.e., soluble aggregates).

\textbf{One collapse on the scFv.} The scFv parent behaves differently (Figure~\ref{fig:fit}c).
Across designs, scattering begins at the foot of the first fluorescence transition, and a median 9.6 °C later the sample precipitates and leaves the light path.
The fluorescence rarely reaches an unfolded baseline, because the protein is gone before it gets there, and an equilibrium midpoint is resolved in only 37\% of wells.
There is one kinetic event, the sample's collapse out of solution, and reported $T_m^{\mathrm{app}}$ and $T_{\mathrm{agg}}$ are two summary statistics of it, about 10 °C apart in the final round.
This fits the known behavior of the format, in which domain separation and unfolding are coupled into one cooperative transition \citep{jager1999}, although we did not observe that separation directly.
Unlike on the VHH, these reported metrics could be predicted across rounds, because both track the collapse, which is a stable property of the sequence.
We replaced them anyway, since they report one event and not two properties, and used $T_{\mathrm{collapse}}$.

In the test, the collapse rose 2.0 °C at half load against 2.8 °C predicted and 4.7 °C at the faster ramp, above the 3 to 4 °C predicted (Figure~\ref{fig:traces}e).

\textbf{The scFv lead} carries six mutations from the 7B1-derived parent (7 V-gene differences from the human germlines IGKV3-20 and IGHV3-7, the same as the parent) and passes the original $K_D$, $T_m^{\mathrm{app}}$ and yield constraints.
Off-rate is 1.00 log slower on two wells, and $K_D$ is 0.63 nM against 3.7 nM (Figure~\ref{fig:sensorgrams}b; 0.96 log and 4.3 nM against the parent's DR0 run).
Collapse temperature is 74.2 °C against 56.3 °C (57.5 °C in DR0, Figure~\ref{fig:traces}b), and reported $T_{\mathrm{agg}}$ rose to 57.7 °C, short of the original 65 °C.
Two of the mutations are the packing repairs of Section 5 (Q91A, S245A), and G230Y carries the largest affinity gain.
The scFv is a notoriously unstable format \citep{wornpluckthun2001}, so an 18 °C gain is substantial.

Not every prediction held on either scaffold (Appendix E).
Paired conditions ran on different days, except the scFv load pair, and on the scFv 43 of 88 fast-ramp capillaries never collapsed in the window, so the ramp shifts are the weaker evidence.

\textbf{Revising the specification.} The goal stayed the same throughout, a stable molecule that does not aggregate, and what changed was how we measured it.
On the VHH, $\mathrm{p}k_{\mathrm{agg}} \geq 1.72$ replaced $T_{\mathrm{agg}}$ and the 70 °C melt constraint moved to $\hat{T}_m$, and on the scFv, $T_{\mathrm{collapse}} \geq 70$ °C replaced $T_m^{\mathrm{app}}$ and $T_{\mathrm{agg}}$, with limits on HMW species added on both (5\% and 3\%, Appendix E).
Against the original specifications in Table~\ref{tab:gates}, none of the 113 final-round scFv designs passes with a measured $K_D$, and two with $\mathrm{p}k_{\mathrm{off}} \geq 4.3$ would pass if counted, each on one well.
The metrics and predictions came before the perturbation plates were read, and we set the final specifications knowing the results (Appendix E).
The scFv threshold did not choose the lead, since the same 7 of the 113 final-round designs pass with a measured $K_D$ at either 65 or 70 °C, and 26 pass at 70 °C with those at $\mathrm{p}k_{\mathrm{off}} \geq 4.3$ counted.

\textbf{The frontier.} Each campaign ended with a set of trade-offs, and the leads are best read as points on that frontier (Figure~\ref{fig:gates}).
We nominated the designs measured cleanly on every axis, which were not always the designs that went furthest on one axis.
On the scFv, one design keeps close to the parent's affinity (3.3 nM against 3.7 nM) and collapses at 81.2 °C, 25 °C above the parent, on three wells.
At the other end, a design collapses at 77.0 °C with $\mathrm{p}k_{\mathrm{off}} \geq 4.3$, on a single well.
The lead (0.63 nM, 74.2 °C) sits just inside this frontier, as one of two designs with $K_D$ measured inside the constraint on two wells, and the more stable of the two, with the tighter replicates.
On the VHH, the lead has the slowest resolved off-rate of the round, the tightest design reads 0.85 nM on three wells (12 times tighter than the parent), one design reaches a 75.7 °C melt at 2.3 nM on one well, and another holds a 70.6 °C melt at 1.2 nM with $\mathrm{p}k_{\mathrm{agg}} \geq 3.0$.
The affinity end of the scFv frontier is a lower bound (the base design of Figure~\ref{fig:strategies}d reads $K_D \leq 0.25$ nM on nine wells, about 15 times tighter than the parent), so that campaign reached at least this far, and the long-window kinetics plate in Section 8 would resolve it.
Both populations improved, with binder rates, counting controls, of 48, 58, 93, 99\% on the VHH and 47, 44, 93\% on the scFv (Figure~\ref{fig:population}).

\section{8. Discussion}\label{discussion}

A general-purpose agent, working with a team of researchers, optimized two antibody leads without relying on a specialized design model, by reasoning about both the molecules and how they were being measured.
The agent proposed most of the mechanisms and analyses, while the research team set the overall design strategy and the trade-off between exploration and exploitation, and the researcher working with the agent reviewed designs and pressed it to read the raw curves and reason about the biophysics directly (Appendix G).
In the researcher's account the agent was often hasty in analysis, leaned on models where reasoning directly was available, and would not have stopped to look at the curves without prompting.
Once prompted, however, it read every curve on both scaffolds, built and compared candidate models, and registered its predictions within days, work that would have taken the human researcher weeks.

This work points to a tension at the center of autonomous science.
Optimization needs a fixed target, but science needs the freedom to change the metrics of success.
Machine learning for molecular design usually takes the specification as an input and concentrates on search, while these campaigns treated it as a working hypothesis about what each assay measures, to be tested as a campaign runs.
Here that meant asking which intrinsic property of the molecule each thermal metric was meant to estimate, and checking the answer against the raw curves.
A specification that can change can also drift toward whatever the data favor, so each change needs a record of what changed and why, predictions written down in advance, and a test that could fail.
The journal and the perturbation plates served that purpose here, and any agent revising its own objective will likely need the same.

Questioning a metric during a campaign depends on the lab as much as on the agent.
Three properties of BigHat's platform made it possible: raw instrument traces stored beside every derived metric and quality-control flag, a single system of record an agent could query directly, and protocols that could be changed between rounds.
As more experimental platforms open their raw data and protocols to agents, we expect more assay development to move inside the campaign, and we believe that prospect justifies the work of rebuilding experimental data infrastructure around AI agents.

\textbf{Limits and next steps.} The closest affinity competitors to the scFv lead sit at or beyond the BLI assay's limit of resolution, as do 51 of the 113 final-round scFv designs, so a long-window kinetics plate with the shortlists and parents, or a more sensitive platform such as surface plasmon resonance, is the next step for ranking them.
Lead values rest on one or two replicates, and the VHH's aggregation value is one well, so confirmation runs of both leads come first.
A melt that holds across concentrations does not by itself rule out aggregation \citep{goyal2014}, and an isothermal hold with size exclusion before and after would test it directly.
Stress testing and expression in a mammalian system would extend these results toward development, and a condition that resolves the scFv's unfolding midpoint would complete its thermal picture.
Size analysis of scFv samples held just below the collapse could also show whether variable domains pair across molecules on the way to aggregation.
Testing a small set of designs spanning the range of $\mathrm{p}k_{\mathrm{agg}}$ at high protein concentration would show whether the metric can serve as a low-material screen for high-concentration behavior.

Looking forward, two views of AI in molecular design now sit side by side.
In one, a general-purpose model orchestrates specialized tools that do the design, and in the other, taken in these campaigns, the general model reasons about the molecule directly.
We see the choice between them as a question of resources more than capability.
Scaling a general model, extending a specialized one and generating data in the lab are all ways of turning compute and skilled labor into better molecules, and the right mix depends on what a team has and the value of the application.
Specialized models keep their own advantage, since they are modular and can be tuned to a particular data distribution far faster than a frontier model can change, and for a valuable enough problem that speed should justify the investment.
Agents also shift this balance by supplying the labor custom models require, so models with likelihoods crafted for a specific assay and dataset once again become practical, as the censored regressions and physical models in these campaigns suggest.
In a world with abundant intelligence, we can afford to investigate nature more deeply than ever before.

\section*{Contributions}

Samuel Stanton led the project and directed the agent.
Claude, working in Claude Code, carried out most of the design and analysis (Appendix G).
BigHat Biosciences provided the platform, proposed the lead molecules and initial molecule specifications, ran the experiments, advised on the assays and shared in the project's scientific decisions.

\section*{Disclosure}

Anthropic paid BigHat for carrying out the laboratory work described here.

\section*{Acknowledgments}

We thank Liz Schwarzbach and Sangeetha Iyer for helpful discussion throughout the project, and Anna Barr-Folinsky, Bessie Wang, Nathan Frey, and Eric Kauderer-Abrams for their support.

\newpage

\section{How to read these appendices}\label{appendix-guide}

Each appendix backs one part of the main text, and the appendices can be read in any order.

\begin{longtable}[]{@{}llr@{}}
\toprule\noalign{}
Appendix & What it holds & Page \\
\midrule\noalign{}
\endhead
\bottomrule\noalign{}
\endlastfoot
\hyperref[appendix-a.-platform-and-protocol]{A} & Parents, platform and assay methods & \pageref{appendix-a.-platform-and-protocol} \\
\hyperref[appendix-b.-the-record-and-its-provenance]{B} & The campaign journal and what it can and cannot show & \pageref{appendix-b.-the-record-and-its-provenance} \\
\hyperref[appendix-c.-library-design-and-the-fiber-method]{C} & Library design, the lattice method and the Walsh regression (Section 5) & \pageref{appendix-c.-library-design-and-the-fiber-method} \\
\hyperref[appendix-d.-reading-the-raw-curves]{D} & Reading the raw curves, and the replacement models (Section 6) & \pageref{appendix-d.-reading-the-raw-curves} \\
\hyperref[appendix-e.-the-designed-perturbation-test]{E} & The designed perturbation test and its scoring (Sections 6 and 7) & \pageref{appendix-e.-the-designed-perturbation-test} \\
\hyperref[appendix-f.-specialized-models-in-full]{F} & Every specialized model, scored prospectively (Section 8) & \pageref{appendix-f.-specialized-models-in-full} \\
\hyperref[appendix-g.-division-of-labor-from-the-journal]{G} & Division of labor, from the journal (Section 8) & \pageref{appendix-g.-division-of-labor-from-the-journal} \\
\hyperref[appendix-h.-outcomes-in-full]{H} & Both leads against both sets of specifications (Section 7) & \pageref{appendix-h.-outcomes-in-full} \\
\hyperref[appendix-i.-limits-in-full]{I} & Limits and next experiments (Section 8) & \pageref{appendix-i.-limits-in-full} \\
\end{longtable}

\textbf{Glossary.}
The appendices use the campaigns' working vocabulary, which we define here once.

\begin{longtable}[]{@{}
  >{\raggedright\arraybackslash}p{(\linewidth - 2\tabcolsep) * \real{0.24}}
  >{\raggedright\arraybackslash}p{(\linewidth - 2\tabcolsep) * \real{0.76}}@{}}
\toprule\noalign{}
Term & Meaning \\
\midrule\noalign{}
\endhead
\bottomrule\noalign{}
\endlastfoot
DR0, DR1, \ldots & Design round 0, 1, and so on. The VHH ran DR0--DR3 and the scFv DR0--DR2. \\
Well, design & A well is one measured sample. A design is one sequence, which may occupy several wells. \\
Block & A group of wells on a plate built to answer one question (Appendix C.2). \\
Anchor & A known molecule placed on a plate as a reference. \\
Lever & A mutation with a measured effect on a target property. \\
Boolean lattice & The set of all combinations of a set of mutations, equivalent to a combinatorial library, usually sampled (Appendix C.3). \\
Cube, factorial & Every combination of a set of mutations, measured. \\
Chassis & The VHH design from DR2 on which the final round was built. \\
Hero block & The chassis plus single mutations, tested in the final VHH round. \\
P & The scFv design combining Q91A, G230Y, Y238A and S245A, on which the final scFv round was built. \\
$\mathrm{p}k_{\mathrm{off}}$ & $-\log_{10}$ of the off-rate, so larger is slower dissociation and tighter binding. \\
Assay floor & The slowest off-rate the measurement window can resolve. Slower values are recorded as bounds (Appendix I.1). \\
$T_m^{\mathrm{app}}$, $T_{\mathrm{agg}}$ & The apparent melting temperature and aggregation onset that BigHat's pipeline reports from each nanoDSF run (Appendix A.1). Figures label them reported Tm and Tagg. \\
$T_m$, $\hat{T}_m$ & The thermodynamic melting temperature, and our fitted estimate of it (the identified melt). \\
$K_D$ & The equilibrium dissociation constant, lower being tighter binding. It equals $k_{\mathrm{off}}/k_{\mathrm{on}}$, the off-rate over the on-rate. \\
HMW & High-molecular-weight fraction by size exclusion, a measure of aggregate in the sample. \\
Melt & A true unfolding midpoint. \\
Collapse & The temperature at which the sample leaves solution, read from the scattering peak. \\
$k_{\mathrm{agg}}$ & $\kappa_N / c$, the native-state aggregation rate per unit of loaded protein, reported as $\mathrm{p}k_{\mathrm{agg}} = -\log_{10} k_{\mathrm{agg}}$. Higher is slower. \\
$T_{\mathrm{collapse}}$ & The temperature of the scattering peak, standardized to a reference loading (Appendix D.5). \\
Neat, half load & The standard loading and one-half of it. \\
Standard condition & Neat load on the 0.3 °C/min ramp. \\
Identified & A well whose fitted rate or midpoint has a finite interval away from its bounds (Appendix D.2). \\
Censored & A value known only as a bound, such as an off-rate beyond the floor. \\
\end{longtable}

\section{Appendix A. Platform and protocol}\label{appendix-a.-platform-and-protocol}

\textbf{A.1 Parents, targets, and what the agent started with.} The VHH campaign began from
an anti-human-serum-albumin single-domain antibody of 115 residues, and the scFv campaign from an anti-mesothelin single-chain fragment of 259 residues, both supplied by
BigHat Biosciences as a parent sequence and a success specification. Baseline measurements of each
parent came with the specification; no other assay results or internal models were available. Public
resources were the patent and journal literature, which on day one identified the VHH as Ablynx's
Alb8 and the scFv as a derivative of Medarex clone 7B1; the Protein Data Bank; and open-source
structure, language, and design models. The specifications were constraints on affinity, thermal
stability, and expression yield, set at $K_D$ ≤ 1.6 nM, $T_m^{\mathrm{app}}$ ≥ 70 °C, $T_{\mathrm{agg}}$ ≥ 70 °C, and yield
\textgreater{} 10 µg/mL for the VHH, and $K_D$ ≤ 1.5 nM, $T_m^{\mathrm{app}}$ ≥ 65 °C, $T_{\mathrm{agg}}$ ≥ 65 °C, and yield \textgreater{} 10 µg/mL for
the scFv. $T_m^{\mathrm{app}}$ is the reported melting temperature, read from the unfolding transition seen in the
protein's intrinsic fluorescence as the sample is heated; $T_{\mathrm{agg}}$ is the aggregation onset, the
temperature at which static light scattering from the same capillary begins to rise. Both are the
values BigHat's analysis pipeline reports from each nanoDSF run.

\textbf{A.2 Platform.} Each design round was 176 design wells (two 96-well plates) plus BigHat's 16
end-to-end controls, which in most rounds included two parent wells and two non-binding-control wells per plate (we found no scFv parent wells in the DR2 assay data).
Variants were expressed cell-free in \emph{E. coli} extract, affinity-captured, quantified, and
measured by bio-layer interferometry at a single 100 nM analyte concentration with a 1:1 kinetic
fit, by nanoDSF on an UNcle instrument (intrinsic fluorescence and static light scattering on a 0.3
°C/min ramp), by size-exclusion chromatography at day 0, and by non-reduced capillary
electrophoresis. The VHH ran four rounds (DR0--DR3) and the scFv three (DR0--DR2) between 2026-06-29
and 2026-09-04, 1,232 design wells in all. Turnaround from design submission to data was 9--14 days (median 10)
for the standard assay set, set mainly by gene synthesis at Twist.

\textbf{A.3 Assay methods.} BigHat ran every assay, and the protocols below are theirs.

\textbf{Binding.} BLI measurements were made on Sartorius Octet instruments. For the VHH molecules, target protein was biotinylated in-house, checked by QC, and captured on streptavidin tips, and select anti-albumin binders carried a GS-6His tag for antibody capture. The anti-mesothelin scFv molecules were measured against a commercial His-tagged antigen. Assays used tip-loading thresholds, positive and end-to-end controls, and buffer-blank referencing, with double referencing in some runs. Fitting parameters were fixed in the method protocols at project start. Kinetics data were uploaded to Reccy and passed through BigHat's QC pipelines, and protocol deviations were flagged in the data.

\textbf{Yield.} Protein yield was assessed by measuring absorbance at 260nm and 280nm using a SpectraMax iD5 spectrophotometer. Sequence-derived molar extinction coefficients and volume-defined path lengths were then combined to determine protein concentration using the Beer-Lambert Law.

\textbf{Assembly.} Capillary electrophoresis was performed to assess protein assembly on the Revvity LabChip GX II Touch microfluidic capillary electrophoresis SDS system. Samples were loaded into a HT Protein Express LabChip according to the manufacturer’s protocol. A manufacturer-provided protein ladder was used to benchmark protein size and validate protein assembly.

\textbf{Thermostability.} Thermostability assessment was performed using nanoDSF on the UNchained Labs UNcle system from 25 °C to 85 °C using a thermal ramp rate of 0.3 °C/min. To determine melting temperature ($T_m^{\mathrm{app}}$), the instrument measures intrinsic fluorescence across a spectrum of wavelengths following 280nm excitation. The inflection point of the barycentric mean (BCM) wavelength versus temperature is identified as the $T_m^{\mathrm{app}}$, with a secondary derivative QC check to evaluate noise. Simultaneously, the instrument also measures static light scattering intensity (arbitrary units) at 266nm to assess protein aggregation. $T_{\mathrm{agg}}$ is the scattering onset called by a baseline-departure algorithm on the 266 nm signal, with additional QC to confirm the onset and exclude calls driven by signal noise. We did not obtain the software's exact rule. Among QC-passing scFv wells, a crossing at about 6\% of the rise from baseline to peak reproduces the reported values to a median of 0.5 °C, and no simple rule we tried fits the VHH wells well. UV-266 attenuation varied between instruments and plate halves (0.50, 0.25 or 0.13), so absolute scattering intensities are not comparable across runs. The load and ramp comparisons of Appendix E were each made within a single instrument.

\textbf{Aggregation and purity.} Aggregation and purity were determined using size exclusion chromatography (SEC) high performance liquid chromatography (HPLC) on an Agilent 1260 Infinity II HPLC system. Monomers and aggregates were evaluated on a 125Å ethylene bridged hybrid (BEH) HPLC column manufactured by Waters. A manufacturer-generated protein ladder was used to benchmark protein size.

\textbf{A.4 Division of labor.} The agent was Claude, running in Claude Code, primarily with Claude Fable 5, directed by the corresponding author (S.S.), with BigHat as the experimental partner. The agent analyzed the data,
designed the library components, and recommended every decision; S.S. gave feedback on the
experiment design, pushed at times for more or less exploration or for more analysis, set or
accepted the objective, agreed the protocol with BigHat, reviewed and edited all partner-facing
communications, and, with BigHat, made the final decisions on the experimental design. In S.S.'s
judgment the human and AI contributions were roughly even and the result would have been substantially worse with
either party removed. See Section 8 for more details.

\section{Appendix B. The record and its provenance}\label{appendix-b.-the-record-and-its-provenance}

\textbf{B.1 The record.} Every design, analysis, finding, and decision was appended as a typed record to a
campaign journal as it happened, with an author field naming the agent process or the researcher; the two
journals hold 387 records through the lead decisions of 2026-09-04 and 443 in all. Predictions for the later design rounds were written to file before the plates
ran and scored against the wet data when it arrived. After each campaign's last round an adversarial
review recomputed the headline results from raw data and code, and its findings supersede earlier
records where they disagree. Three caveats on what is cited. The order-1 main effects of the VHH DR3
factorial are one author's least-squares fit reproduced by two checkers and not a pipeline table.
One cross-scaffold learnability record that we cite could not be located in the journal. And the
journal for the VHH's second round is short (20 records in the saved file) and
may be incomplete.

\section{Appendix C. Library design and the lattice
method}\label{appendix-c.-library-design-and-the-fiber-method}

\setcounter{figure}{0}
\renewcommand{\thefigure}{C\arabic{figure}}

\textbf{C.1 Lineage.} The agent identified both parents on the first day. The VHH matched Ablynx's
Alb8 at all 115 positions, which brought with it a patent family, a published successor set of six
framework and CDR1 substitutions (Alb23; \citealp{ablynx2012}), and a crystal structure of a family member's binding mode
(PDB 8OI2; \citealp{defelice2023}). The scFv was traced through a patent search to
Medarex clone 7B1, from which it differs at three CDR-H3 positions,
four VH framework positions and one VL framework position, with no structure available.
Positions are numbered sequentially along each expressed construct (the scFv runs VL, linker, VH).
CDRs follow the Kabat definition, except that VHH CDR1 spans residues 26--35 to include the Chothia loop.
The scFv's human germlines were assigned with ANARCI \citep{dunbar2016}, and IGHV3-23 is the human reference for the VHH.
Both parents came from lineages whose
affinity had been engineered once already, and both bound more weakly than the best members of those
lineages; the agent treated the lineage as the first library: on the VHH the six Alb23 substitutions
and six other family members (Alb1, Alb3--6 and Alb10; \citealp{ablynx2006}) went on the round-0 plate as single and combined variants; on the scFv
the three CDR-H3 differences from 7B1 became a five-well ancestral-reversion block. Every well in
the VHH lineage blocks bound, the best of those six read 1.42 nM in round 0 against a parent in
the 10 nM range, and the largest single affinity lever of the campaign, E1A, came from
Alb1. All five ancestral-reversion wells on the scFv bound, three of them beyond the assay's
dissociation floor, and one reversion, G230Y, took the parent beyond that floor in every subsequent
round. The harder problem was thermal: zero of 176 scFv round-0 designs reached the reported $T_{\mathrm{agg}}$
constraint of 65 °C, and on the VHH the two thermal constraints at 70 °C sat about 8 °C above the parent's
melting temperature and 25--30 °C above its aggregation onset. An 8 °C gain in melting temperature
is a large ask for a 115-residue domain; a 25--30 °C gain in aggregation onset is larger than any
gain we found published for this scaffold, and the size of that gap was the first sign that the $T_{\mathrm{agg}}$
metric was not reporting the fold (Appendix D).

\begin{figure}[t]
\centering
\includegraphics[width=0.62\linewidth,height=\textheight,keepaspectratio,alt={The parent's binding mode, from the deposited Alb8:albumin crystal structure (PDB 8Z8V). (a) The VHH (pink; CDRs dark red) bound to albumin domain II (gold; domains I and III gray). (b) The interface, with albumin contact side chains in gold and VHH contact side chains in dark red; the eight positions mutated in the lead are drawn as spheres. Four sit at the interface (E1A, S31R and S103R within 3--4 Å of albumin; S30R at 5 Å) and four are on the far side of the domain (L45P 9 Å, T77N 12 Å, V93R 15 Å, N16G 25 Å): the mutations that raised thermal stability and reduced aggregation act on the scaffold, away from the paratope (i.e., the binding surface). On the first day the agent used the structure of the related Alb1 complex (PDB 8OI2) as a map for what could be changed without touching the binding mode; the figure is drawn from the complex of the parent itself.}]{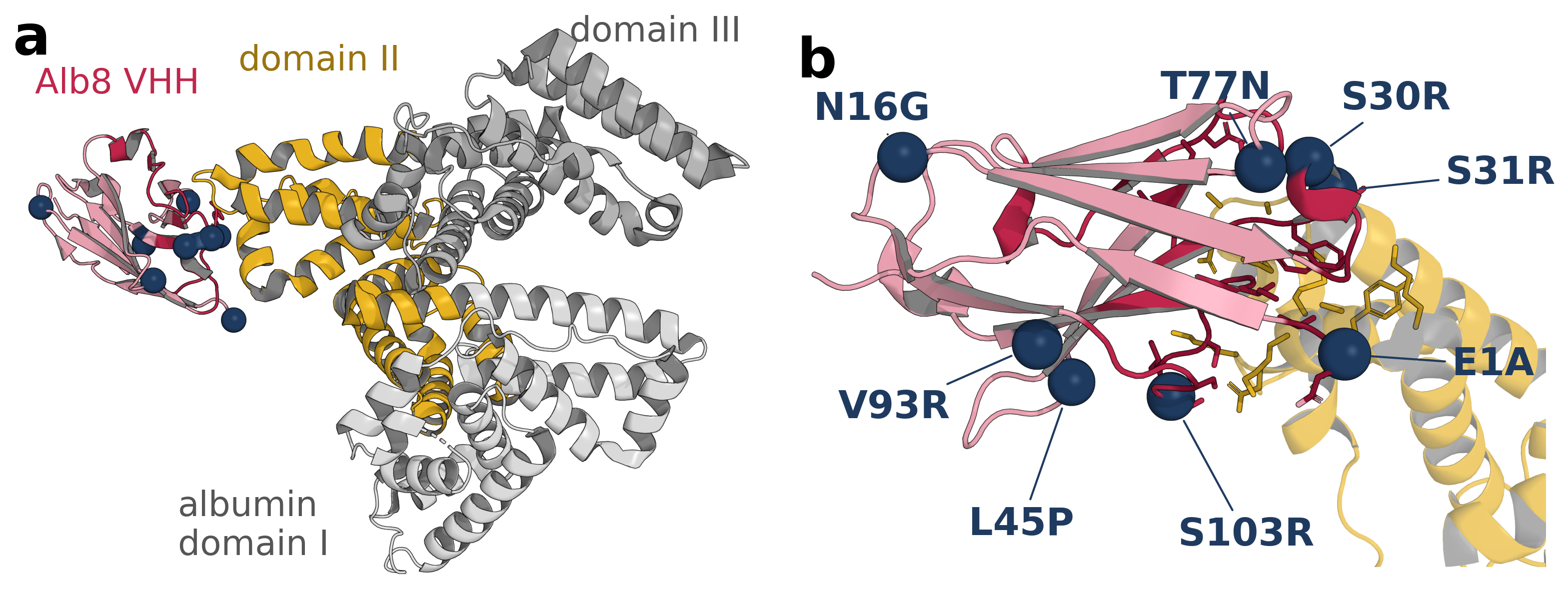}
\caption{\textbf{The parent's binding mode, from the deposited Alb8:albumin crystal structure (PDB
8Z8V; \citealp{mima2024}).} (a) The VHH (pink; CDRs dark red) bound to albumin domain
II (gold; domains I and III gray). (b) The interface, with albumin contact side chains in gold and
VHH contact side chains in dark red; the eight positions mutated in the lead are drawn as spheres.
Four sit at the interface (E1A, S31R and S103R within 3--4 Å of albumin; S30R at 5 Å) and four are
on the far side of the domain from the binding surface (L45P 9 Å, T77N 12 Å, V93R 15 Å, N16G 25 Å). On the first day the agent used the structure of
the related Alb1 complex (PDB 8OI2; \citealp{defelice2023}) as a map for what could be
changed without touching the binding mode; the figure is drawn from the complex of the parent
itself.}\label{fig:structure}
\end{figure}

\begin{figure}[t]
\centering
\includegraphics[width=1\linewidth,height=\textheight,keepaspectratio,alt={How the plates were built. (a) A Boolean lattice over four mutations is the 2⁴ cube of their presence-absence combinations; a Sobol sample of the cube (filled) picks cells so that every mutation, and every pair, is present in about half the sampled cells, which keeps the main effects of C.3 well conditioned at a fraction of the cube, with higher orders partly aliased. (b, c) Design wells per round by block family, from the design tables.}]{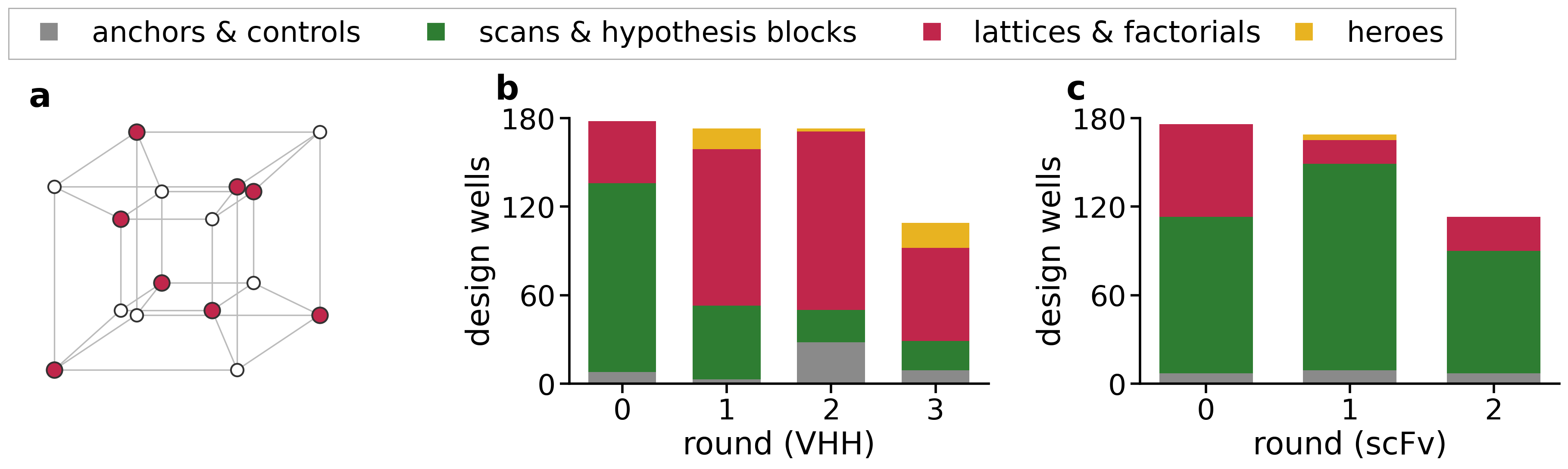}
\caption{\textbf{How the plates were built.} (a) A Boolean lattice over four mutations is the 2⁴ cube of their
presence-absence combinations; a Sobol sample of the cube (filled) picks cells so that every
mutation, and every pair, is present in about half the sampled cells, which keeps the
main effects of C.3 well conditioned at a fraction of the cube, with higher orders partly aliased. (b, c) Design wells per round by block
family, from the design tables.}\label{fig:c1}
\end{figure}

\textbf{C.2 Blocks.} Each plate was assembled from four kinds of block (Figure~\ref{fig:c1}). \emph{Anchors}
are known molecules, lineage members, literature variants and the parent, that serve as centroids
around which to explore, and as the reference wells every other measurement on the plate is compared
against. \emph{Scans} change one position at a time over a region, so that every well is a main
effect on the parent background; the round-0 alanine scans of the scFv CDRs found two packing
repairs, Q91A in CDR-L3 and S245A at the VH J-stem, each worth 6--7 °C on the collapse temperature as a fitted module effect (single scan wells read 5.3--7.7 °C),
and a tyrosine-to-phenylalanine scan established that the paratope needed an aromatic ring but not a
hydroxyl at certain positions (8 of 8 binders against 1 of 8 for tyrosine-to-serine).
\emph{Hypothesis blocks} test a stated hypothesis in a few wells: a two-well control that combined
the scFv's affinity reversions with its two packing repairs, later called P, beat every
hypothesis-designed variant in round 1 and became the putative lead. \emph{Lattice samples and factorials}
sample the combinatorial space over a set of mutations and are read by the regression of C.3. Round
0 of each campaign was mostly scans; the VHH's rounds 1--3 were mostly lattice samples and factorials once
the levers were known; the scFv's rounds 1--2 stayed hypothesis-driven because its levers carried
affinity costs that had to be tested one at a time.

\textbf{C.3 Boolean lattices and the Walsh regression.} A Boolean lattice over \(k\) mutations is the set
\(\{0,1\}^k\) of their presence--absence combinations, from a wild type (\(\mathbf{0}\)) to the full
mutant (\(\mathbf{1}\)). A full factorial measures every cell; a lattice sample measures \(n \ll 2^k\) of
them, chosen by a scrambled Sobol sequence \citep{sobol1967, owen1998} in \([0,1]^k\)
thresholded at one half, a trick borrowed from Bayesian optimization in continuous domains. The
low-discrepancy sample holds each mutation near 50\% occupancy and keeps the design matrix well
conditioned (condition number 110.8 for the VHH round-0 lattice sample of 13 mutations in 37 wells), which is
what allows the model below to be fit at sub-plate budgets.

Lattice samples are read in the Walsh parity basis, the Fourier basis of Boolean functions in which the
epistasis of a sequence--function map is a sparse set of coefficients of definite order
\citep{poelwijk2016, tsui2025}. Code each mutation as
\(s_j = 2x_j - 1 \in \{-1, +1\}\) and define, for every subset \(S\) of the mutations, the parity
function \(\phi_S(\mathbf{x}) = \prod_{j \in S} s_j\). These functions are orthogonal over the full
cube, and a model of order \(K\) uses all subsets with \(1 \le |S| \le K\):
\[ y(\mathbf{x}) \;=\; \bar y \;+\; \sum_{1 \le |S| \le K} w_S\, \phi_S(\mathbf{x}) \;+\; \varepsilon, \qquad \varepsilon \sim \mathcal{N}(0, \beta^{-1}). \]
Order 0 is the mean alone, order 1 is additive (main effects only), order 2 adds pairwise
interactions. In this coding the effect of adding mutation \(j\), averaged over backgrounds, is
\(2 w_{\{j\}}\), and the interaction contrast \(y_{11} - y_{10} - y_{01} + y_{00}\) for a pair is
\(4 w_{\{i,j\}}\); the figures below report \(2w\) for main effects and the raw coefficient for
pairs.

These coefficients are also attributions in the game-theoretic sense.
Treat the mutations as players and \(y\) as the value of a coalition.
The averaged effect \(2 w_{\{j\}}\) is the Banzhaf value of mutation \(j\).
Its Shapley value \citep{shapley1953}, the basis of SHAP explanations of machine-learning models \citep{lundberg2017, tsui2025}, is \(2 \sum_{S \ni j,\, |S|\, \mathrm{odd}} w_S / |S|\).
Even-order terms cancel, so for a model of order 1 or 2 the two coincide, and the main effects we report are the Shapley attribution of the change from the parent to the full combination, each with a posterior uncertainty.

The weights carry a Gaussian prior \(\mathbf{w} \sim \mathcal{N}(\mathbf{0}, \alpha^{-1} I)\). With
\(\Phi\) the \(n \times M\) design matrix of parity columns and
\(\mathbf{y}_c = \mathbf{y} - \bar y\), the posterior is Gaussian with
\[ A = \alpha I + \beta\, \Phi^{\mathsf T} \Phi, \qquad \mathbf{m} = \beta A^{-1} \Phi^{\mathsf T} \mathbf{y}_c, \qquad \mathrm{Cov}[\mathbf{w}] = A^{-1}, \]
so every effect comes with a posterior standard deviation. The two precisions are set by type-II
maximum likelihood in MacKay's evidence framework \citep{mackay1992}: iterate
\(\gamma = \sum_i \beta \lambda_i / (\alpha + \beta \lambda_i)\) over the eigenvalues \(\lambda_i\)
of \(\Phi^{\mathsf T}\Phi\), \(\alpha \leftarrow \gamma / \mathbf{m}^{\mathsf T}\mathbf{m}\),
\(\beta \leftarrow (n - \gamma) / \lVert \mathbf{y}_c - \Phi \mathbf{m} \rVert^2\), where \(\gamma\)
is the effective number of well-determined parameters. The log marginal likelihood at the fitted
precisions is
\[ \ln p(\mathbf{y} \mid K) \;=\; \tfrac{M}{2} \ln \alpha + \tfrac{n}{2} \ln \beta - \tfrac{\beta}{2} \lVert \mathbf{y}_c - \Phi \mathbf{m} \rVert^2 - \tfrac{\alpha}{2}\, \mathbf{m}^{\mathsf T}\mathbf{m} - \tfrac{1}{2} \ln |A| - \tfrac{n}{2} \ln 2\pi, \]
and the model order is chosen by maximizing it over \(K \in \{0, 1, 2\}\) (3 where the sample
supports it). The \(\tfrac{1}{2}\ln|A|\) term penalizes added columns, so the selected order reflects the epistasis the data support. Where a target is censored (an off-rate beyond the dissociation floor, an aggregation rate
at its bound), the fit is completed by expectation--maximization in the Tobit form \citep{tobin1958}: each censored
value \(y_i \ge c_i\) is replaced by its conditional expectation under the current fit,
\(\mu_i + \sigma\, \varphi(z_i) / (1 - \Phi_{\mathcal N}(z_i))\) with \(z_i = (c_i - \mu_i)/\sigma\)
and \(\sigma^2 = \beta^{-1}\), and the regression is refit until the weights stop moving. For blocks
that are not reversion cubes the design uses centered 0/1 indicators in place of parity columns; the
centering matters, since an uncentered indicator pins the reference design at \(\bar y\) and biases
every main effect.

\begin{figure}[t]
\centering
\includegraphics[width=1\linewidth,height=\textheight,keepaspectratio,alt={The scFv round-0 germline-reversion lattice. (a) Binders among expressed designs with each position reverted to germline or kept somatic; seven of eight reversions look costly. (b) Walsh main effects on the probability of binding with ± 2σ posterior intervals: three positions carry binding and five are free. (c) Log evidence by model order relative to order 0, for binding (n = 58), reported Tagg (n = 58) and off-rate among binders (n = 14); order 1 is selected on all three.}]{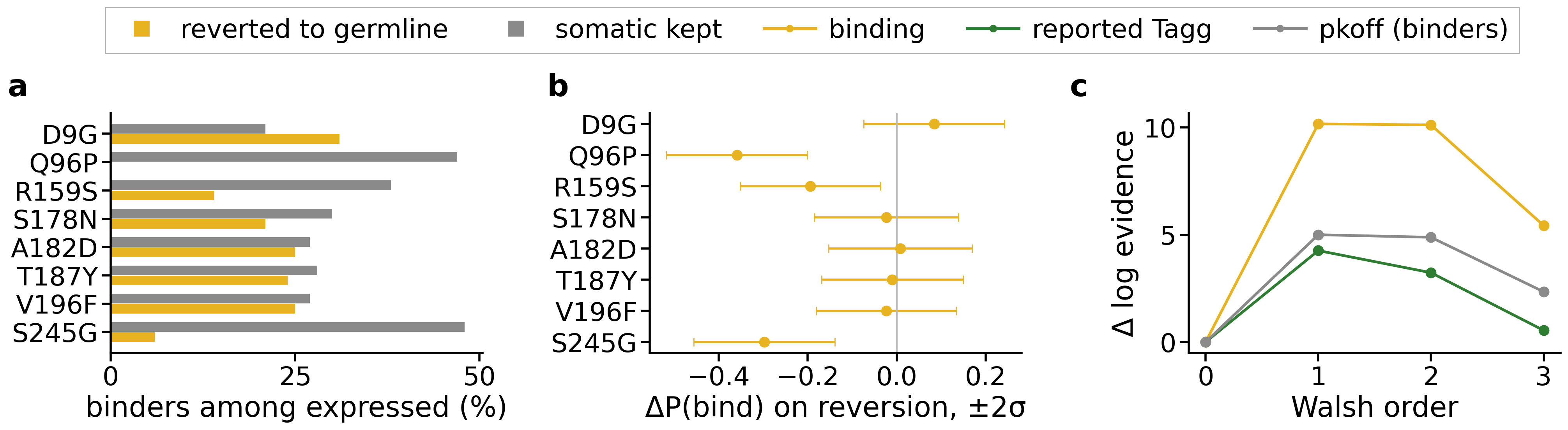}
\caption{\textbf{The scFv round-0 germline-reversion lattice.} (a) Binders among expressed designs
with each position reverted to germline or kept somatic; seven of eight reversions look costly. (b) Walsh
main effects on the probability of binding with ± 2σ posterior intervals: three positions carry
binding and five are free. (c) Log evidence by model order relative to order 0, for binding (n =
58), reported $T_{\mathrm{agg}}$ (n = 58) and off-rate among binders (n = 14); order 1 is selected on all
three.}\label{fig:c2}
\end{figure}

\textbf{C.4 The scFv round-0 lattice, read two ways.} The first lattice sample the campaign ran held 63 designs, 56 from a Sobol sequence and 7 added by hand, over the eight somatic mutations that separate the scFv from its germline
(Figure~\ref{fig:c2}). Read by marginal binding rates, reverting seven of the eight positions lowered the fraction of binders (D9G is the exception),
and the natural conclusion was that nearly every somatic mutation was necessary. Read in the Walsh basis,
the evidence selects order 1 for binding, reported $T_{\mathrm{agg}}$ and off-rate alike, and only three positions
carry binding (ΔP(bind) −0.36, −0.30 and −0.19, each ± 0.08); the other five are free on binding,
and four of them, S178N, A182D, T187Y and V196F, are candidate thermal levers, each estimated at +0.8 to +1.6 °C of
reported $T_{\mathrm{agg}}$. The marginal differences at those five positions are consistent with noise. Those four positions became the stabilizers of the DR2 dose cube (C.6), and one of the four residues, T187Y, is in the final lead, beside V196A, a milder form of a second. In round 1, among designs with at least three of the five stabilizing reversions (D9G, S178N, A182D, T187Y, V196F) and otherwise only the modules of that round (D190E, S191T, Q91A, S245A, G230Y, Y238A), 0 of 7 bound without G230Y and Y238A (five non-binders and two ambiguous wells), 3 of 4 bound with one of the pair, and 14 of 15 bound with both.

\begin{figure}[t]
\centering
\includegraphics[width=1\linewidth,height=\textheight,keepaspectratio,alt={The VHH DR3 2\^{}6 cube (recomputed). Main effects of adding each lever to the parent, ± 1σ, on (a) off-rate (n = 63 quantified cells), (b) consensus melting temperature (n = 52), (c) aggregation rate (n = 57 unbounded cells); (d) log evidence by model order relative to order 0, order 1 selected on all three.}]{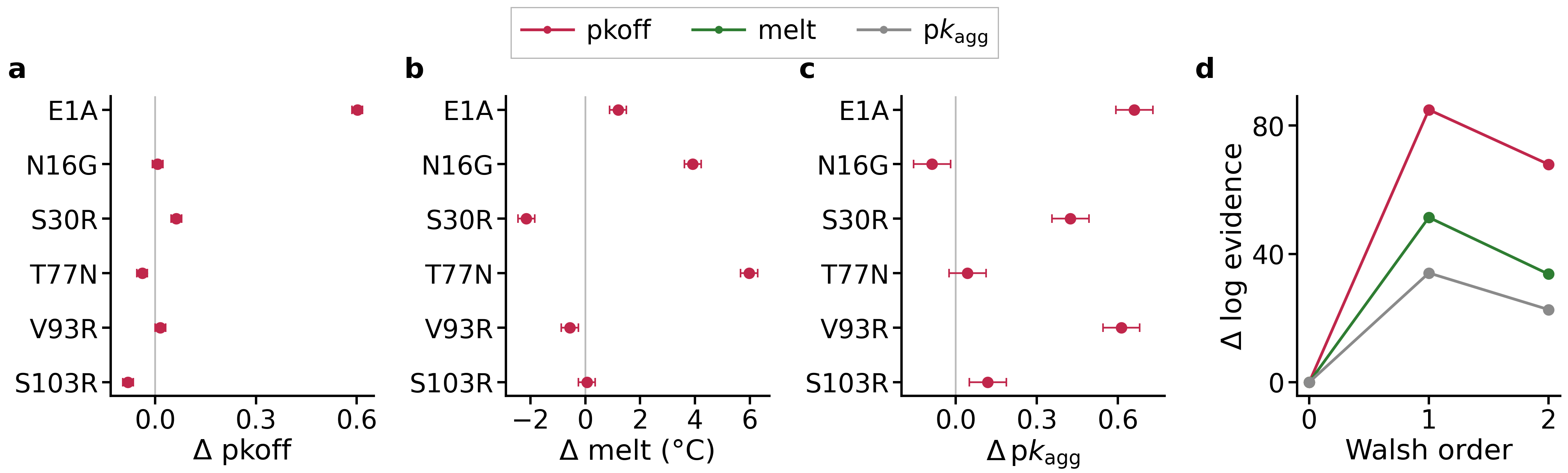}
\caption{\textbf{The VHH DR3 \(2^6\) cube (recomputed).} Main effects of adding each lever to the
parent, ± 1σ, on (a) off-rate (n = 63 quantified cells), (b) consensus melting temperature (n = 52),
(c) aggregation metric \(\mathrm{p}k_{\mathrm{agg}}\) (n = 57 unbounded cells); (d) log evidence by model order relative to
order 0, order 1 selected on all three.}\label{fig:c3}
\end{figure}

\textbf{C.5 The VHH DR3 cube.} The VHH's final round put a full \(2^6\) factorial on the six levers
found in DR0--DR2 (E1A, N16G, S30R, T77N, V93R, S103R) on the parent background, one well per cell,
64 cells with the parent (Figure~\ref{fig:c3}). The evidence selects order 1 on all three objectives: additive
to within the replicate noise, with one affinity lever (E1A, +0.60 $\mathrm{p}k_{\mathrm{off}}$), two melt levers (N16G
+3.9 °C, T77N +6.0 °C) and three aggregation levers (E1A, S30R, V93R, each +0.4 to +0.7 on \(\mathrm{p}k_{\mathrm{agg}}\)).
S30R costs about 2 °C of melt and S103R about 0.08 of $\mathrm{p}k_{\mathrm{off}}$, the two trade-offs the lead accepts.
The recomputation here is on the design-level values, with the consensus melting temperature in
place of the identified two-state midpoint the campaign used; it reproduces the campaign's signs and ranks. Pair terms on the
melt are resolved only to 0.7--2.6 °C, and two reach \textbar z\textbar{} \textgreater{} 2 on the
replicate error (E1A × T77N −1.3 °C, S30R × T77N +1.2 °C) without exceeding 2 °C; the selected order
is 1 on every target.

\begin{figure}[t]
\centering
\includegraphics[width=1\linewidth,height=\textheight,keepaspectratio,alt={The scFv DR2 dose cube on P. Walsh coefficients ± 1σ for the five main effects and ten pair terms on (a) off-rate, order 2 selected, log₁₀ BF(2 vs 1) = +17.9; (b) T\_\{\textbackslash mathrm\{collapse\}\}, order 1 selected; (c) reported Tagg, order 2 by +0.2. Coefficients are in the ± 1 parity coding (a main effect of adding the mutation is twice the coefficient).}]{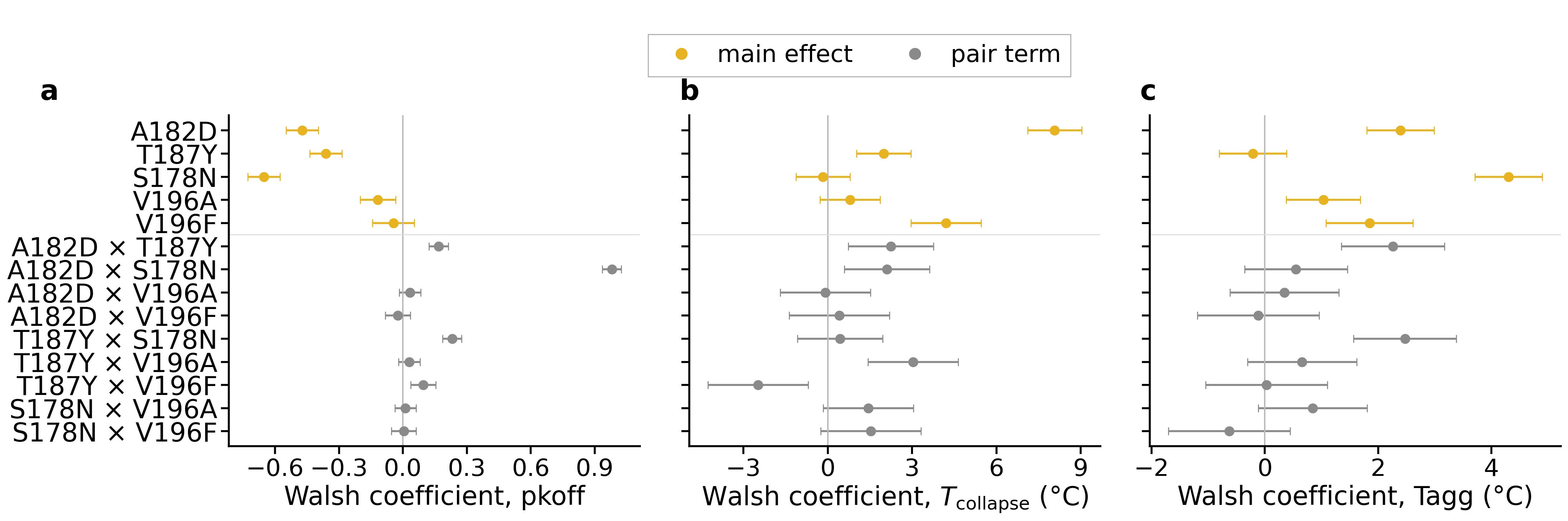}
\caption{\textbf{The scFv DR2 dose cube on P.} Coefficients ± 1σ for the five main effects and
nine pair terms (V196A and V196F cannot occur together) on (a) off-rate, order 2 selected, log₁₀ BF(2 vs 1) = +17.9; (b)
\(T_{\mathrm{collapse}}\), order 1 selected; (c) reported $T_{\mathrm{agg}}$, order 2 by +0.2. Coefficients are on 0/1 indicators, so a main effect is the effect of adding the mutation to P.}\label{fig:c4}
\end{figure}

\textbf{C.6 The scFv DR2 dose cube.} The one real non-additivity either campaign found came from a
\(2^4\) cube on P over the four stabilizers of C.4 plus V196A as a milder alternative to V196F
(Figure~\ref{fig:c4}). This cube is not a reversion cube, so it uses the 0/1 indicators of C.3, and a coefficient is the effect of adding the mutation to P. On off-rate the evidence selects order 2 with log₁₀ Bayes factor +17.9 over the
additive model, and the structure is one large pair term: A182D × S178N, coefficient +0.98, against
mains in the same model of about −0.97 (A182D), −0.44 (T187Y) and −1.19 (S178N). The additive model, which the evidence rejects, gives −0.47, −0.36 and −0.65. The practical reading is that the first
stabilizer costs about one log of off-rate and further ones cost nothing, because P is beyond the
dissociation floor and the cost saturates at $\mathrm{p}k_{\mathrm{off}}$ about 3.1, outside the $K_D$ constraint. On
\(T_{\mathrm{collapse}}\) the evidence selects order 1 (A182D +8.1 °C, V196F +4.2 °C, T187Y +2.0 °C), and on reported $T_{\mathrm{agg}}$ order 2 by a negligible margin. The lead took
T187Y and V196A, two of the milder levers, with V196A preferred to V196F on HMW and replication.

\textbf{C.7 Selection.} Where the agent selected by stated hypothesis, choosing anchors, scans,
controls, the DR3 hero block (the chassis plus single mutations) and the corner set of DR2, the
selections produced the leads. Where selection was delegated to a learned acquisition function, the
120-well DR2 lattice sample chosen by a composite utility over three discriminative models, the set moved the
population and the utility did not rank inside it (Appendix F).

\section{Appendix D. Reading the raw curves}\label{appendix-d.-reading-the-raw-curves}

\setcounter{figure}{0}
\renewcommand{\thefigure}{D\arabic{figure}}

\textbf{D.1 What the instrument records.} nanoDSF heats each capillary on a ramp (0.3 °C/min in the
standard condition) and records two signals: the barycentric mean of the protein's intrinsic
tryptophan fluorescence, BCM\((T)\), which red-shifts as the protein unfolds, and static light
scattering at 266 nm, SLS\((T)\), which rises as particles form. The standard summary reduces these
to two metrics, $T_m^{\mathrm{app}}$ (the peak of \(|d\mathrm{BCM}/dT|\), Figure~\ref{fig:meltevent}) and $T_{\mathrm{agg}}$ (a scattering onset on
SLS), and treats them as two separate properties of the molecule. The agent fit the raw traces of every capillary (916 VHH wells over four rounds and 352 scFv wells in the final round) and found that on each scaffold one of the two metrics did not measure what its name implied.

\textbf{D.2 The measurement model.} Each well is fit with the two processes of Section 6.
Fluorescence follows Equation~\ref{eq:F}, with $\phi_N$ from Equation~\ref{eq:phi}.
Scattering is the instrument baseline plus what has accumulated from each state,
\[ \mathrm{SLS}(T) = b_0 + b_1 (T - T_0) + \kappa_N \int_{T_0}^{T} \phi_N(T')^2\, dT' + \kappa_U \int_{T_0}^{T} \big(1 - \phi_N(T')\big)^2\, dT', \]
where $\kappa_N$ and $\kappa_U$ are rates from folded and unfolded protein, in scattering counts per °C of ramp.
Whichever integral carries the rise says which state is aggregating.

\emph{Fitting.} The fit is staged.
Fluorescence gives $T_m$ and $\Delta H$, and scattering then gives $\kappa_N$ and $\kappa_U$ with $\phi_N$ held, so scattering cannot move the melt.
Scattering is fit only up to its peak, beyond which settling makes the signal fall.
While $\phi_N \approx 1$ the slope $b_1$ and $\kappa_N$ are collinear, so $b_0$ is held near the observed baseline, $|b_1|$ is bounded, and the two are separated only by the curvature $\phi_N^2$ imposes.

\emph{What counts as identified.} A well is \emph{identified} when $\kappa_N$ has a finite posterior interval away from its bounds (63\% of VHH wells).
A further 23\% are censored, where a flat trace gives only an upper bound, and 14\% have no usable two-state fit.
The design-level value is the median over identified standard-condition wells.
The scattering fit is rough in many wells (median $R^2$ 0.76 over the 590 wells with a value of $\mathrm{p}k_{\mathrm{agg}}$, lower on flat traces), so a single well's rate is approximate.

\emph{What the model does not establish.} The squared terms are an empirical form, not a mechanism.
On the identified VHH wells ($n = 531$), a form linear in $\phi_N$ fits equally well (median $R^2$ 0.85 against 0.83) and ranks $\kappa_N$ nearly the same way (Spearman 0.91).
Scattering scales close to first order in protein (D.3), where literal pairwise association would predict second order.
Adding a mixed term $\phi_N(1-\phi_N)$ improves a minority of wells and leaves $\kappa_N$ nearly unchanged (Spearman 0.95), but it makes $\kappa_U$ poorly identified, so no metric rests on $\kappa_U$.
A temperature-dependent $\kappa_N$ is preferred by BIC in 59\% of VHH wells with a fit, but the apparent activation energy is not a stable property of a molecule (intraclass correlation 0.38), so we report the window-average rate.

\begin{figure}[t]
\centering
\includegraphics[width=1\linewidth,height=\textheight,keepaspectratio,alt={The model on the parents and leads (the raw traces). Data are the averaged raw traces; dashed curves are the two-state plus native-aggregation fit of D.2 with constant \textbackslash kappa\_N, refit here on the averaged traces. (a) VHH parent: T\_m 63.9 °C, \textbackslash Delta H 211 kJ/mol, and a scattering rise carried by \textbackslash kappa\_N (the \textbackslash phi\_N\^{}2 term), i.e., the folded protein aggregating 20 °C below its melt. (b) VHH lead: T\_m 70.6 °C and a flat scattering trace, \textbackslash kappa\_N at its lower bound (censored). (c) scFv parent (DR1) and (d) scFv lead (DR2): the two-state model does not fit; the scattering peak (red dotted) coincides with the steepest fluorescence change (green dotted) and the fluorescence has no post-transition baseline, because the sample has left solution. The parent's BCM rises 3 nm and then falls; the lead's falls from 25 °C before the collapse. Every reported Tm on this scaffold is the position of that collapse.}]{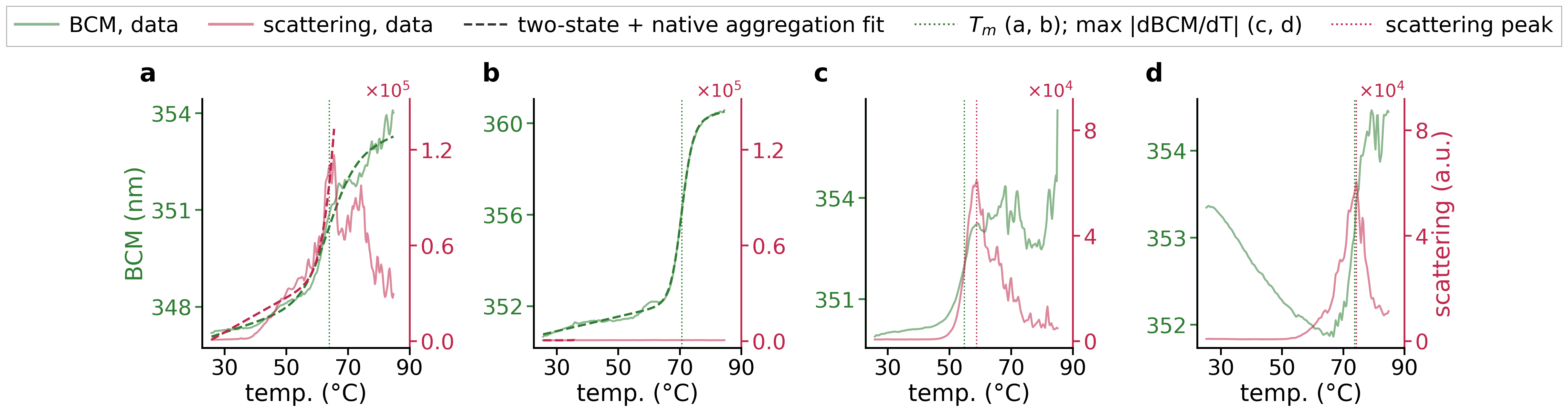}
\caption{\textbf{The model on the parents and leads.} Data are the averaged
traces of Figure~\ref{fig:traces}; dashed curves are the two-state plus native-aggregation fit of D.2 with constant
\(\kappa_N\), refit here on the averaged traces. (a) VHH parent: \(\hat{T}_m\) 64.0 °C and \(\Delta H\) 202
kJ/mol on the full range (\(\hat{T}_m\) falls to 62.0 °C if the fit stops at 75 °C), and a scattering rise that starts 20 °C below the melt. The per-well fits assign that rise to \(\kappa_N\) (the \(\phi_N^2\) term), the folded protein aggregating. On this averaged trace \(\kappa_N\) carries about 40\% of the rise at the peak and \(\kappa_U\) the rest. (b) VHH lead: \(\hat{T}_m\) 70.6 °C and a flat scattering trace,
\(\kappa_N\) at its lower bound (censored). (c) scFv parent (DR1) and (d) scFv lead (DR2): the
two-state model does not identify a melt; the scattering peak (red dotted) falls near the steepest
fluorescence change (green dotted; 3.5 °C apart on the parent, 0.5 °C on the lead) and the fluorescence has no post-transition baseline, because the
sample has left solution. The parent's BCM rises 3 nm and then falls; the lead's falls from 25 °C
before the collapse. Every reported $T_m^{\mathrm{app}}$ on this scaffold is the position of that
collapse.}\label{fig:d1}
\end{figure}

\textbf{D.3 The VHH: a real melt, and scattering that does not report it.} The parent unfolds at about 62 °C on every estimate we tried, matching BigHat's baseline of 61.8 ± 1.2 °C.
On the lead, the model-free derivative peak, the coupled fit and the multistate fit of D.5 agree within 0.4 °C.

\begin{figure}[t]
\centering
\includegraphics[width=1\linewidth,height=\textheight,keepaspectratio,alt={The thermal metrics across the final-round populations. (a) VHH DR3, standard-condition wells (n = 69): scattering onset against the identified melting temperature; onset sits near 40 °C regardless of the melt, the parent (open diamond) among the designs. (b) VHH DR3 load pairs (n = 41 designs): \textbackslash log\_\{10\}\textbackslash kappa\_N at the neat loading against the same design at half loading; the dotted line is the first-order expectation (+0.30), the dashed line identity. (c, d) scFv DR2 (n = 101 and 108): reported Tm and reported Tagg against the collapse temperature; both are summary statistics of the same event, about 10 °C apart.}]{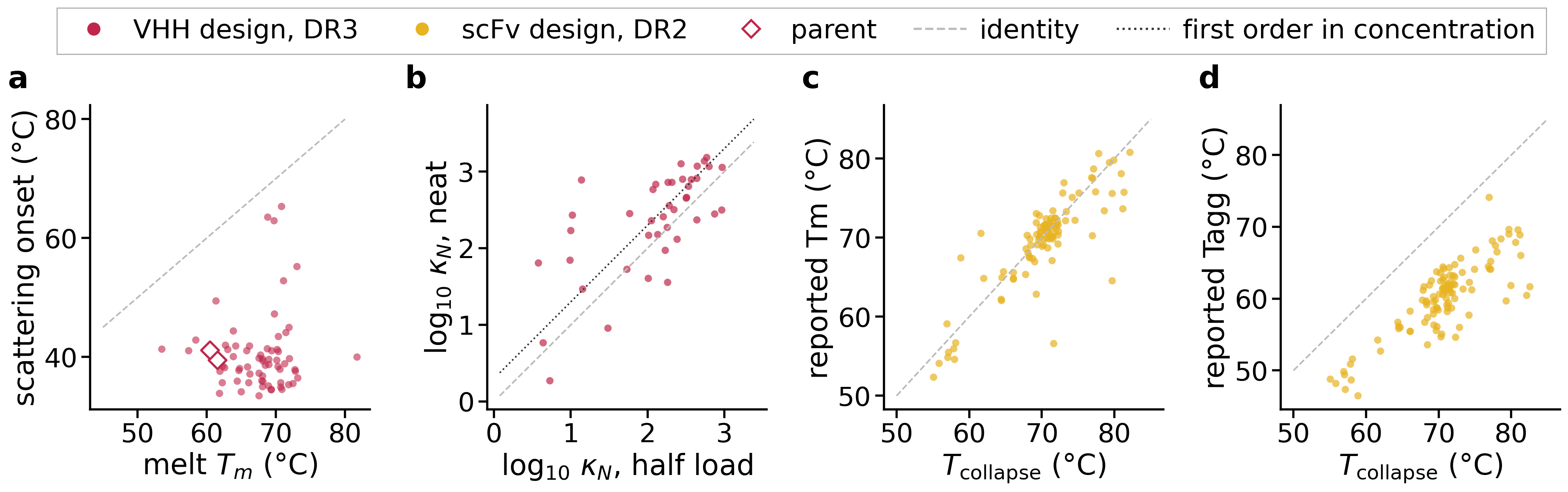}
\caption{\textbf{The thermal metrics across the final-round populations.} (a) VHH DR3,
standard-condition wells (n = 69): scattering onset against the identified melting temperature;
onset sits near 40 °C regardless of the melt, the parent (open diamond) among the designs. (b) VHH DR3
load pairs (n = 41 designs): \(\log_{10}\kappa_N\) at the neat loading against the same design at
half loading; the dotted line is the first-order expectation (+0.30), the dashed line identity. (c,
d) scFv DR2 (n = 101 and 108): reported $T_m^{\mathrm{app}}$ and reported $T_{\mathrm{agg}}$ against the collapse temperature; both
are summary statistics of the same event, about 10 °C apart.}\label{fig:d2}
\end{figure}

\emph{Onset does not track the melt.} Across the final-round designs, the fitted scattering onset sits near 40 °C regardless of the melt (Figure~\ref{fig:d2}a).
It is 40--46 °C on the parent (melt 62 °C), 38--48 °C on a non-binding control (melt 72 °C), and a median 39 °C over 69 standard-condition wells whose melts span 53--82 °C, with a correlation of 0.08.
The native term carries the rise in 55\% of wells, and the unfolded term alone in 9\%.
Scattering on this scaffold therefore reports folded protein associating well below its melt.

\emph{Onset does not track the rate.} The fitted \(\kappa_N\) of parent and control differ by 18× in this round and by 57× to 680× in earlier ones, though their onsets do not.
BigHat's QC recognizes the behavior, classifying 491 of the 916 VHH wells, 18 of the parent's 20 among them, as ``Aggregation before Tm call or aggregation only''.
The flag does not say how fast a molecule aggregates, so the metric became the rate itself, standardized for loading,
\[ k_{\mathrm{agg}} = \kappa_N / c, \]
reported as \(\mathrm{p}k_{\mathrm{agg}} = -\log_{10} k_{\mathrm{agg}}\), with \(c\) the loaded concentration in ng/µL.
On 41 designs run at two loadings on one plate, \(\log_{10}\kappa_N\) fell by a median 0.27 at half load, against 0.30 for a first-order process (Figure~\ref{fig:d2}b), so \(k_{\mathrm{agg}}\) holds across the twofold range tested.

\emph{The lead has no onset to call.} The onset model has nothing to fit on a flat trace.
The lead's standard-condition scattering ends the ramp within about 1.1$\times$ of baseline, with isolated spikes to 3.8$\times$ in one well and one fast-ramp well drifting to about 1.5$\times$, against 10 to 300$\times$ on the parent.
The software's onset calls on the lead (32--59 °C) all failed QC, as they did for four of the five finalists (Figure~\ref{fig:d1}b).
The pipeline records a QC-failed onset as 90 °C, while the design table's pass flag uses the failed onset and reads false for every DR3 design, and the lead decision recorded the constraint as not met.
We treat the lead's $T_{\mathrm{agg}}$ as right-censored above 85 °C on the evidence of the raw traces.

\textbf{D.4 The scFv: no melt to measure.} At the assay's 0.15--1.3 mg/mL, scattering begins at the foot of the first fluorescence transition and the sample precipitates a median 9.6 °C later, before any unfolding midpoint is reached (Figure~\ref{fig:d1}c, d).
The two-state model of D.2 has no upper baseline to fit and returns \(\hat{T}_m\) at a bound.

The reported $T_m^{\mathrm{app}}$ and $T_{\mathrm{agg}}$ on this scaffold are two readings of one kinetic collapse.
Across the 101 final-round designs with a QC-passing $T_m^{\mathrm{app}}$, it tracks the collapse temperature at Spearman 0.78 with a median offset of −0.2 °C.
Across the 108 with a reported $T_{\mathrm{agg}}$, that metric tracks it at 0.74 with a median offset of −9.8 °C (Figure~\ref{fig:d2}c, d).
The two constraints are therefore one requirement measured twice, and $T_{\mathrm{agg}}$ ≥ 65 °C is the one that binds.

BigHat had stated the rule before the model existed.
Its answers of 20 July give the specification as the lower of $T_m^{\mathrm{app}}$ and $T_{\mathrm{agg}}$, and note that a $T_m^{\mathrm{app}}$ above $T_{\mathrm{agg}}$ may be aggregate melting and that $T_{\mathrm{agg}}$ depends on concentration.
The model of D.5 makes that rule quantitative, and the test of Appendix E checks it.

\textbf{D.5 The multistate model and the collapse temperature.} To read a cascade rather than a single melt, the agent fit each well with a family of schemes: two-state and three-state, each with an optional term that lets aggregated protein carry its own fluorescence signature.
The scheme is chosen by BIC, provided no parameter sits at a bound and the transition falls inside the fitted window.

\emph{On the VHH} (548 wells), the model classes 42\% as two-state, 42\% as two-state with an aggregation-linked shoulder, 7\% as carrying a real intermediate and 9\% as poor fits.
Its identified midpoint agrees with the coupled fit of D.2 to a median 0.36 °C with no bias.
On the parent, 8 of the 20 reported $T_m^{\mathrm{app}}$ calls fall at 44--48 °C, on an aggregation-linked shoulder.
BigHat's QC flags 6 of those 8 wells, and the model returns the melt on them as well, in agreement with BigHat's baseline.

\emph{On the scFv}, the scheme that fits is the collapse, whose transition is the scattering peak.
Equation~\ref{eq:collapse} standardizes it to $c_{\mathrm{ref}} = 350$ ng/µL in every round, with $\gamma_c = -2.6 \pm 0.4$ °C per e-fold fit on the designed load pair of Appendix E.
Earlier rounds used a within-design prior of −4.0 ± 1.1.
Applying that prior to the final round moves the scFv lead by 0.7 °C and leaves the count of designs passing the revised specification unchanged.

We adopted the collapse temperature over the reported call for its coverage (100\% of wells against 80\%) and cross-round stability (−1.5 °C drift against −8.0 °C for a two-state fit).
Its replicate SD, 1.1--2.0 °C, is no better than the reported call's.
Its main limit is that second-order kinetics predict a design-dependent slope, while one pooled slope is applied across designs whose concentrations differ by more than 3 e-folds.

\textbf{D.6 What the two interpretations changed.} They explain why the same thermal recipes behaved differently on the two scaffolds.
Designed disulfides and interface-methionine designs, aimed at raising a melt, cost aggregation on the VHH and failed outright on the scFv, where the limit is the collapse and the two repairs that raised it, Q91A and S245A, are packing fixes found by scanning.

They also explain the learnability result of Appendix F.
The reported thermal metrics were learnable on the scFv, where they are summary statistics of the collapse, and near zero across rounds on the VHH, where $T_{\mathrm{agg}}$ says little about the molecule (among wells with a QC-passing $T_{\mathrm{agg}}$, onset and $\mathrm{p}k_{\mathrm{agg}}$ correlate at Spearman $+0.29$, n = 464).

\section{Appendix E. The designed perturbation
test}\label{appendix-e.-the-designed-perturbation-test}

\setcounter{figure}{0}
\renewcommand{\thefigure}{E\arabic{figure}}
\setcounter{table}{0}
\renewcommand{\thetable}{E\arabic{table}}

\textbf{E.1 Why the protocol had to change.} A model of the traces makes predictions that the
standard protocol cannot test, because every capillary is run at one concentration and one ramp
rate. The model of Appendix D says which metrics are thermodynamic and which are kinetic. A
thermodynamic midpoint does not depend on how much protein is in the capillary or how fast it is
heated; an aggregation rate is first order in concentration, so \(\kappa_N\) halves at half load and
\(k_{\mathrm{agg}} = \kappa_N / c\) does not move; and a kinetic collapse, a threshold crossed on a ramp, moves later
at lower concentration and at a faster ramp. On the VHH that separates the melt (invariant) from the
scattering onset (moves). On the scFv it says that every thermal metric should move, by predictable
amounts, and that the reported $T_m^{\mathrm{app}}$, if it were a melt, should not.

\textbf{E.2 What was run.} For the final round of each campaign the agent proposed running each
design at two loadings (neat and one-half) on one plate and at two ramp rates on the other, as
designed perturbations. BigHat proposed a slower second ramp; the agent asked for a faster one, 1.0
°C/min, on the reasoning that a fast ramp outruns aggregation and might expose the unfolding
transition before the collapse, and BigHat agreed. An isothermal hold was deferred to mammalian
material because cell-free material is below 1 mg/mL, and a one-quarter loading was too dilute to
run. The synthesis scale was doubled to supply the extra capillaries. A protocol record (2026-08-12)
and a prediction file (2026-08-19) wrote down, before the plates ran, which metrics should move,
which should not, and by how much (Table~\ref{tab:e1}).

\begin{figure}[t]
\centering
\includegraphics[width=1\linewidth,height=\textheight,keepaspectratio,alt={The designed perturbation test, scored per design. Paired shifts, half load minus neat and fast ramp minus standard, for every design measured in both conditions; bars are medians; gray bands and dotted lines are the pre-registered predictions. (a) VHH identified melt (n = 74, 55): invariant. (b) VHH scattering onset (n = 49, 49): moves. (c) scFv collapse temperature (n = 74, 42, observed collapses only): +2.0 °C against 2.8 predicted at half load; +4.7 °C at the faster ramp, above the pre-registered +3--4 °C and inside the 2--6 °C band written at scoring. (d) scFv reported calls: Tagg tracks the collapse (n = 84, 77); Tm moves the wrong way under load (n = 73) and with the collapse under ramp (n = 63).}]{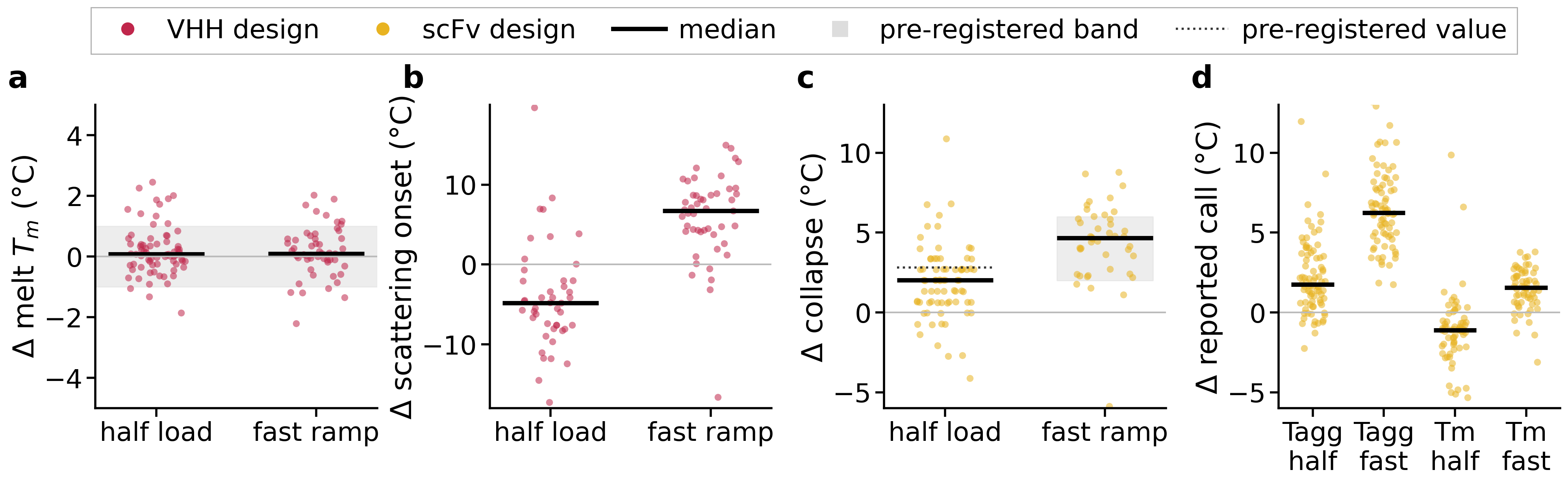}
\caption{\textbf{The designed perturbation test, scored per design.} Paired shifts, half load minus
neat and fast ramp minus standard, for every design measured in both conditions; bars are medians;
the axes clip 22 extreme points (5 in a, 5 in b, 12 in d); gray bands and dotted lines are the pre-registered predictions. (a) VHH identified melt (n = 74,
55): invariant. (b) VHH scattering onset (n = 49, 49): moves. (c) scFv collapse temperature (n = 74,
42, observed collapses only): +2.0 °C against 2.8 °C predicted at half load; +4.7 °C at the faster ramp, above the pre-registered +3--4 °C and inside the 2--6 °C band written at scoring. (d) scFv reported calls: $T_{\mathrm{agg}}$ tracks the collapse (n = 84, 77); $T_m^{\mathrm{app}}$
moves the wrong way under load (n = 73) and with the collapse under ramp (n = 63).}\label{fig:e1}
\end{figure}

\textbf{E.3 Scoring.} We re-scored every prediction from the per-capillary and per-condition tables,
pairing each design with itself across conditions and taking the median over designs of the paired shift, with its
interquartile range (Figure~\ref{fig:e1}, Table~\ref{tab:e1}). The unit is the design unless stated, and a design with several capillaries enters once. On the VHH the identified melt moved +0.08 °C at half
load (n = 74 pairs) and +0.09 °C at the faster ramp (n = 55), inside the pre-registered 1 °C band;
\(\mathrm{p}k_{\mathrm{agg}}\) moved −0.03 log at half load (n = 41; Figure~\ref{fig:d2}b); the scattering onset moved −4.8 °C at half
load and +6.7 °C at the faster ramp (n = 49 each). Reported $T_{\mathrm{agg}}$ moved −1.8 °C at half load (n = 68, p = 0.13) and +4.6 °C at the faster ramp, and the sign of the half-load shift reverses across plates. On the scFv the collapse rose 2.0 °C at half load
(n = 74 designs with an observed collapse in both conditions, from 75 capillary pairs) against 2.8 °C predicted, and 4.7 °C at the
faster ramp (n = 42 designs; 4.45 °C over 45 capillary pairs), above the pre-registered +3--4 °C and inside the 2--6 °C band written at scoring; 43 of the 88 fast-ramp capillaries
never collapsed inside the ramp and are right-censored, so the ramp shift is a lower bound on the
population. \(\log_{10}\kappa_N\) fell by a median 0.41 at half load (IQR −0.64 to −0.04, n = 51
designs with an identified rate in both conditions) against the 0.30 that halving predicts; the multistate fit's rate estimator gives −0.32 per design (n = 84) and −0.30 per capillary (n = 88). Reported $T_{\mathrm{agg}}$ tracked the
collapse (+1.8 and +6.2 °C); reported $T_m^{\mathrm{app}}$ moved the wrong way under load (−1.1 °C, opposite in sign
to the collapse) and coherently under ramp (+1.6 °C). Three predictions failed: the reported $T_m^{\mathrm{app}}$ was predicted erratic under both perturbations and was erratic only under
load, and the identified two-state midpoint on the 20 scFv wells where the multistate model resolves
one moved −1.4 and +2.4 °C, outside the 1 °C band, and the VHH scattering onset moved the wrong way at half load. The scFv ramp shift was also not larger for early aggregators, as had been predicted.

\textbf{E.4 What the test established and what it did not.} The VHH melt held, the scFv kinetic metrics moved in the predicted direction at close to the predicted size, and the metrics the constraints were written on moved with the perturbation.

The comparisons differ in quality.
The scFv load pair ran about two hours apart.
The VHH load pair shared a plate design but ran about 19 hours apart, and the journal records a run-day term on its onset contrast.
The two ramp rates ran on different days in both campaigns, with no instrument control, so a day term (about +5 to +7 °C on the VHH onset) is aliased one-to-one with the ramp contrast.
We therefore report the ramp shifts as confounded, along with the concentration order and activation energy inferred from them.
The melt result does not depend on this, since the melt held steady across the days as well.

\begin{table}[t]
\centering
\footnotesize
\caption{Pre-registered predictions of the perturbation test (protocol record 2026-08-12; prediction
file 2026-08-19) and their scores, recomputed from the per-condition tables.\label{tab:e1}}
\begin{tabular}{@{}
  >{\raggedright\arraybackslash}p{(\linewidth - 10\tabcolsep) * \real{0.2128}}
  >{\raggedright\arraybackslash}p{(\linewidth - 10\tabcolsep) * \real{0.0638}}
  >{\raggedright\arraybackslash}p{(\linewidth - 10\tabcolsep) * \real{0.1064}}
  >{\raggedright\arraybackslash}p{(\linewidth - 10\tabcolsep) * \real{0.1277}}
  >{\raggedright\arraybackslash}p{(\linewidth - 10\tabcolsep) * \real{0.2872}}
  >{\raggedright\arraybackslash}p{(\linewidth - 10\tabcolsep) * \real{0.2021}}@{}}
\toprule\noalign{}
\begin{minipage}[b]{\linewidth}\raggedright
prediction
\end{minipage} & \begin{minipage}[b]{\linewidth}\raggedright
scaffold
\end{minipage} & \begin{minipage}[b]{\linewidth}\raggedright
metric
\end{minipage} & \begin{minipage}[b]{\linewidth}\raggedright
pre-registered
\end{minipage} & \begin{minipage}[b]{\linewidth}\raggedright
observed (paired median, IQR)
\end{minipage} & \begin{minipage}[b]{\linewidth}\raggedright
verdict
\end{minipage} \\
\midrule\noalign{}
melt invariant at half load & VHH & \(\hat{T}_m\) & \textbar shift\textbar{} \textless{} 1 °C & +0.08
{[}−0.25, +0.60{]}, n = 74 & confirmed \\
melt invariant at fast ramp & VHH & \(\hat{T}_m\) & \textbar shift\textbar{} \textless{} 1 °C & +0.09
{[}−0.25, +0.60{]}, n = 55 & confirmed \\
\(k_{\mathrm{agg}}\) invariant at half load & VHH & \(\mathrm{p}k_{\mathrm{agg}}\) & ≈ 0 & −0.03 {[}−0.28, +0.24{]} log, n = 41 &
confirmed \\
scattering onset moves with load & VHH & onset & moves & −4.8 {[}−8.0, −2.0{]} °C, n = 49 & refuted on sign; run-day term not separated \\
scattering onset moves with ramp & VHH & onset & moves & +6.7 {[}+4.1, +8.9{]} °C, n = 49 & moves;
day-aliased \\
collapse +2.8 °C at half load & scFv & \(T_{\mathrm{collapse}}\) & +2.8 °C & +2.0 {[}+0.6, +2.7{]},
n = 74 & confirmed in sign, 70\% of size \\
collapse +3--4 °C at fast ramp & scFv & \(T_{\mathrm{collapse}}\) & +3--4 °C (band of +2 to +6 °C written at scoring) & +4.7 {[}+3.1, +5.9{]}, n = 42; 43/88 censored & above the prediction, inside the scoring band; day-aliased \\
\(\kappa_N\) halves at half load & scFv & \(\log_{10}\kappa_N\) & not pre-registered; first-order expectation −0.30 & −0.41 {[}−0.64, −0.04{]},
n = 51 designs (second estimator −0.32, n = 84) & consistent within the IQR \\
reported $T_{\mathrm{agg}}$ tracks the collapse & scFv & $T_{\mathrm{agg}}$ & moves with collapse & +1.8 (load), +6.2 (ramp) &
confirmed \\
reported $T_m^{\mathrm{app}}$ erratic & scFv & $T_m^{\mathrm{app}}$ & erratic & −1.1 {[}−2.3, −0.5{]} (load); +1.6 {[}+0.7, +2.5{]}
(ramp) & confirmed on load, refuted on ramp \\
resolved unfolding midpoint invariant & scFv & two-state \(\hat{T}_m\) & \textbar shift\textbar{}
\textless{} 1 °C & −1.4 (load, n = 20); +2.4 (ramp, n = 21) & refuted \\
\bottomrule\noalign{}
\end{tabular}
\end{table}

\textbf{E.5 The revised specifications.} \(\mathrm{p}k_{\mathrm{agg}}\) was defined on 30 July and \(T_{\mathrm{collapse}}\) adopted on 10 August, and the predictions were filed on 13 and 19 August, before the perturbation plates were read.
The final specifications were set knowing the results, on 3 September for the VHH and 4 September for the scFv (from a draft of 26 August), and the leads were chosen on 4 September.

Against the original specifications, the VHH lead passes with its $T_{\mathrm{agg}}$ right-censored above the ramp (Appendix D.3).
On the scFv, 0 of 113 DR2 designs pass $K_D$, $T_m^{\mathrm{app}}$ and $T_{\mathrm{agg}}$ together with a measured $K_D$, and two pass with $\mathrm{p}k_{\mathrm{off}} \geq 4.3$ counted (15 pass $K_D$, 84 pass $T_m^{\mathrm{app}}$, 17 pass $T_{\mathrm{agg}}$).

The lead decisions used replacements the agent proposed.
For the VHH, these were \(\mathrm{p}k_{\mathrm{agg}}\) at or above the value of the chassis, the DR2 design the final round was built on (+1.72 on three identified wells), and day-0 high-molecular-weight (HMW) fraction by size exclusion ≤ 5\% in place of $T_{\mathrm{agg}}$, with the melt constraint kept at 70 °C on the identified midpoint.
For the scFv, these were \(T_{\mathrm{collapse}}\) ≥ 70 °C in place of $T_m^{\mathrm{app}}$ and $T_{\mathrm{agg}}$, with $K_D$ ≤ 1 nM, HMW ≤ 3\% and purity ≥ 95\%.
Seven of 113 DR2 designs pass with a measured $K_D$, the lead among them, and 26 pass with those at $\mathrm{p}k_{\mathrm{off}} \geq 4.3$ counted.
Both leads pass their revised specifications.
BigHat agreed that both campaigns were completed successfully.

\FloatBarrier

\section{Appendix F. Specialized models in full}\label{appendix-f.-specialized-models-in-full}

\setcounter{figure}{0}
\renewcommand{\thefigure}{F\arabic{figure}}
\setcounter{table}{0}
\renewcommand{\thetable}{F\arabic{table}}

\textbf{F.1 The test.} The agent ran specialized models inside the campaigns, against pre-registered predictions. Every score below is prospective unless marked: trained on earlier rounds, scored on the next round's data
against a prediction file written before the plate ran. Rank correlations are Spearman ρ with 90\%
bootstrap intervals; Table~\ref{tab:f1} gives the register.

\textbf{F.2 Structure and sequence models.} Structure co-folding (ESMFold2, \citealp{candido2026}; Protenix, \citealp{protenix2025}; Boltz-2,
\citealp{boltz2}) did not separate binders from non-binders on either scaffold: on
the VHH two of three models ranked binders below non-binders, on the scFv the interface confidence
(ipTM; \citealp{evans2021}) gave an AUROC of 0.13 against decoys designed for another target, and the four tightest scFv binders on the
plate had the lowest median ipTM of any class. An in-sample binder-versus-non-binder AUROC of 0.72
at a wider mesothelin crop fell to 0.48 on the next two rounds. The scores tracked distance from the parent. ProteinMPNN \citep{dauparas2022} redesigns of the scFv CDR-H3 gave
three non-binders in three wells, and its per-position preferences were
right at two positions and wrong at two, including a probability of 1.00 for the glycine whose
reversion, G230Y, is the campaign's largest affinity lever. Antibody language-model naturalness
(IgBert; \citealp{kenlay2024}) was anti-correlated with binding on the scFv (AUROC 0.28 over 359 designs,
retrospective, masked marginals on the parent), because the most natural sequences on the plate are germline reversions that do not
bind; language-model embeddings did no better than one-hot encodings. Published thermostability
predictors were checked against their own validation and rejected; the one run, NanoMelt
\citep{ramon2025}, dropped nothing from 89 designs and predicted +6.5 °C where +1
°C was measured.

\begin{figure}[t]
\centering
\includegraphics[width=1\linewidth,height=\textheight,keepaspectratio,alt={The specialized models, scored prospectively. (a) VHH: rank correlation of the Gaussian process and the CNN trained on DR0+DR1 and scored on the 131 DR2 designs, per target, with 90\% intervals; the refit metrics transfer, the reported metrics do not. (b) scFv: DR0 → DR1 (n = 50--116); reported calls for the CNN only, from the ensemble table. (c) VHH DR2: the same models inside the 120-well balanced lattice sample (light) and on the 30 designs outside it (dark). (d) The DR2 acquisition step: rank correlation between the design-time BOtied utility and its realized value on the 84 selected designs measured, for the mean of the two models, each model alone, and the corner utility (n = 98).}]{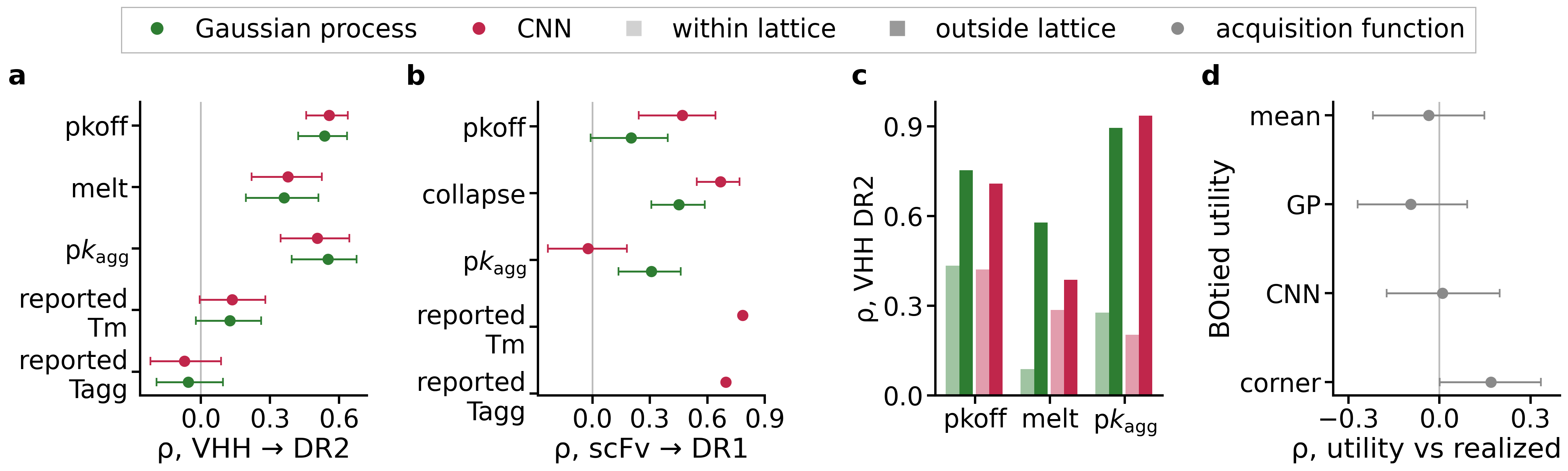}
\caption{\textbf{The specialized models, scored prospectively.} (a) VHH: rank correlation of the
Gaussian process and the CNN trained on DR0+DR1 and scored on the 131 DR2 designs, per target, with
90\% intervals; the refit metrics transfer, the reported metrics do not. (b) scFv: DR0 → DR1 (n =
50--116); reported calls for the CNN only, from the ensemble table. (c) VHH DR2: the same models
inside the 120-well balanced lattice sample (light) and on the 30 designs outside it (dark). (d) The DR2
acquisition step: rank correlation between the design-time BOtied utility and its realized value on
the 84 selected designs measured, for the mean of the two models, each model alone, and the corner
utility (n = 98).}\label{fig:f1}
\end{figure}

\textbf{F.3 Discriminative models.} Two discriminative models were carried through both campaigns:
an observation-only Gaussian process on a sequence kernel, and a 16,000-parameter convolutional
network trained as a five-seed ensemble (Figure~\ref{fig:f1}a, b). The GP transferred off-rate rank across
rounds at ρ 0.59 (DR0 → DR1, n = 74) and 0.54 (DR0+DR1 → DR2, n = 131) on the VHH with calibrated 95\%
intervals (coverage 0.96 and 0.99), against 0.10 for counting mutations from the parent; a
Tobit-EM variant that won in-round cross-validation was the worse arm prospectively (coverage 0.45,
bias −0.9 log) and was retired. The CNN matched the GP on off-rate (0.60 and 0.56) and beat it on
the scFv collapse temperature, 0.67 {[}0.55, 0.77{]} against 0.45 {[}0.31, 0.59{]} (n =
116). On the VHH DR3 designs never measured before, scored from the predictions filed on 2026-08-19, the CNN reached ρ 0.78 on the melt against the GP's 0.70 (n = 60) and 0.83 on \(\mathrm{p}k_{\mathrm{agg}}\) against the GP's 0.87 (n = 69). Its intervals under-covered everywhere
(0.23--0.88), so we used it for ranking and the GP for calibrated intervals. Two results in Figure~\ref{fig:f1} bear on Section 6.
First, the reported thermal metrics were learnable on the scFv (CNN ρ 0.78 on reported $T_m^{\mathrm{app}}$, 0.70 on
reported $T_{\mathrm{agg}}$, n = 132 and 140) and at zero on the VHH (CNN ρ 0.13 and −0.07 for reported $T_m^{\mathrm{app}}$ and
$T_{\mathrm{agg}}$), for the reason in Appendix D: on the scFv they are summary statistics of the collapse; on the VHH $T_{\mathrm{agg}}$
says little about the molecule. The refit metrics were learnable on both. Second, inside
a single balanced lattice sample every model fell to ρ 0.09--0.43 (Figure~\ref{fig:f1}c): across rounds the models rank the population as a whole, and inside a balanced lattice sample they do not rank individual designs.

\textbf{F.4 The acquisition step.} For the VHH's DR2 the agent ran a model-driven selection: the GP
and the CNN scored a Sobol candidate pool of 4,096 over 17 mutations, a multi-objective acquisition
function (after BOtied; \citealp{park2023}) ranked the pool by Pareto rank over a Gaussian
copula of the three objectives, and a greedy rebalancing step chose 120 wells so that every mutation
stayed near 50\% occupancy; no generative proposal model was trained. The correlation between
predicted and realized utility inside the selected set was ρ −0.04 {[}−0.22, +0.15{]} (n = 84;
Figure~\ref{fig:f1}d), and neither model alone did better (−0.10 and +0.01); the top and bottom halves by
predicted utility were indistinguishable on every objective. The set as a whole was good, at the
0.95 quantile of everything measured before, and the record attributes that to the anchor frame and
the additive levers moving the block, which the models had learned, rather than to the selection
within it, which they had not. The lattice sample is the set on which the models had the most data and the
tightest design, so it is the fairest test the campaigns offer; a composite of three weak, mutually
anti-correlated rankers would be difficult for any acquisition function. This is not the full
LaMBO-2 loop \citep{gruver2023}, which also trains a
guided generative model to propose candidates; it is the discriminative and acquisition half of it.

\textbf{F.5 The additive model as a predictor.} The Walsh regression of Appendix C, fit on DR0--DR2
and applied to DR3, predicted the chassis off-rate at 3.29 against a measured 3.15 and, scored from the predictions filed on 2026-08-19, ranked the
new DR3 designs at ρ 0.95 on off-rate (n = 69), 0.67 on the melt (n = 44) and 0.78 on \(\mathrm{p}k_{\mathrm{agg}}\) (n = 68);
the off-rate figure is inflated by the E1A split the cube builds in (the CNN's 0.90 on the same
designs falls to about 0.4 within either half of the cube). It over-predicted the melt of the hero block by about
5 °C, which is the cost of assuming additivity across the chassis boundary. It was useful for
off-rate and aggregation, where the levers are additive, and as an experiment-design tool, where it
says which cell to measure next.

\begin{table}[t]
\centering
\footnotesize
\caption{The verdict register: each model family the campaigns ran, its prospective score, and our
verdict.\label{tab:f1}}
\begin{tabular}{@{}
  >{\raggedright\arraybackslash}p{(\linewidth - 4\tabcolsep) * \real{0.2769}}
  >{\raggedright\arraybackslash}p{(\linewidth - 4\tabcolsep) * \real{0.4308}}
  >{\raggedright\arraybackslash}p{(\linewidth - 4\tabcolsep) * \real{0.2923}}@{}}
\toprule\noalign{}
\begin{minipage}[b]{\linewidth}\raggedright
component
\end{minipage} & \begin{minipage}[b]{\linewidth}\raggedright
prospective score
\end{minipage} & \begin{minipage}[b]{\linewidth}\raggedright
verdict
\end{minipage} \\
\midrule\noalign{}
Co-fold scores (three models) & decoy AUROC 0.13 (scFv); 2 of 3 wrong-signed (VHH) & not a filter \\
ProteinMPNN & 0 of 3 grafts bound; per-position 2 right, 2 wrong & did not help as a generator \\
IgBert naturalness / PLM embeddings & AUROC 0.28 (retrospective); ≤ one-hot & did not help here \\
$T_m$ proxies (NanoMelt; pLDDT, \citealp{jumper2021}; FoldX, \citealp{schymkowitz2005}) & 0 drops; predicted +6.5 °C vs measured +1 & did not help here \\
GP, observation-only & ρ 0.59 / 0.54 (off-rate); cov95 0.96 / 0.99 & useful; the calibrated arm \\
GP, Tobit-EM & cov95 0.45; bias −0.9 log & retired \\
CNN & ρ 0.56 (off-rate); 0.67 (collapse); 0.83 (aggregation) & useful for ranking; uncalibrated \\
Any model inside a balanced lattice sample & ρ 0.09--0.43 & weak \\
BOtied composite acquisition & ρ −0.04 {[}−0.22, +0.15{]} within the selected set & no ranking signal \\
Additive Walsh model & chassis off-rate 3.29 vs 3.15; melt +5 °C over & useful for off-rate and
aggregation \\
Lineage and literature priors & 100\% binders in VHH lineage blocks; 5 of 8 lead mutations &
highest yield per well \\
\bottomrule\noalign{}
\end{tabular}
\end{table}

\section{Appendix G. Division of labor, from the
journal}\label{appendix-g.-division-of-labor-from-the-journal}

\setcounter{figure}{0}
\renewcommand{\thefigure}{G\arabic{figure}}
\setcounter{table}{0}
\renewcommand{\thetable}{G\arabic{table}}

\begin{figure}[t]
\centering
\includegraphics[width=1\linewidth,height=\textheight,keepaspectratio,alt={The seven rounds and where the researcher appears in the record. Every journal record by date for (a) the VHH and (b) the scFv campaign; rounds run from one design submission to the next, with the record count in each; the dotted line is the lead decisions. Filled circles are records the researcher authored or co-signed, open circles agent records that name the researcher, squares the partner.}]{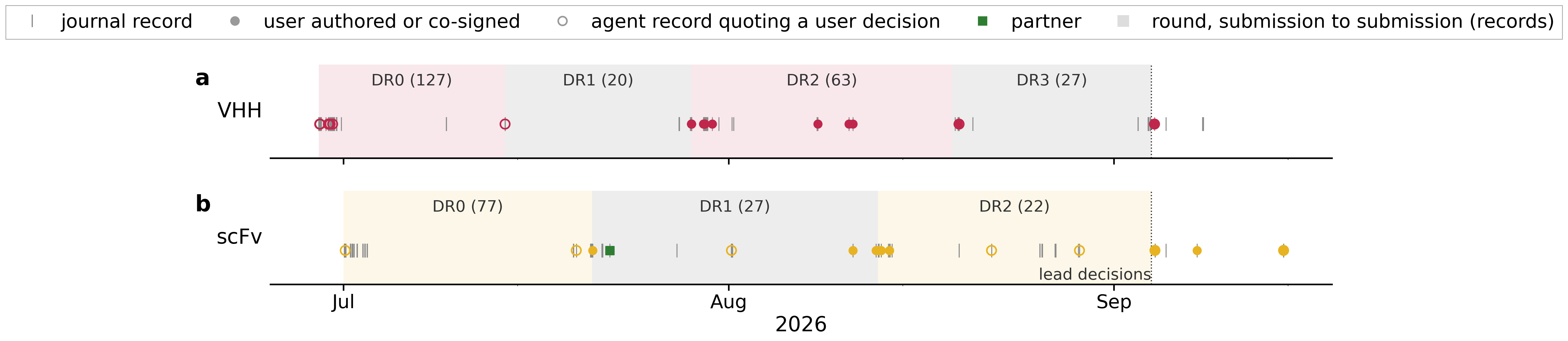}
\caption{\textbf{The seven rounds and where the researcher appears in the record.} Every journal record by
date for (a) the VHH and (b) the scFv campaign; rounds run from one design submission to the next,
with the record count in each; the dotted line is the lead decisions. Filled circles are records the
researcher authored or co-signed, open circles agent records that name the researcher, squares the
partner.}\label{fig:g1}
\end{figure}

\textbf{G.1 The record.} Each journal record carries an author field: \texttt{coord} (the round-0
orchestrating process) and \texttt{cc} (the agent in Claude Code, sometimes with a named
sub-process) for the agent, \texttt{A1}--\texttt{A9} for sub-agents, \texttt{sam} or
\texttt{samuelstanton} for the researcher, \texttt{bighat} for the partner, and compound entries such as
\texttt{cc+sam} for decisions taken together. Of the 387 journal records through the lead decisions, 27 carry the researcher in the
author field, 2 the partner (one of them shared with the researcher), and a further 19 agent-authored records name the researcher
inside the body, by a simple word match (Figure~\ref{fig:g1}); the rest are the agent's designs, analyses, findings and corrections.
The human-attributed records cluster at the design-round boundaries, where wells were committed, and
at the lead decisions; the long runs of agent-only records between them are analysis.

\textbf{G.2 The interventions, by category.} Table~\ref{tab:g1} gives one or more cited records per category.
Everything partner-facing went through the researcher: no plate submission, protocol request or question
list reached BigHat without him sending it. He set each campaign's objective from the partner's specification. Plate composition and budget is the
densest category: the DR3 block budget is quoted in his words, and the reallocation for the last VHH
round (``ideally this should be our last [VHH] round, so we want to be greedy'') cut exploration
wells and enlarged the hero block, the chassis-plus-one designs, from which the lead came. The
final-round protocol changes were agreed three ways; the agent's experimental suggestions were
useful, and BigHat amended them for two kinds of reason. Some ran into constraints of the protocol
itself: the isothermal hold was deferred because cell-free yields are below 1 mg/mL, and a
one-quarter loading was too dilute to run. Others were amended for time, cost and labor. The agent
is not always calibrated on the difficulty of the experiments it requests, and the experimental side
is where the most room remains. The multistate thermal model and the learnability studies are
co-signed. Both lead decisions record that the ranking was the agent's and the decision the researcher's.

\begin{table}[t]
\centering
\footnotesize
\caption{The researcher's interventions by category, with the journal records that attest to them (dates
are 2026).\label{tab:g1}}
\begin{tabular}{@{}
  >{\raggedright\arraybackslash}p{(\linewidth - 6\tabcolsep) * \real{0.1176}}
  >{\raggedright\arraybackslash}p{(\linewidth - 6\tabcolsep) * \real{0.3725}}
  >{\raggedright\arraybackslash}p{(\linewidth - 6\tabcolsep) * \real{0.1569}}
  >{\raggedright\arraybackslash}p{(\linewidth - 6\tabcolsep) * \real{0.3529}}@{}}
\toprule\noalign{}
\begin{minipage}[b]{\linewidth}\raggedright
category
\end{minipage} & \begin{minipage}[b]{\linewidth}\raggedright
record (campaign, kind @ date)
\end{minipage} & \begin{minipage}[b]{\linewidth}\raggedright
author field
\end{minipage} & \begin{minipage}[b]{\linewidth}\raggedright
what the researcher did
\end{minipage} \\
\midrule\noalign{}
partner interface & VHH \texttt{submission\_plan} @ 06-29; scFv \texttt{dr1\_sent\_to\_bighat} @
07-21 & coord; sam & nothing reached BigHat without the researcher sending it \\
tool policy & VHH \texttt{policy\_decision} @ 06-29 & coord & released a parked
thermostability-predictor run (``all sys go'') \\
plate ownership & VHH \texttt{plate\_B\_v1} @ 06-29 & coord & confirmed BigHat's eight control wells
per plate; overlapping anchors dropped \\
objective & scFv \texttt{spec\_read} @ 07-01; VHH \texttt{constraint} @ 07-29 & coord; samuelstanton
& set the campaign objective from the partner's specification \\
design co-decisions & VHH \texttt{reframe}, \texttt{v4\_design}, \texttt{v5\_design},
\texttt{v5\_shipped} @ 07-29; \texttt{r2\_v5\_heroes} @ 07-30 & cc+sam & the DR2 reframe (the
affinity block became an aggregation-reduction block) and the swap of two designed hero variants
into the plate \\
budget & VHH \texttt{r3\_selector\_spec} @ 07-29; \texttt{r3\_design\_v1\_locked} @ 08-10;
\texttt{r3\_design\_v2\_greedy} @ 08-19 & cc+sam & the D ≥ k rule of thumb; the ``16 + 20 + 8''
block budget; the greedy reallocation of the last round \\
protocol & scFv \texttt{protocol\_r2} @ 08-12 & sam+bighat+cc & the load and ramp perturbations
agreed three ways \\
analysis & \texttt{multistate\_thermal\_model} @ 08-10 (both); VHH \texttt{biophysics\_retro\_r0r2}
@ 08-08; \texttt{sec\_vs\_nanodsf}, \texttt{full\_panel\_learnability} @ 07-30 &
cc(autoresearch)\allowbreak{}+sam; cc+sam & co-signed the thermal model and the learnability and SEC studies \\
decisions & \texttt{lead\_decision} @ 09-04 (both) & cc (recommendation); samuelstanton (accepted) &
``the ranking is cc's; the decision is Sam's'' \\
\bottomrule\noalign{}
\end{tabular}
\end{table}

\textbf{G.3 What the record does not show.} Four corrections were habitual and are the researcher's
account, not the record's. The agent was often hasty in analysis and modeling, and the researcher pressed
it to double-check its work; the checks often found mistakes, and finding them made him press more.
He routinely prompted it to rely less on models and to reason about the biophysics directly. He
usually had to steer it away from spending wells on replicates, and had to prod it to fill leftover
wells with more creative exploratory designs. The journal records what was written down; a redirect
inside a session that produced no record is invisible here, so Figure~\ref{fig:g1} and Table~\ref{tab:g1} are a lower
bound on the steering. In the researcher's judgment the two contributions were roughly even and the result
would have been substantially worse with either party removed. The record shows a narrower division: the agent proposed most of the analyses and designs, and the researcher set the design strategy, managed the trade-off between exploration and exploitation, and decided with BigHat how the experiments were run. Without that direction the agent tended toward fitting models; the move to reasoning about the biophysics from the raw
data was a direction the researcher set.

\section{Appendix H. Outcomes in full}\label{appendix-h.-outcomes-in-full}

\setcounter{table}{0}
\renewcommand{\thetable}{H\arabic{table}}
\setcounter{figure}{0}
\renewcommand{\thefigure}{H\arabic{figure}}

\begin{figure}[t]
\centering
\includegraphics[width=1\linewidth,keepaspectratio,alt={Binding sensor signal against time for parent and lead. (a) VHH parent and lead in the final round. (b) scFv parent in round 1 and lead in the final round. Both leads dissociate much more slowly than their parents.}]{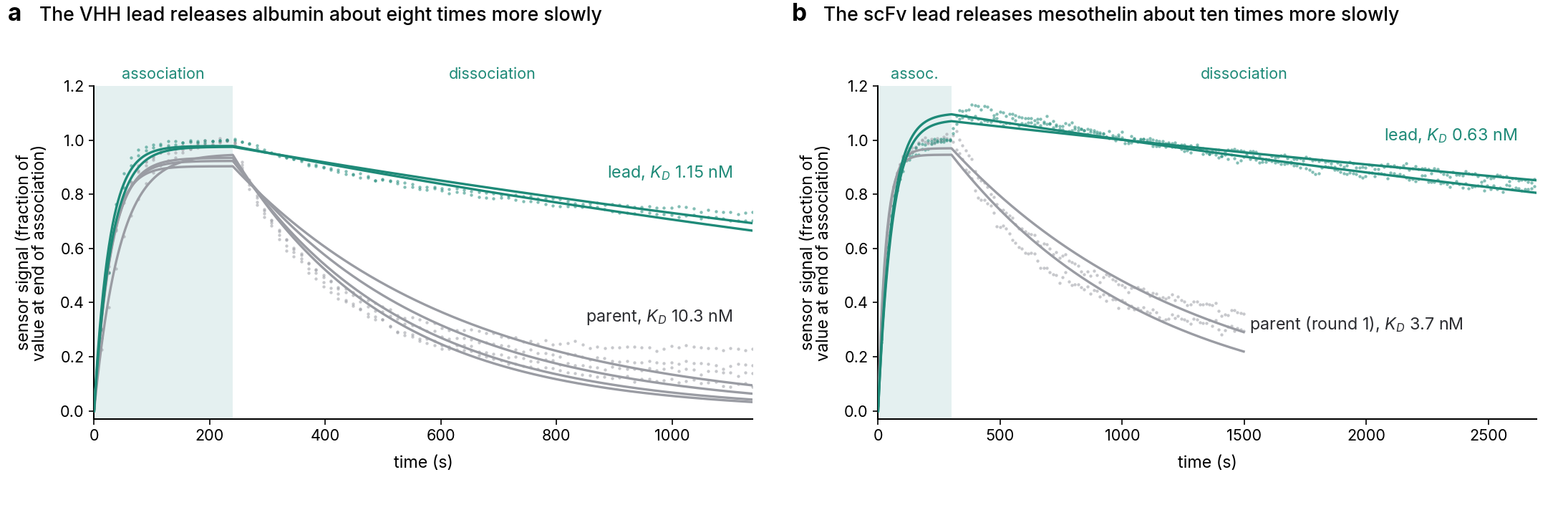}
\caption{\textbf{Binding curves of the parents and leads.} Sensor signal in the binding assay at a single 100 nM injection, scaled to each well's value at the end of association. Points are the raw signal from each well shown and lines are the platform's 1:1 fits. (a) The VHH parent (four wells) and lead (two wells) in the final-round run, with an 899 s dissociation window. (b) The scFv parent in round 1 (the two wells that gave a signal, 1,200 s window) and the lead in the final round (two wells, 2,400 s window). Each $K_D$ is the median over the wells shown.}\label{fig:sensorgrams}
\end{figure}

\begin{figure}[t]
\centering
\includegraphics[width=1\linewidth,height=\textheight,keepaspectratio,alt={Every design by round. (a, b) Off-rate on the VHH and scFv. (c) VHH aggregation rate. (d) scFv collapse temperature.}]{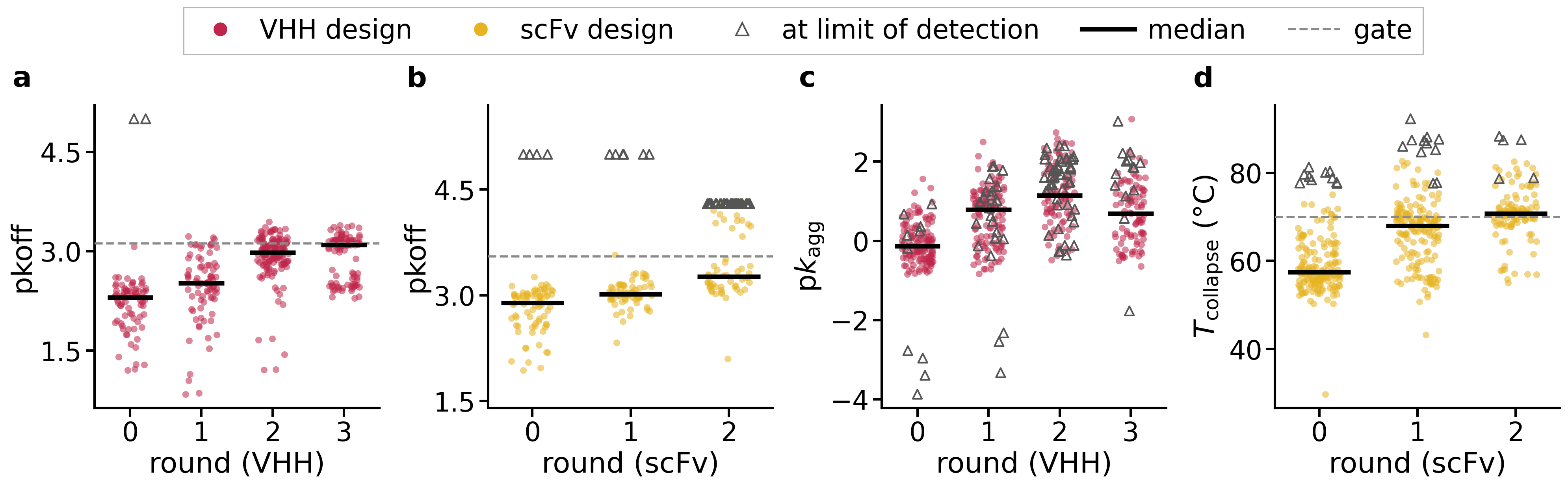}
\caption{\textbf{Every design by round.} Off-rate on (a) the VHH and (b) the scFv, (c) the VHH aggregation metric \(\mathrm{p}k_{\mathrm{agg}}\), and (d) the scFv \(T_{\mathrm{collapse}}\). Bars are medians; triangles are at the limit of detection; dashed lines are the constraints, and panel (c) does not draw the revised \(\mathrm{p}k_{\mathrm{agg}} \geq 1.72\).}\label{fig:population}
\end{figure}

\textbf{H.1 Summary.} Both campaigns ended with a lead that improves on its parent on affinity, thermal stability and
aggregation and holds expression and purity (Tables~\ref{tab:h1} and~\ref{tab:h2}, Figure~\ref{fig:sensorgrams}; scFv HMW 2.3\% against 4.2\% for the parent in DR1 and 2.3\% in DR0), and both populations moved as blocks across rounds (Figure~\ref{fig:population}). BigHat agreed that both campaigns were completed successfully. n is given for every lead value; bounds are written as bounds.

\textbf{H.2 The VHH lead.} The VHH lead, CAND\_L45P\_S31R, carries eight mutations on Alb8: one affinity lever that also slows aggregation (E1A), three
melt levers (N16G, T77N, L45P), two rim-charge aggregation levers (S30R, V93R), a further rim charge with little measured effect on aggregation (S103R), and one
interface residue from a six-well scan (S31R). On the DR3 run its off-rate is 0.90 log slower than
the parent's ($\mathrm{p}k_{\mathrm{off}}$ 3.39 on two wells against 2.49 on four), 1.15 nM raw $K_D$; the identified melt is
70.6 °C (parent 62.2), 0.6 °C above the constraint, about one replicate SD; the aggregation metric is
+2.00 on one identified well against −0.39 on the parent; HMW is not detected; yield is 51 µg/mL. Two
caveats: the aggregation value is one well, and the $K_D$ margin, 0.14 log, is smaller than the 0.20
log ambiguity in the run reference between DR2 and DR3.

\begin{table}[t]
\centering
\footnotesize
\caption{The VHH lead against its parent (DR3 run), with the original specifications and the
campaign's revised specifications.\label{tab:h1}}
\begin{tabular}{@{}
  >{\raggedright\arraybackslash}p{(\linewidth - 8\tabcolsep) * \real{0.1842}}
  >{\raggedright\arraybackslash}p{(\linewidth - 8\tabcolsep) * \real{0.1579}}
  >{\raggedright\arraybackslash}p{(\linewidth - 8\tabcolsep) * \real{0.2105}}
  >{\raggedright\arraybackslash}p{(\linewidth - 8\tabcolsep) * \real{0.1842}}
  >{\raggedright\arraybackslash}p{(\linewidth - 8\tabcolsep) * \real{0.2632}}@{}}
\toprule\noalign{}
\begin{minipage}[b]{\linewidth}\raggedright
metric
\end{minipage} & \begin{minipage}[b]{\linewidth}\raggedright
parent Alb8 (n)
\end{minipage} & \begin{minipage}[b]{\linewidth}\raggedright
lead CAND\_L45P\_S31R (n)
\end{minipage} & \begin{minipage}[b]{\linewidth}\raggedright
original constraint
\end{minipage} & \begin{minipage}[b]{\linewidth}\raggedright
revised constraint
\end{minipage} \\
\midrule\noalign{}
off-rate, $\mathrm{p}k_{\mathrm{off}}$ & 2.49 (4) & 3.39 (2) & ≥ 3.12 ($K_D$ ≤ 1.6 nM at the reference on-rate) & same \\
$K_D$, raw & 10.3 nM & 1.15 nM & ≤ 1.6 nM & same \\
melt, identified $\hat{T}_m$ (°C) & 62.2 (4) & 70.6 (2) & ≥ 70 on reported $T_m^{\mathrm{app}}$ & ≥ 70 on $\hat{T}_m$ \\
reported $T_{\mathrm{agg}}$ (°C) & 40--45 (4 per round) & no onset to 85 (2) & ≥ 70 & replaced by \(\mathrm{p}k_{\mathrm{agg}}\) ≥
chassis value and HMW ≤ 5\% \\
aggregation, \(\mathrm{p}k_{\mathrm{agg}}\) & −0.39 (4) & +2.00 (1) & none & ≥ +1.72 (the chassis's
value, 3 wells) \\
SEC HMW / recovery & 2.8\% / 70\% (4) & not detected / 57\% (2) & information & ≤ 5\% \\
yield (µg/mL) & 43 & 51 (2) & \textgreater{} 10 & same \\
\bottomrule\noalign{}
\end{tabular}
\end{table}

\textbf{H.3 The scFv lead.} The scFv lead, P\_dose\_010A, is P (the parent plus Q91A, G230Y, Y238A and S245A) plus two stability
residues from the DR2 dose cube, T187Y and V196A. P itself is beyond the assay floor on all nine of
its DR2 wells, so its affinity is a bound; the lead is measured, at $\mathrm{p}k_{\mathrm{off}}$ 3.96 on two wells ($K_D$ 0.63
nM) against 2.95 on the parent (3.00 in DR0), and was chosen over P because it is measured.
P has the better affinity bound, two other
designs have better thermal metrics, and T187Y costs 0.32 log of off-rate for no resolved gain in
collapse. Its collapse temperature is 74.2 °C against 56.3 on the parent (57.5 in DR0) and 2.3 °C above P (inside
1.2 replicate SDs); it passes the reported $T_m^{\mathrm{app}}$ constraint and fails the reported $T_{\mathrm{agg}}$ constraint, as does every
shortlisted design. Every lead value rests on two wells.

\begin{table}[t]
\centering
\footnotesize
\caption{The scFv lead against P and the parent (DR2 run; parent values from its DR1 and DR0 runs, in that order).\label{tab:h2}}
\begin{tabular}{@{}
  >{\raggedright\arraybackslash}p{(\linewidth - 10\tabcolsep) * \real{0.1625}}
  >{\raggedright\arraybackslash}p{(\linewidth - 10\tabcolsep) * \real{0.1250}}
  >{\raggedright\arraybackslash}p{(\linewidth - 10\tabcolsep) * \real{0.1625}}
  >{\raggedright\arraybackslash}p{(\linewidth - 10\tabcolsep) * \real{0.1750}}
  >{\raggedright\arraybackslash}p{(\linewidth - 10\tabcolsep) * \real{0.1750}}
  >{\raggedright\arraybackslash}p{(\linewidth - 10\tabcolsep) * \real{0.2000}}@{}}
\toprule\noalign{}
\begin{minipage}[b]{\linewidth}\raggedright
metric
\end{minipage} & \begin{minipage}[b]{\linewidth}\raggedright
parent, DR1 / DR0 (n)
\end{minipage} & \begin{minipage}[b]{\linewidth}\raggedright
P (n)
\end{minipage} & \begin{minipage}[b]{\linewidth}\raggedright
lead P\_dose\_010A (n)
\end{minipage} & \begin{minipage}[b]{\linewidth}\raggedright
original constraint
\end{minipage} & \begin{minipage}[b]{\linewidth}\raggedright
revised constraint
\end{minipage} \\
\midrule\noalign{}
off-rate, $\mathrm{p}k_{\mathrm{off}}$ & 2.95 / 3.00 (4) & ≥ 4.3, bound (9) & 3.96 (2) & ≥ 3.55 ($K_D$ ≤ 1.5 nM) & $K_D$ ≤ 1 nM,
quantified \\
$K_D$ & 3.7 / 4.3 nM & ≤ 0.25 nM, bound & 0.63 nM & ≤ 1.5 nM & ≤ 1 nM \\
collapse, \(T_{\mathrm{collapse}}\) (°C) & 56.3 / 57.5 (4) & 71.9 (9) & 74.2 (2) & none & ≥ 70 \\
reported $T_m^{\mathrm{app}}$ (°C) & 54.6 / 55.0 & 71.5 & 75.1 & ≥ 65 & replaced by collapse \\
reported $T_{\mathrm{agg}}$ (°C) & 50.5 / 48.3 & 61.3 & 57.7 & ≥ 65 & replaced by collapse \\
SEC HMW / recovery & 4.2\% / 59\%; 2.3\% / 71\% & 2.4\% / 76\% (9) & 2.3\% / 74\% (2) & information & ≤ 3\% \\
yield (µg/mL) & 16 / 39 & 30 & 41 & \textgreater{} 10 & same \\
\bottomrule\noalign{}
\end{tabular}
\end{table}

\textbf{H.4 Populations.} Figure~\ref{fig:population} shows every design in every round on the primary metrics. On the VHH the binder rate went
48 → 58 → 93\% and the median off-rate 2.30 → 2.52 → 2.98 over DR0 to DR2 (n = 178, 173, 173), with
the median aggregation metric among unbounded designs rising from −0.14 to +1.15. On the scFv the binder rate went 47 → 44
→ 93\%, the median off-rate among quantified binders 2.89 → 3.02 → 3.26 with 4, 6 and 51 designs
beyond the dissociation floor, and the median collapse temperature 57 → 68 → 71 °C (n = 176, 169,
113).

\section{Appendix I. Limits in full}\label{appendix-i.-limits-in-full}

\setcounter{figure}{0}
\renewcommand{\thefigure}{I\arabic{figure}}

These are the limits the lead decisions record, with the reviews' corrections, and the limits of the report's argument.

\begin{figure}[t]
\centering
\includegraphics[width=1\linewidth,height=\textheight,keepaspectratio,alt={The off-rate floor by round. Measured off-rates for every quantified binder by round on (a) the VHH and (b) the scFv, with the floor above which an off-rate is recorded as a bound: the fitter's own limit (pkoff 5.0) in the early rounds, the campaign's reported floor for the scFv DR2 2400 s window (4.3; the 5\% rule alone would put it at 4.67), and an 899 s window (4.24) for VHH DR3. Triangles are the bounds. (c) The fraction of binding signal lost over the window as a function of off-rate, for the two windows used; the 5\% crossing is at 4.24 for 899 s and 4.67 for 2400 s.}]{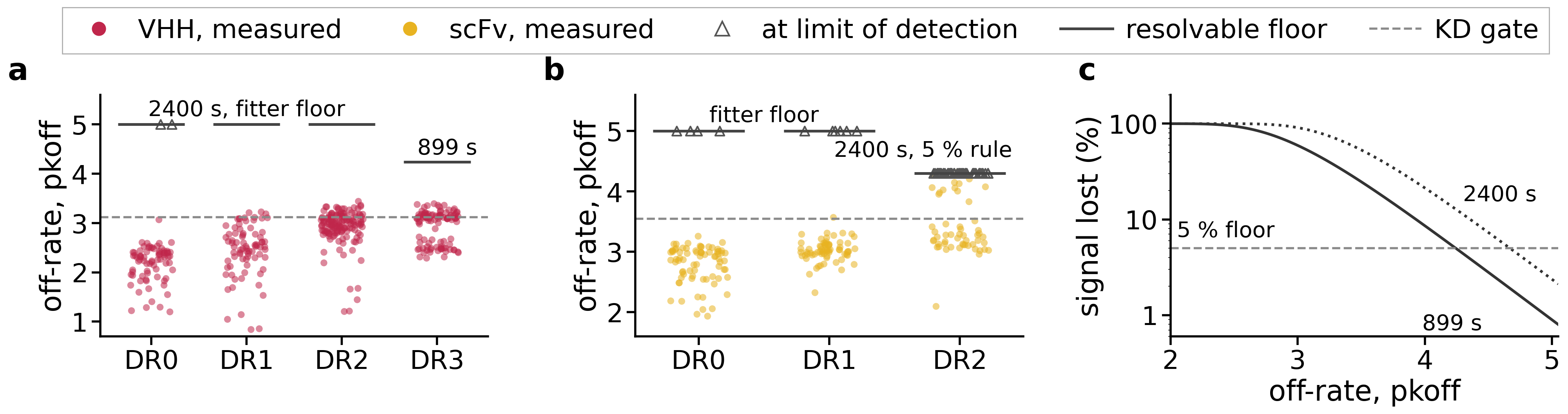}
\caption{\textbf{The off-rate floor by round.} Measured off-rates for every quantified binder by
round on (a) the VHH and (b) the scFv, with the floor above which an off-rate is recorded as a
bound: the fitter's own limit ($\mathrm{p}k_{\mathrm{off}}$ 5.0) in the early rounds, the campaign's reported floor for the
scFv DR2 2400 s window (4.3; the 5\% rule alone would put it at 4.67), and an 899 s window (4.24)
for VHH DR3. Triangles are the bounds. (c) The fraction of binding signal lost over the window as a
function of off-rate, for the two windows used; the 5\% crossing is at 4.24 for 899 s and 4.67 for
2400 s.}\label{fig:i1}
\end{figure}

\textbf{I.1 The assay floor.} A single 100 nM BLI injection with a 1:1 fit resolves an off-rate only
if the signal falls by more than about 5\% over the dissociation window; slower off-rates are
recorded as a bound at the window's floor (Figure~\ref{fig:i1}c).  The VHH's window was 899 s in every round, so off-rates slower than $\mathrm{p}k_{\mathrm{off}}$ 4.24 are
unresolved; no measured DR3 design exceeds 3.39, and the lead's $K_D$ margin over the constraint is 0.14 log
(Figure~\ref{fig:i1}a). BigHat offered a longer dissociation window in July, and the 2400 s window in the
protocol record of 2026-08-12 was run on the scFv's final round. There 51 of 113 designs, P among
them, are at its floor, reported as $\mathrm{p}k_{\mathrm{off}}$ 4.3 (the 5\% rule alone would give 4.67; the campaign's
value is the more conservative bound of record); their ordering replicates only weakly across plates
(over the 17 designs at the floor on both plates, Spearman +0.23 on the measured signal drop and +0.53 on the fitter's extrapolated off-rates), and the lead was chosen over P because it is measured (Figure~\ref{fig:i1}b). The earlier
rounds' floor of 5.0 is the Octet fitter's own limit, not a window rule. A single long-window plate
with the shortlists and parents on it would settle both rankings. The VHH DR3 run also carries an
unresolved offset from DR2, under which the lead's $K_D$ is 1.0, 1.15 or 1.6 nM depending on the
reference, and plate G reads about 0.05 $\mathrm{p}k_{\mathrm{off}}$ below plate H; offsets were estimated and reported,
not corrected.

\textbf{I.2 Thermal and replicate limits.} The two ramp rates ran on different days in both
campaigns, with a day term of +5 to +7 °C on scattering onset that the ramp-rate shifts cannot be
separated from (Appendix E). The VHH lead's aggregation value is one well; every scFv lead value
rests on two, and its thermal advantage over P is inside noise. The VHH $T_{\mathrm{agg}}$ constraint is passed on our reading of the raw curves: the lead's scattering stays near baseline, so the onset model has nothing to fit, and we read $T_{\mathrm{agg}}$ as right-censored above 85 °C (Appendix D.3). The scFv collapse temperature is standardized with one pooled
concentration slope across designs whose loadings differ by more than 3 e-folds, where second-order
kinetics predict a design-dependent slope.

\textbf{I.3 Next experiments.} The campaigns measured cell-free material on the standard assay set.
The next experiments are stress testing (hold, freeze-thaw and heat), HIC, a polyreactivity panel, DSC, charge-variant analysis and expression in a mammalian system.
Polyreactivity is of particular interest for the VHH lead, since its mutations add about five positive charges (E1A, S30R, S31R, V93R, S103R) and positive charge can raise nonspecific binding \citep{tiller2017}, and a condition that resolves the scFv's true unfolding midpoint would complete its thermal picture.
The parent VHH is on an older codon set than the designs, so a matched comparison would use one codon set.
Confirmation runs of both leads would test decisions made on the data as measured.

\textbf{I.4 Limits of the argument.} The comparison between designed blocks and learned acquisition
(Appendix F) is within one campaign. We ran no comparison library designed another way, so we cannot say how much of the gain came from the selection. The division of labor (Section 8, Appendix G)
rests on a journal that records what was written down, and the researcher's habitual corrections are his
account. And two scaffolds on one platform with one partner is the whole sample.


\clearpage
\begin{thebibliography}{99}
\bibitem[Ablynx(2006)]{ablynx2006} Ablynx N.V. Serum albumin binding proteins. WO 2006/122787 A1 (2006).
\bibitem[Ablynx(2012)]{ablynx2012} Ablynx N.V. Serum albumin binding proteins. WO 2012/175400 A1 (2012).
\bibitem[Andrews and Roberts(2007)]{andrews2007} Andrews, J. M. \& Roberts, C. J. A Lumry--Eyring nucleated polymerization model of protein aggregation kinetics: 1. Aggregation with pre-equilibrated unfolding. \emph{Journal of Physical Chemistry B} \textbf{111}, 7897--7913 (2007). \href{https://doi.org/10.1021/jp070212j}{doi:10.1021/jp070212j}
\bibitem[Bisht et~al.(2026)]{bisht2026} Bisht, H., Kumar, V., Jablonka, K. M., Mausam \& Krishnan, N. M. A. Agentic AI scientists are not built for autonomous scientific discovery. \emph{NeurIPS Position Paper Track} (2026). \href{https://arxiv.org/abs/2605.08956}{arXiv:2605.08956}
\bibitem[ByteDance AML AI4Science Team(2025)]{protenix2025} ByteDance AML AI4Science Team. Protenix: Advancing structure prediction through a comprehensive AlphaFold3 reproduction. \emph{bioRxiv} (2025). \href{https://doi.org/10.1101/2025.01.08.631967}{doi:10.1101/2025.01.08.631967}
\bibitem[Candido et~al.(2026)]{candido2026} Candido, S. \emph{et al.} Language modeling materializes a world model of protein biology. \emph{bioRxiv} (2026). \href{https://doi.org/10.64898/2026.06.03.729735}{doi:10.64898/2026.06.03.729735}
\bibitem[Carter and Lazar(2018)]{carter2018} Carter, P. J. \& Lazar, G. A. Next generation antibody drugs: pursuit of the `high-hanging fruit'. \emph{Nature Reviews Drug Discovery} \textbf{17}, 197--223 (2018). \href{https://doi.org/10.1038/nrd.2017.227}{doi:10.1038/nrd.2017.227}
\bibitem[Chen et~al.(2025)]{chen2024} Chen, A. \emph{et al.} Generalists vs. Specialists: Evaluating LLMs on highly-constrained biophysical sequence optimization tasks. In \emph{Proceedings of the 42nd International Conference on Machine Learning}, \emph{PMLR} \textbf{267}, 9029--9072 (2025). \href{https://arxiv.org/abs/2410.22296}{arXiv:2410.22296}
\bibitem[Chi et~al.(2003)]{chi2003} Chi, E. Y., Krishnan, S., Randolph, T. W. \& Carpenter, J. F. Physical stability of proteins in aqueous solution: mechanism and driving forces in nonnative protein aggregation. \emph{Pharmaceutical Research} \textbf{20}, 1325--1336 (2003). \href{https://doi.org/10.1023/A:1025771421906}{doi:10.1023/A:1025771421906}
\bibitem[Claude Science and Shanehsazzadeh(2026)]{shaneh2026} Claude Science \& Shanehsazzadeh, A. Autonomous \emph{de novo} protein binder design with Claude. \emph{Anthropic technical report} (18 August 2026).
\bibitem[Dauparas et~al.(2022)]{dauparas2022} Dauparas, J. \emph{et al.} Robust deep learning-based protein sequence design using ProteinMPNN. \emph{Science} \textbf{378}, 49--56 (2022). \href{https://doi.org/10.1126/science.add2187}{doi:10.1126/science.add2187}
\bibitem[De~Felice et~al.(2024)]{defelice2023} De Felice, S. \emph{et al.} Crystal structure of human serum albumin in complex with megabody reveals unique human and murine cross-reactive binding site. \emph{Protein Science} \textbf{33}, e4887 (2024). \href{https://doi.org/10.1002/pro.4887}{doi:10.1002/pro.4887}
\bibitem[Dudgeon et~al.(2012)]{dudgeon2012} Dudgeon, K. \emph{et al.} General strategy for the generation of human antibody variable domains with increased aggregation resistance. \emph{Proceedings of the National Academy of Sciences} \textbf{109}, 10879--10884 (2012). \href{https://doi.org/10.1073/pnas.1202866109}{doi:10.1073/pnas.1202866109}
\bibitem[Dunbar and Deane(2016)]{dunbar2016} Dunbar, J. \& Deane, C. M. ANARCI: antigen receptor numbering and receptor classification. \emph{Bioinformatics} \textbf{32}, 298--300 (2016). \href{https://doi.org/10.1093/bioinformatics/btv552}{doi:10.1093/bioinformatics/btv552}
\bibitem[Ebo et~al.(2020)]{ebo2020} Ebo, J. S. \emph{et al.} An in vivo platform to select and evolve aggregation-resistant proteins. \emph{Nature Communications} \textbf{11}, 1816 (2020). \href{https://doi.org/10.1038/s41467-020-15667-1}{doi:10.1038/s41467-020-15667-1}
\bibitem[Evans et~al.(2021)]{evans2021} Evans, R. \emph{et al.} Protein complex prediction with AlphaFold-Multimer. \emph{bioRxiv} (2021). \href{https://doi.org/10.1101/2021.10.04.463034}{doi:10.1101/2021.10.04.463034}
\bibitem[Fannjiang and Listgarten(2020)]{fannjiang2020} Fannjiang, C. \& Listgarten, J. Autofocused oracles for model-based design. In \emph{Advances in Neural Information Processing Systems} \textbf{33} (2020). \href{https://arxiv.org/abs/2006.08052}{arXiv:2006.08052}
\bibitem[Fleming and DeMets(1996)]{fleming1996} Fleming, T. R. \& DeMets, D. L. Surrogate end points in clinical trials: are we being misled? \emph{Annals of Internal Medicine} \textbf{125}, 605--613 (1996). \href{https://doi.org/10.7326/0003-4819-125-7-199610010-00011}{doi:10.7326/0003-4819-125-7-199610010-00011}
\bibitem[Foote and Winter(1992)]{foote1992} Foote, J. \& Winter, G. Antibody framework residues affecting the conformation of the hypervariable loops. \emph{Journal of Molecular Biology} \textbf{224}, 487--499 (1992). \href{https://doi.org/10.1016/0022-2836(92)91010-M}{doi:10.1016/0022-2836(92)91010-M}
\bibitem[Frey et~al.(2025)]{frey2025} Frey, N. C. \emph{et al.} Lab-in-the-loop therapeutic antibody design with deep learning. \emph{bioRxiv} (2025). \href{https://doi.org/10.1101/2025.02.19.639050}{doi:10.1101/2025.02.19.639050}
\bibitem[Gao et~al.(2023)]{gao2023} Gao, L., Schulman, J. \& Hilton, J. Scaling laws for reward model overoptimization. In \emph{Proceedings of the 40th International Conference on Machine Learning}, \emph{PMLR} \textbf{202}, 10835--10866 (2023). \href{https://arxiv.org/abs/2210.10760}{arXiv:2210.10760}
\bibitem[Gottweis et~al.(2026)]{gottweis2026} Gottweis, J. \emph{et al.} Accelerating scientific discovery with Co-Scientist. \emph{Nature} \textbf{655}, 487--496 (2026). \href{https://doi.org/10.1038/s41586-026-10644-y}{doi:10.1038/s41586-026-10644-y}
\bibitem[Goyal et~al.(2014)]{goyal2014} Goyal, M., Chaudhuri, T. K. \& Kuwajima, K. Irreversible denaturation of maltodextrin glucosidase studied by differential scanning calorimetry, circular dichroism, and turbidity measurements. \emph{PLOS ONE} \textbf{9}, e115877 (2014). \href{https://doi.org/10.1371/journal.pone.0115877}{doi:10.1371/journal.pone.0115877}
\bibitem[Greenfield(2006)]{greenfield2006} Greenfield, N. J. Using circular dichroism collected as a function of temperature to determine the thermodynamics of protein unfolding and binding interactions. \emph{Nature Protocols} \textbf{1}, 2527--2535 (2006). \href{https://doi.org/10.1038/nprot.2006.204}{doi:10.1038/nprot.2006.204}
\bibitem[Gruver et~al.(2023)]{gruver2023} Gruver, N. \emph{et al.} Protein design with guided discrete diffusion. In \emph{Advances in Neural Information Processing Systems} \textbf{36} (2023). \href{https://arxiv.org/abs/2305.20009}{arXiv:2305.20009}
\bibitem[Hie et~al.(2024)]{hie2023} Hie, B. L. \emph{et al.} Efficient evolution of human antibodies from general protein language models. \emph{Nature Biotechnology} \textbf{42}, 275--283 (2024). \href{https://doi.org/10.1038/s41587-023-01763-2}{doi:10.1038/s41587-023-01763-2}
\bibitem[Huang et~al.(2026)]{biomni2026} Huang, K. \emph{et al.} Autonomous biomedical research with an artificial intelligence agent. \emph{Science} \textbf{393}, eadz4351 (2026). \href{https://doi.org/10.1126/science.adz4351}{doi:10.1126/science.adz4351}
\bibitem[J\"ager and Pl\"uckthun(1999)]{jager1999} J\"ager, M. \& Pl\"uckthun, A. Domain interactions in antibody Fv and scFv fragments: effects on unfolding kinetics and equilibria. \emph{FEBS Letters} \textbf{462}, 307--312 (1999). \href{https://doi.org/10.1016/S0014-5793(99)01532-X}{doi:10.1016/S0014-5793(99)01532-X}
\bibitem[Jain et~al.(2017)]{jain2017} Jain, T. \emph{et al.} Biophysical properties of the clinical-stage antibody landscape. \emph{Proceedings of the National Academy of Sciences} \textbf{114}, 944--949 (2017). \href{https://doi.org/10.1073/pnas.1616408114}{doi:10.1073/pnas.1616408114}
\bibitem[Jeon et~al.(2026)]{jeon2026} Jeon, M., Song, J., Kim, J. \& Zhong, E. D. Agentic campaign control for high-throughput \emph{de novo} binder design. \emph{bioRxiv} (2026). \href{https://doi.org/10.64898/2026.09.22.753604}{doi:10.64898/2026.09.22.753604}
\bibitem[Jumper et~al.(2021)]{jumper2021} Jumper, J. \emph{et al.} Highly accurate protein structure prediction with AlphaFold. \emph{Nature} \textbf{596}, 583--589 (2021). \href{https://doi.org/10.1038/s41586-021-03819-2}{doi:10.1038/s41586-021-03819-2}
\bibitem[Kenlay et~al.(2024)]{kenlay2024} Kenlay, H. \emph{et al.} Large scale paired antibody language models. \emph{PLOS Computational Biology} \textbf{20}, e1012646 (2024). \href{https://doi.org/10.1371/journal.pcbi.1012646}{doi:10.1371/journal.pcbi.1012646}
\bibitem[Khan et~al.(2023)]{khan2023} Khan, A. \emph{et al.} Toward real-world automated antibody design with combinatorial Bayesian optimization. \emph{Cell Reports Methods} \textbf{3}, 100374 (2023). \href{https://doi.org/10.1016/j.crmeth.2022.100374}{doi:10.1016/j.crmeth.2022.100374}
\bibitem[Kotov et~al.(2021)]{kotov2021} Kotov, V. \emph{et al.} In-depth interrogation of protein thermal unfolding data with MoltenProt. \emph{Protein Science} \textbf{30}, 201--217 (2021). \href{https://doi.org/10.1002/pro.3986}{doi:10.1002/pro.3986}
\bibitem[Lepock et~al.(1992)]{lepock1992} Lepock, J. R. \emph{et al.} Influence of transition rates and scan rate on kinetic simulations of differential scanning calorimetry profiles of reversible and irreversible protein denaturation. \emph{Biochemistry} \textbf{31}, 12706--12712 (1992). \href{https://doi.org/10.1021/bi00165a023}{doi:10.1021/bi00165a023}
\bibitem[Li et~al.(2023)]{li2023} Li, L. \emph{et al.} Machine learning optimization of candidate antibody yields highly diverse sub-nanomolar affinity antibody libraries. \emph{Nature Communications} \textbf{14}, 3454 (2023). \href{https://doi.org/10.1038/s41467-023-39022-2}{doi:10.1038/s41467-023-39022-2}
\bibitem[Lu et~al.(2026)]{lu2026} Lu, C. \emph{et al.} Towards end-to-end automation of AI research. \emph{Nature} \textbf{651}, 914--919 (2026). \href{https://doi.org/10.1038/s41586-026-10265-5}{doi:10.1038/s41586-026-10265-5}
\bibitem[Lumry and Eyring(1954)]{lumry1954} Lumry, R. \& Eyring, H. Conformation changes of proteins. \emph{Journal of Physical Chemistry} \textbf{58}, 110--120 (1954). \href{https://doi.org/10.1021/j150512a005}{doi:10.1021/j150512a005}
\bibitem[Lundberg and Lee(2017)]{lundberg2017} Lundberg, S. M. \& Lee, S.-I. A unified approach to interpreting model predictions. In \emph{Advances in Neural Information Processing Systems} \textbf{30} (2017). \href{https://arxiv.org/abs/1705.07874}{arXiv:1705.07874}
\bibitem[MacKay(1992)]{mackay1992} MacKay, D. J. C. Bayesian interpolation. \emph{Neural Computation} \textbf{4}, 415--447 (1992). \href{https://doi.org/10.1162/neco.1992.4.3.415}{doi:10.1162/neco.1992.4.3.415}
\bibitem[Makabe et~al.(2008)]{makabe2008} Makabe, K. \emph{et al.} Thermodynamic consequences of mutations in vernier zone residues of a humanized anti-human epidermal growth factor receptor murine antibody, 528. \emph{Journal of Biological Chemistry} \textbf{283}, 1156--1166 (2008). \href{https://doi.org/10.1074/jbc.M706190200}{doi:10.1074/jbc.M706190200}
\bibitem[Makowski et~al.(2022)]{makowski2022} Makowski, E. K. \emph{et al.} Co-optimization of therapeutic antibody affinity and specificity using machine learning models that generalize to novel mutational space. \emph{Nature Communications} \textbf{13}, 3788 (2022). \href{https://doi.org/10.1038/s41467-022-31457-3}{doi:10.1038/s41467-022-31457-3}
\bibitem[Manheim and Garrabrant(2018)]{manheim2018} Manheim, D. \& Garrabrant, S. Categorizing variants of Goodhart's law. \emph{arXiv} (2018). \href{https://arxiv.org/abs/1803.04585}{arXiv:1803.04585}
\bibitem[Mason et~al.(2021)]{mason2021} Mason, D. M. \emph{et al.} Optimization of therapeutic antibodies by predicting antigen specificity from antibody sequence via deep learning. \emph{Nature Biomedical Engineering} \textbf{5}, 600--612 (2021). \href{https://doi.org/10.1038/s41551-021-00699-9}{doi:10.1038/s41551-021-00699-9}
\bibitem[Medarex(2009)]{medarex} Medarex, Inc. Human antibodies that bind mesothelin, and uses thereof. WO 2009/045957 A1 (2009).
\bibitem[Mima et~al.(2024)]{mima2024} Mima, M. \emph{et al.} Structural design of the anti-TNF\(\alpha\) therapeutic NANOBODY compound, ozoralizumab, to support its potent and sustained clinical efficacy. \emph{Biochemical and Biophysical Research Communications} \textbf{734}, 150454 (2024). \href{https://doi.org/10.1016/j.bbrc.2024.150454}{doi:10.1016/j.bbrc.2024.150454}
\bibitem[Owen(1998)]{owen1998} Owen, A. B. Scrambling Sobol' and Niederreiter--Xing points. \emph{Journal of Complexity} \textbf{14}, 466--489 (1998). \href{https://doi.org/10.1006/jcom.1998.0487}{doi:10.1006/jcom.1998.0487}
\bibitem[Park et~al.(2024)]{park2023} Park, J. W., Tagasovska, N., Maser, M., Ra, S. \& Cho, K. BOtied: Multi-objective Bayesian optimization with tied multivariate ranks. In \emph{Proceedings of the 41st International Conference on Machine Learning}, \emph{PMLR} \textbf{235}, 39813--39833 (2024). \href{https://arxiv.org/abs/2306.00344}{arXiv:2306.00344}
\bibitem[Passaro et~al.(2025)]{boltz2} Passaro, S. \emph{et al.} Boltz-2: Towards accurate and efficient binding affinity prediction. \emph{bioRxiv} (2025). \href{https://doi.org/10.1101/2025.06.14.659707}{doi:10.1101/2025.06.14.659707}
\bibitem[Perchiacca et~al.(2014)]{perchiacca2014} Perchiacca, J. M., Lee, C. C. \& Tessier, P. M. Optimal charged mutations in the complementarity-determining regions that prevent domain antibody aggregation are dependent on the antibody scaffold. \emph{Protein Engineering, Design and Selection} \textbf{27}, 29--39 (2014). \href{https://doi.org/10.1093/protein/gzt058}{doi:10.1093/protein/gzt058}
\bibitem[Poelwijk et~al.(2016)]{poelwijk2016} Poelwijk, F. J., Krishna, V. \& Ranganathan, R. The context-dependence of mutations: a linkage of formalisms. \emph{PLOS Computational Biology} \textbf{12}, e1004771 (2016). \href{https://doi.org/10.1371/journal.pcbi.1004771}{doi:10.1371/journal.pcbi.1004771}
\bibitem[Poelwijk et~al.(2019)]{poelwijk2019} Poelwijk, F. J., Socolich, M. \& Ranganathan, R. Learning the pattern of epistasis linking genotype and phenotype in a protein. \emph{Nature Communications} \textbf{10}, 4213 (2019). \href{https://doi.org/10.1038/s41467-019-12130-8}{doi:10.1038/s41467-019-12130-8}
\bibitem[Ramon et~al.(2025)]{ramon2025} Ramon, A. \emph{et al.} Prediction of protein biophysical traits from limited data: A case study on nanobody thermostability through NanoMelt. \emph{mAbs} \textbf{17}, 2442750 (2025). \href{https://doi.org/10.1080/19420862.2024.2442750}{doi:10.1080/19420862.2024.2442750}
\bibitem[Rapp et~al.(2024)]{rapp2024} Rapp, J. T., Bremer, B. J. \& Romero, P. A. Self-driving laboratories to autonomously navigate the protein fitness landscape. \emph{Nature Chemical Engineering} \textbf{1}, 97--107 (2024). \href{https://doi.org/10.1038/s44286-023-00002-4}{doi:10.1038/s44286-023-00002-4}
\bibitem[Raybould et~al.(2019)]{raybould2019} Raybould, M. I. J. \emph{et al.} Five computational developability guidelines for therapeutic antibody profiling. \emph{Proceedings of the National Academy of Sciences} \textbf{116}, 4025--4030 (2019). \href{https://doi.org/10.1073/pnas.1810576116}{doi:10.1073/pnas.1810576116}
\bibitem[Rollins et~al.(2024)]{rollins2024} Rollins, Z. A., Widatalla, T., Cheng, A. C. \& Metwally, E. AbMelt: Learning antibody thermostability from molecular dynamics. \emph{Biophysical Journal} \textbf{123}, 2921--2933 (2024). \href{https://doi.org/10.1016/j.bpj.2024.06.003}{doi:10.1016/j.bpj.2024.06.003}
\bibitem[Romero et~al.(2013)]{romero2013} Romero, P. A., Krause, A. \& Arnold, F. H. Navigating the protein fitness landscape with Gaussian processes. \emph{Proceedings of the National Academy of Sciences} \textbf{110}, E193--E201 (2013). \href{https://doi.org/10.1073/pnas.1215251110}{doi:10.1073/pnas.1215251110}
\bibitem[S\'anchez-Ruiz et~al.(1988)]{sanchezruiz1988} S\'anchez-Ruiz, J. M., L\'opez-Lacomba, J. L., Cortijo, M. \& Mateo, P. L. Differential scanning calorimetry of the irreversible thermal denaturation of thermolysin. \emph{Biochemistry} \textbf{27}, 1648--1652 (1988). \href{https://doi.org/10.1021/bi00405a039}{doi:10.1021/bi00405a039}
\bibitem[S\'anchez-Ruiz(1992)]{sanchezruiz1992} S\'anchez-Ruiz, J. M. Theoretical analysis of Lumry-Eyring models in differential scanning calorimetry. \emph{Biophysical Journal} \textbf{61}, 921--935 (1992). \href{https://doi.org/10.1016/S0006-3495(92)81899-4}{doi:10.1016/S0006-3495(92)81899-4}
\bibitem[Schymkowitz et~al.(2005)]{schymkowitz2005} Schymkowitz, J. \emph{et al.} The FoldX web server: an online force field. \emph{Nucleic Acids Research} \textbf{33}, W382--W388 (2005). \href{https://doi.org/10.1093/nar/gki387}{doi:10.1093/nar/gki387}
\bibitem[Shapley(1953)]{shapley1953} Shapley, L. S. A value for \emph{n}-person games. In \emph{Contributions to the Theory of Games}, vol. II (eds Kuhn, H. W. \& Tucker, A. W.) 307--317 (Princeton University Press, 1953).
\bibitem[Sobol'(1967)]{sobol1967} Sobol', I. M. On the distribution of points in a cube and the approximate evaluation of integrals. \emph{USSR Computational Mathematics and Mathematical Physics} \textbf{7}, 86--112 (1967). \href{https://doi.org/10.1016/0041-5553(67)90144-9}{doi:10.1016/0041-5553(67)90144-9}
\bibitem[Stanton et~al.(2022)]{stanton2022lambo} Stanton, S. \emph{et al.} Accelerating Bayesian optimization for biological sequence design with denoising autoencoders. In \emph{Proceedings of the 39th International Conference on Machine Learning}, \emph{PMLR} \textbf{162}, 20459--20478 (2022). \href{https://arxiv.org/abs/2203.12742}{arXiv:2203.12742}
\bibitem[Swanson et~al.(2025)]{swanson2025} Swanson, K., Wu, W., Bulaong, N. L., Pak, J. E. \& Zou, J. The Virtual Lab of AI agents designs new SARS-CoV-2 nanobodies. \emph{Nature} \textbf{646}, 716--723 (2025). \href{https://doi.org/10.1038/s41586-025-09442-9}{doi:10.1038/s41586-025-09442-9}
\bibitem[Tiller et~al.(2017)]{tiller2017} Tiller, K. E. \emph{et al.} Arginine mutations in antibody complementarity-determining regions display context-dependent affinity/specificity trade-offs. \emph{Journal of Biological Chemistry} \textbf{292}, 16638--16652 (2017). \href{https://doi.org/10.1074/jbc.M117.783837}{doi:10.1074/jbc.M117.783837}
\bibitem[Tobin(1958)]{tobin1958} Tobin, J. Estimation of relationships for limited dependent variables. \emph{Econometrica} \textbf{26}, 24--36 (1958). \href{https://doi.org/10.2307/1907382}{doi:10.2307/1907382}
\bibitem[Tsui et~al.(2025)]{tsui2025} Tsui, D., Musharaf, A., Erginbas, Y. E., Kang, J. S. \& Aghazadeh, A. SHAP zero explains biological sequence models with near-zero marginal cost for future queries. In \emph{Advances in Neural Information Processing Systems} \textbf{38} (2025). \href{https://arxiv.org/abs/2410.19236}{arXiv:2410.19236}
\bibitem[Watson et~al.(2023)]{watson2023} Watson, J. L. \emph{et al.} \emph{De novo} design of protein structure and function with RFdiffusion. \emph{Nature} \textbf{620}, 1089--1100 (2023). \href{https://doi.org/10.1038/s41586-023-06415-8}{doi:10.1038/s41586-023-06415-8}
\bibitem[W\"orn and Pl\"uckthun(2001)]{wornpluckthun2001} W\"orn, A. \& Pl\"uckthun, A. Stability engineering of antibody single-chain Fv fragments. \emph{Journal of Molecular Biology} \textbf{305}, 989--1010 (2001). \href{https://doi.org/10.1006/jmbi.2000.4265}{doi:10.1006/jmbi.2000.4265}
\bibitem[Wu et~al.(2019)]{wu2019} Wu, Z., Kan, S. B. J., Lewis, R. D., Wittmann, B. J. \& Arnold, F. H. Machine learning-assisted directed protein evolution with combinatorial libraries. \emph{Proceedings of the National Academy of Sciences} \textbf{116}, 8852--8858 (2019). \href{https://doi.org/10.1073/pnas.1901979116}{doi:10.1073/pnas.1901979116}
\bibitem[Yang et~al.(2019)]{yang2019} Yang, K. K., Wu, Z. \& Arnold, F. H. Machine-learning-guided directed evolution for protein engineering. \emph{Nature Methods} \textbf{16}, 687--694 (2019). \href{https://doi.org/10.1038/s41592-019-0496-6}{doi:10.1038/s41592-019-0496-6}
\end{thebibliography}
\end{document}